\documentclass[a4paper, 11pt]{article}   	
\usepackage{geometry}                		
\usepackage{graphicx}				
\usepackage{physics}								
\usepackage{amssymb}
\usepackage{amsmath}
\usepackage{algorithm}
\usepackage{authblk}
\usepackage{blindtext}
\usepackage{algpseudocode}
\usepackage[colorlinks=true,linkcolor=blue, citecolor = blue]{hyperref}
\newcommand{\bx}{\mathbf{x}}
\newcommand{\Cov}{\mathbb{C}\text{ov}}
\newcommand{\Var}{\mathbb{V}\text{ar}}
\newcommand{\Exp}{\mathbb{E}}
\usepackage{tikz}
\usetikzlibrary{patterns, arrows,positioning}
\usepackage[toc,title]{appendix}
\usepackage{mathtools}

\title{Emulating dynamic non-linear simulators using Gaussian processes}
\author[1, 2]{Hossein Mohammadi \thanks{Corresponding Author: h.mohammadi@exeter.ac.uk}}
\author[1, 2]{Peter Challenor \thanks{p.g.challenor@exeter.ac.uk}}
\author[1, 2]{Marc Goodfellow \thanks{m.goodfellow@exeter.ac.uk}}
\affil[1]{College of Engineering, Mathematics and Physical Sciences, University of Exeter, Exeter, UK}
\affil[2]{EPSRC Centre for Predictive Modelling in Healthcare, University of Exeter, Exeter, UK}
\date{}							
\begin{document}
	\maketitle

\begin{abstract}
The dynamic emulation of non-linear deterministic computer codes where the output is a time series, possibly multivariate, is examined. Such computer models simulate the evolution of some real-world phenomenon over time, for example models of the climate or the functioning of the human brain. The models we are interested in are highly non-linear and exhibit tipping points, bifurcations and chaotic behaviour. However, each simulation run could be too time-consuming to perform analyses that require many runs, including quantifying the variation in model output with respect to changes in the inputs. Therefore, Gaussian process emulators are used to approximate the output of the code. To do this, the flow map of the system under study is emulated over a short time period. Then, it is used in an iterative way to predict the whole time series.
A number of ways are proposed to take into account the uncertainty of inputs to the  emulators, after fixed initial conditions, and the correlation between them through the time series.
The methodology is illustrated with two examples: the highly non-linear dynamical systems described by the Lorenz and Van der Pol equations. In both cases, the predictive performance is relatively high and the measure of uncertainty provided by the method reflects the extent of predictability in each system.
\end{abstract}


\section{Introduction}
Computer models, e.g. numerical simulators, are sophisticated mathematical representations of some real-world phenomenon implemented in computer programs \cite{ohagan2006}. Such models are widely used in many fields of science and technology to aid our understanding of physical processes or because conducting physical experiments is too costly, highly time-consuming or even impossible in some cases  \cite{sacks1989}. Often, simulators are available as commercial packages and the underlying functions are unknown to the user. In most applications, it is crucial to understand the sensitivity of model outputs to variation or uncertainty in inputs \cite{ohagan2006}. Performing such quantitative studies requires a large number of simulation runs, see for example \cite{ferrat2018}. It becomes impractical if each simulation run is time-consuming.

Emulators, also known as \emph{surrogate models}, \emph{metamodels} or \emph{response surfaces} \cite{kleijnen2009} provide a ``fast'' approximation of complex simulation models using a limited number of training runs. The most popular classes of emulators are neural networks, splines, regression models, etc. We refer the reader to \cite{jin2000, chen2006, forrester2009} for more information on different types of emulators and their properties. Among the diverse types of emulators, Gaussian processes (GPs) have become increasingly popular over the last two decades in the field of the design and analysis of computer experiments \cite{sacks1989, santner2003, jones2009}. Also known as Kriging, especially in geostatistics \cite{cressie1993}, GPs have been effectively used in many real-world applications including wireless communication \newline \cite{schwaighofer2004}, metallurgy \cite{bailer-jones1997}, biology \cite{swain2016, kalaitzis2011, lawrence2006}, environmental science \cite{lee2011, challenor2004}, and sensor placements \cite{krause2008}. 

There are several reasons for the popularity of GPs. Firstly, they can be used to fit any smooth (with different degrees of smoothness), continuous function thanks to the variety of covariance kernels available \cite{neal1998}. See Section~\ref{GP_intro_sec} for more details on kernels. Secondly, GPs are non-parametric models, i.e., no strong assumptions about the form of the underlying function are required, unlike polynomial regression  \cite{rawlings2006}. Moreover, the prediction performance of GPs is comparable to (if not better than) other methods such as neural networks \cite{rasmussen1997, kamath2018}. The limit of a single layer neural network as the number of neurons tends to infinity is a Gaussian process \cite{mackay1998, neal1996}. The main advantage of GPs  is that they provide not only a mean predictor but also a quantification of the associated uncertainty. This uncertainty reflects the prediction accuracy and can serve as a criterion to enhance prediction capability of the emulator \cite{jin2002}.

This paper deals with the emulation of dynamic computer models that  simulate phenomena evolving with time. The output of a dynamic simulator is a time series for each input. The time series represents the values of the state variables at each time step. Such models are often expressed by a system of differential equations. 

Dynamic simulators appear in many applications. For instance, Stommel's box model \cite{stommel1961} simulates the evolution of temperature and salinity to determine the ocean density. In \cite{birrell2011} a dynamic model is developed whose output is a time series of general practice consultations for the 2009 A/H1N1 influenza epidemic in London. Since this model is computationally expensive,  a GP emulator is developed for calibration \cite{Farah2014}. Another real-world example of dynamic computer models is presented in \cite{Kuczera2006} where a saturated path hydrology model simulates the movement of water at catchment scales. In \cite{williamson2014} large climate models with time series output that exhibit chaotic behaviour are emulated using Bayesian dynamic linear model Gaussian processes. We refer to \cite{conti2010} for more examples on such simulators. 

There are many different proposed approaches for emulating dynamic simulators. According to \cite{reichert2011}, these approaches can be divided into four categories:
\begin{enumerate}
	\item One method is to use a multi-output emulator for predicting time series output \cite{conti2010}. In this case, the dimension of output space is $q = T$ where $T$ is the number of time steps the simulator is run for. However, when $T$ is large, the efficiency will reduce or may cause numerical problems. In addition, prediction is possible only for a fixed time horizon and one needs to repeat the prediction procedure for different time horizons.
	\noindent Building $q$ separate emulators for $q$ time points has the drawback of losing some information, as the correlation between various outputs (which we expect to be high) is not considered. Such correlation is taken into account in \cite{fricker2012, rougier2008} within multivariate emulators. However, as mentioned earlier, multivariate emulators are not efficient when the simulator's output is highly multivariate. 
	A common approach to alleviate this problem is to perform dimension reduction techniques on the output space such as principal components analysis \cite{higdon2008}  and wavelet decomposition \cite{bayarri2007}. A potential drawback of these techniques is that we may lose information by leaving out some components.
	\item A second approach is to treat time as an additional model input \cite{kennedy2001}. Gaussian processes have a computational complexity of $\mathcal{O}(n^3)$ where $n$ is the number of sample points.  Considering time as an extra parameter increases the computational cost to $\mathcal{O}(n^3 + T^3)$ using a separable covariance function \cite{plumlee2014}. As a result, the method can be burdensome when $T$ is large. Moreover, it is shown in \cite{conti2010} that the performance of multi-output emulators exceeds emulators with time as an extra input.
	\item One-step ahead emulations are another example in which the basic assumption is that the model output at a given time depends only on the previous output in time. Then, the transition function needs to be approximated. This method is reported to be efficient, \cite{conti2009}.  
	\item Finally, methods have been described that combine  stochastic dynamic models with innovation terms in the form of GPs. For example, in \cite{liu2009} a time-varying auto regression time series, which is a type of dynamic linear model, combined with GPs is used to emulate a dynamic computer code in a hydrological system. Similar work is carried out in \cite{williamson2014} with application to climate models.
\end{enumerate}

We propose a methodology based on iterative one-step ahead predictions. Given that simulating a long time series from the system is computationally expensive, our strategy is to emulate the flow map of the system over a short period of time. Then, we use the estimated flow map, which is computationally cheaper, to approximate the whole time series in an iterative way similar to the work in \cite{conti2009}. However, our method is different from that work in several ways. First, we build separate emulators to approximate each state variable that allows to have different covariance properties.  
Second, we propose a methodology to incorporate the uncertainty of inputs to the  emulators at time $t$ and the correlation between them through the time series, starting from a fixed initial condition. This is an important aspect of one-step ahead predictions because input to the GP model is uncertain after the first time step. Besides, it can be used as a criterion to estimate the predictability horizon of an emulator. Third, we emulate the flow map which is novel to our knowledge and has not yet been pursued.
\section{Gaussian processes as emulators}
\label{GP_intro_sec}
Let $f$ be the underlying function of an expensive simulator we want to approximate (or predict) defined as $f : \mathcal{X} \longmapsto \mathcal{F}$. Here, $\mathcal{X} \subset \mathbb{R}^d$ and $\mathcal{F} \subset \mathbb{R}^q$ are the input and output space respectively.
We further assume that $f$ is a ``black-box" function; there is no analytic expression for it and additional information such as gradients are not available. Also throughout this paper we assume the simulator to be \emph{deterministic} (vs. \emph{stochastic}); i.e. if it is run twice with the same inputs, the outputs will be identical. 

A GP defines a distribution over functions which can be regarded as a generalisation of the normal distribution to infinite dimensions. Formally, a GP indexed by $\mathcal{X}$ is a collection of random variables $\left\lbrace Z_\bx :~ \bx\in \mathcal{X} \right\rbrace$ such that for any $N \in \mathbb{N}$ and any $\bx^1, ..., \bx^N \in \mathcal{X}$, $\left(Z_{\bx^1}, ..., Z_{\bx^N} \right)^\top$ follows a multivariate Gaussian distribution \cite{GPML}. 
GPs are fully characterized by their mean function $\mu(.)$ and covariance kernel $k(.,.)$, which are defined as
\begin{align}
\label{prior_mean}
&\mu : \mathcal{X} \longmapsto \mathbb{R} ~; ~ \mu(\bx) = \Exp[Z_\bx]\\
&k : \mathcal{X} \times \mathcal{X} \longmapsto \mathbb{R} ~; ~ k(\bx, \bx^\prime) = \Cov\left(Z_\bx, Z_{\bx^\prime}\right).
\label{prior_var}
\end{align}

The mean function reflects our prior belief about the form of $f$. That is why $\mu(.)$ is also called the ``prior" mean within the Bayesian framework. While $\mu(.)$ could be any function, $k(.,.)$ must be symmetric positive definite. The most commonly used kernel is the squared exponential (SE) which has the form
\begin{eqnarray} 
\label{SE_kernel}
k\left(\bx, \bx^\prime\right) = \sigma^2 \prod\limits_{l=1}^d \exp \left(-\frac{\mid x_l- x^\prime_l\mid ^2}{2\theta^2_l} \right).
\end{eqnarray}
\noindent In the above equation, the parameter $\sigma^2$ is referred to as the \emph{process variance} and controls the scale of the amplitude of  sample paths.
The parameter $\theta_l$ is called the \emph{characteristic length-scale} and controls the degree of smoothness of sample paths along the coordinate $l~, 1 \leq l \leq d$. The Mat\'ern family of covariance functions is also widely used \cite{GPML}. Usually the kernel parameters are unknown and need to be estimated. Choosing appropriate kernel parameters has a huge impact on the accuracy of emulators. Maximum likelihood, cross validation or Bayesian estimation are common methods for this purpose.

\begin{figure}[!hb] 
	\centering
	\includegraphics[width=0.49\textwidth]{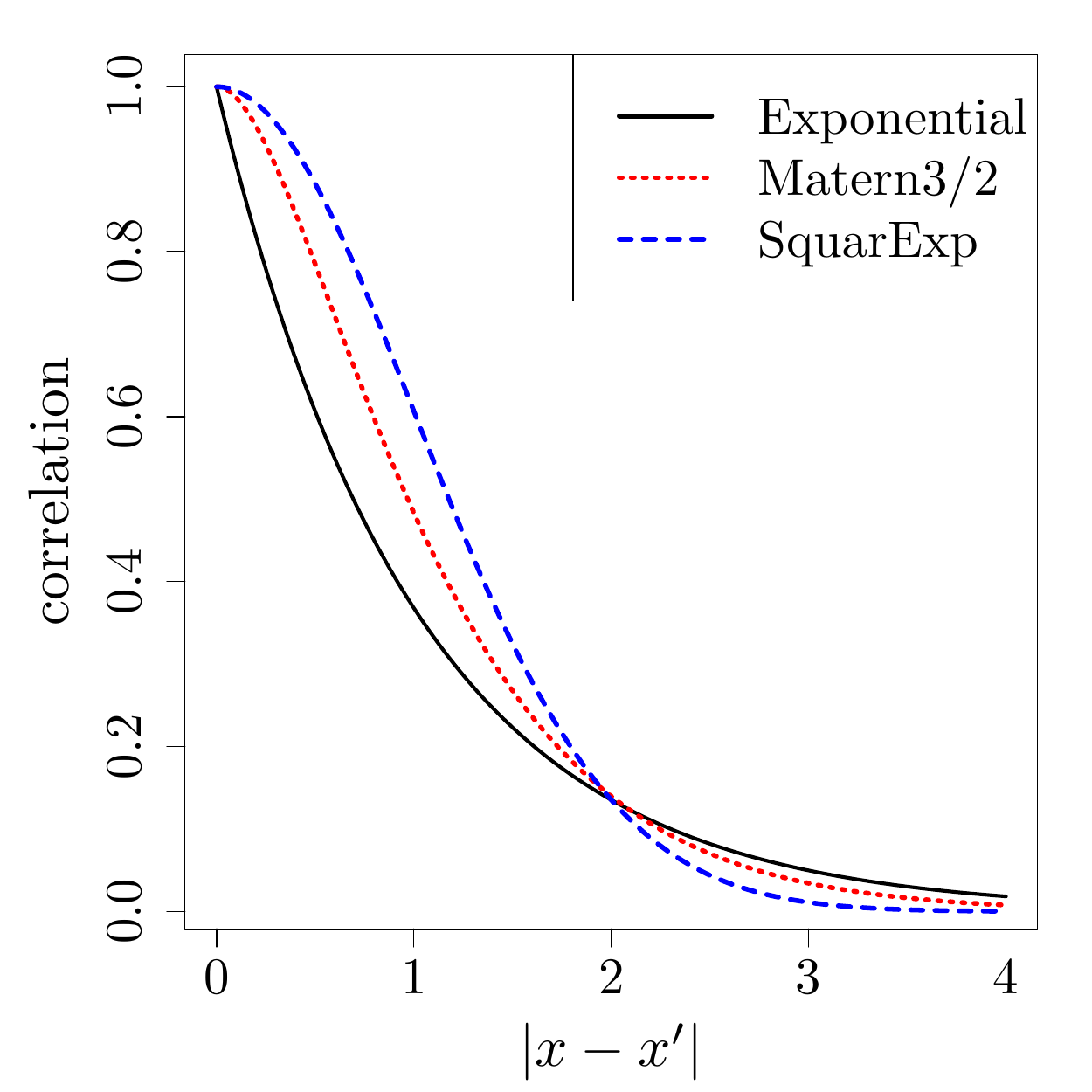}
	\includegraphics[width=0.49\textwidth]{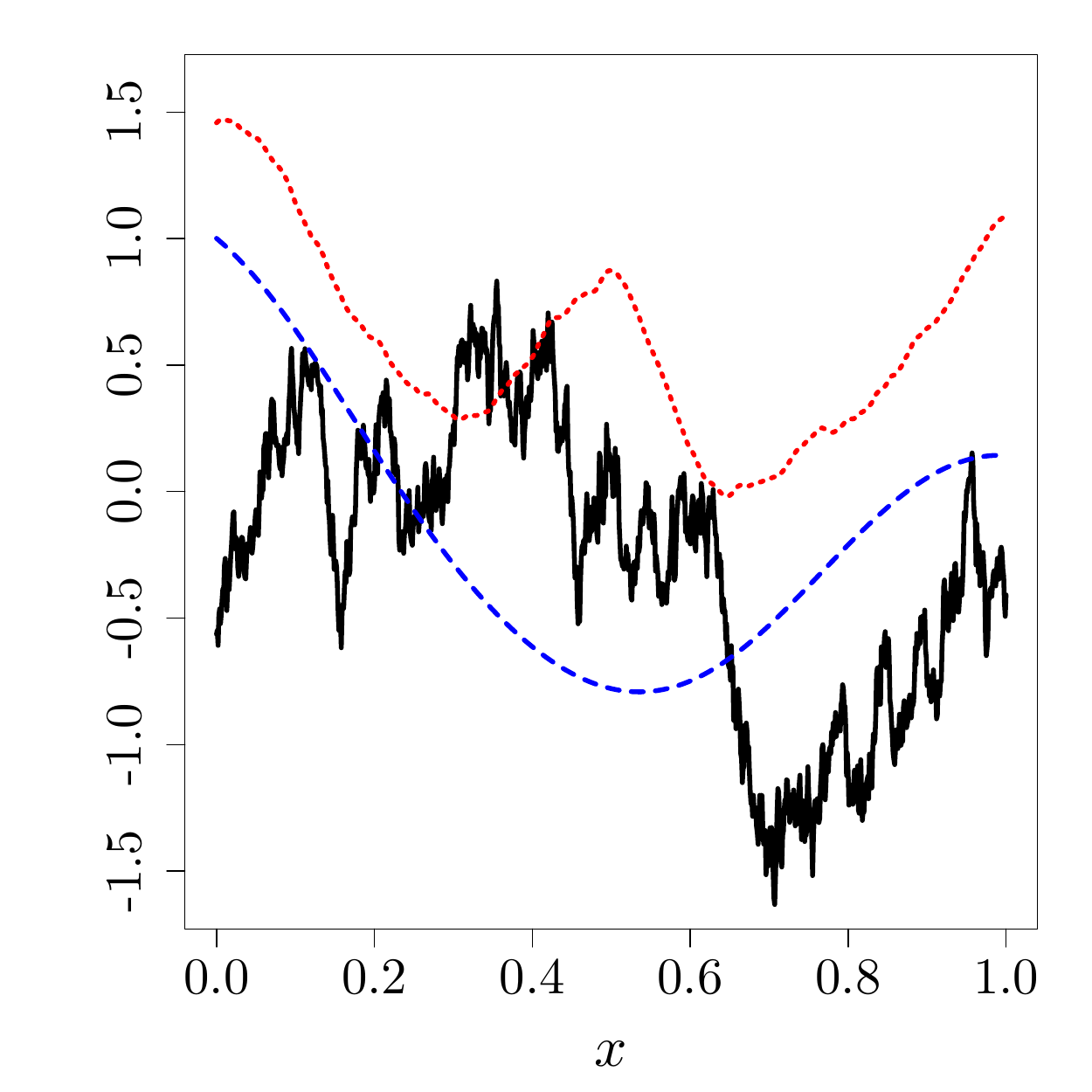}
	\caption{The structure of GP sample paths is determined by the covariance kernel. Left: Graphs of three (stationary) kernels: exponential (solid), Mat\' ern 3/2 (dotted), and squared exponential (dashed). Right: Three sample paths corresponding to the covariance kernels shown on the left picture. The process with squared exponential kernel is infinitely differentiable, whilst with Mat\' ern 3/2 the process is only once differentiable. The process with exponential kernel is not differentiable.} 
	\label{gp_paths}
\end{figure}

Covariance kernels play an important role in GP modelling. They customize the structure of sample paths of GPs. As an example, three different kernels (exponential, Mat\' ern 3/2, and SE, see \cite{GPML} for more information) and the associated sample paths are illustrated in Fig. \ref{gp_paths}. While in a process incorporating the SE kernel the sample paths are smooth (infinitely differentiable), they are only continuous (not differentiable) when the exponential kernel is used.  Herein, we consider \emph{stationary} covariance kernels that are translation invariant. The value of a stationary kernel depends only on the difference between input vectors. In other words, $k(\bx, \bx^\prime) = k(\bx + \mathbf{h} ,  \bx^\prime +  \mathbf{h})$ for any  $\mathbf{h} \in \mathbb{R}^d$.

\begin{figure}[!hb] 
	\centering
	\includegraphics[width=0.6\textwidth]{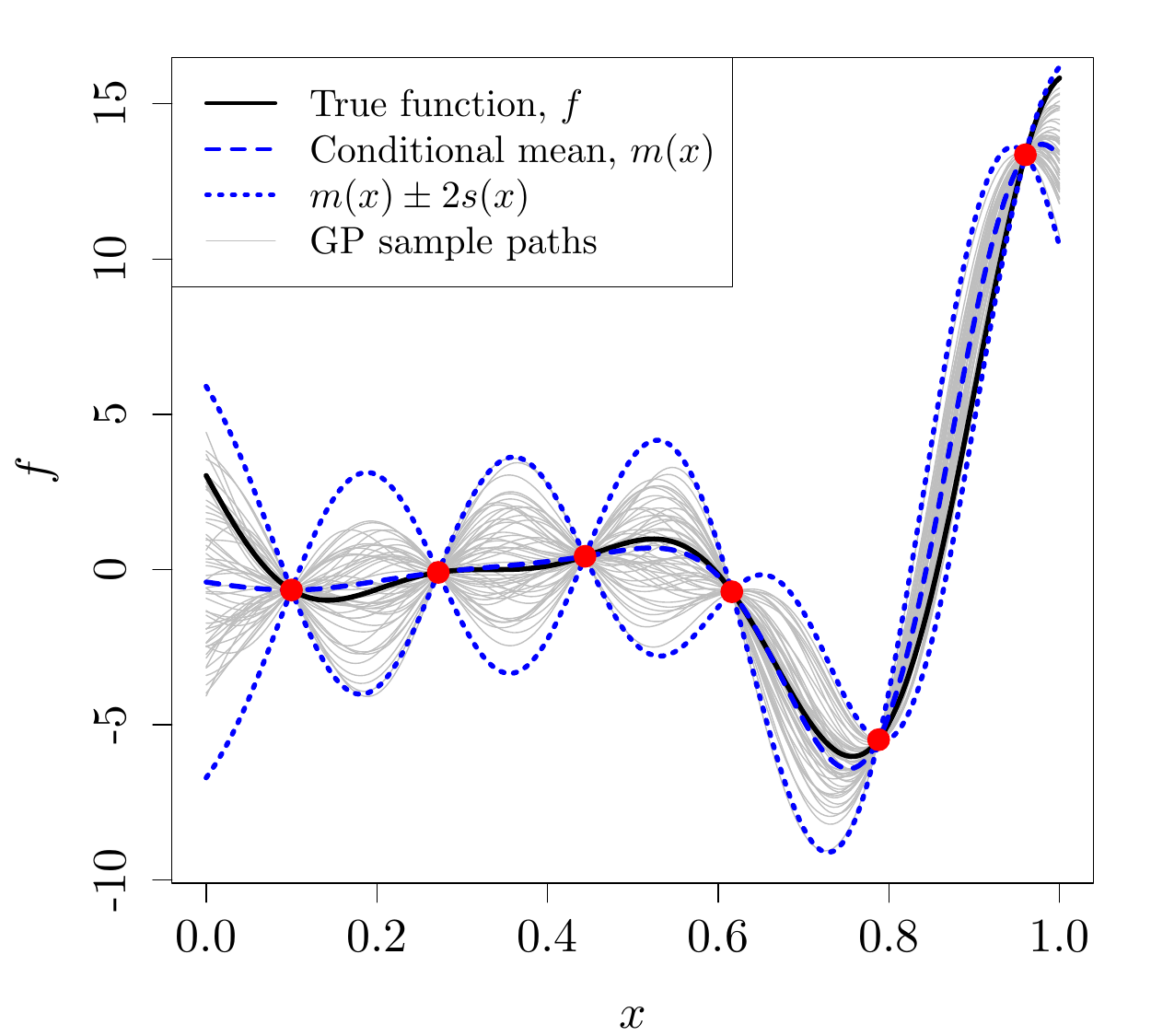}
	\caption{Gaussian process mean (thick  blue dashed line) conditional on 6 training samples (red bullets), which is also known as the conditional mean (denoted by $m(x)$), along with confidence intervals (thick dotted blue lines) equal to $m(x) \pm 2s(x)$. The true function is the thick solid line. The thin lines are 50 sample paths of the GP.} 
	\label{GP_prediction}
\end{figure}

To fit a GP, the true function $f$ is evaluated at $n$ locations $\mathbf{X}_n= \left\lbrace \bx^1, \dots, \bx^n \right\rbrace$ with the corresponding outputs (observations) $\mathbf{y} = \left( f(\bx^1), \dots, f(\bx^n) \right) ^\top$. Together,  $\mathbf{X}_n$ and $\mathbf{y}$ form the set of \emph{training} samples/data denoted by $D = \left\lbrace \mathbf{X}_n, \mathbf{y} \right\rbrace$. Then the conditional distribution of $Z_{\bx}$ is calculated as: \newline $ \left\lbrace  Z_{\bx}  ~\vert~ D : Z_{\bx^1} = f(\bx^1), \dots, Z_{\bx^n} = f(\bx^n) \right\rbrace $. 
If the mean function $\mu(.)$ is known, the prediction (conditional mean, $m(.)$) and its uncertainty (conditional variance, $s^2(.)$) at a generic location $\bx$ are of the form 
\begin{align}
\label{post_mean}
m(\bx) &=\mu(\bx) + \mathbf{k}(\bx)^\top  \mathbf{K}^{-1}(\mathbf{y} - \mu(\mathbf{X}_n) )\\
s^2(\bx) &= k(\bx, \bx) - \mathbf{k}(\bx)^\top  \mathbf{K}^{-1} \mathbf{k} (\bx), 
\label{post_var}
\end{align}
where $\mathbf{k}(\bx) = \left( k(\bx, \bx^i) \right)_{1 \leq i \leq n}$ is the vector of covariances between the observation at $\bx$ and the outputs at the $\bx^i$s and $\mathbf{K} = \left( k(\bx^i, \bx^j) \right)_{1 \leq i, j \leq n} $ denotes the matrix of covariances between sample outputs. Also, $\mu(\mathbf{X}_n)$ is the vector of mean function values at the training samples. The mean predictor obtained by Eq. (\ref{post_mean}) interpolates the points in the training data. Moreover, the prediction uncertainty vanishes at the training points and grows as we get further from them. An illustrative example is shown in Fig. \ref{GP_prediction}.
\section{Emulating dynamical simulators}
\label{dynamic_emulator}  
\subsection{One-step ahead emulation: general methodology} 
We wish to predict the output of a computationally expensive dynamical simulator relying on a $d$-dimensional autonomous system of ordinary differential equations (ODEs) of which the state variable is given by the real-valued vector $\bx(t) = \left(x_1(t), \dots, x_d(t) \right)^\top$. This system gives rise to the flow map $\Phi: \mathbb{R}^d \times \mathbb{R} \longmapsto \mathbb{R}^d $ such that $\bx( t_0 + \Delta t )= \Phi(\bx(t_0), \Delta t)$ for any $\bx(t_0)$ and $ \Delta t$. We are interested in a ``short" \emph{fixed} time step $\Delta t$ to give $\bx(t_1)$ at time $t_1 = t_0 + \Delta t$. Since $\Delta t$ is fixed, we consider the flow map as a function of $\bx(t_0)$ only.

\begin{figure}[!hb]
	\hspace{-2cm}
	\begin{tikzpicture}
	\draw (0, 0) rectangle (3.5, 3.5);
	\node[draw=none,fill=none] at (1.75, -0.25) {$x_1$};
	\node[draw=none,fill=none] at (-0.25, 1.75) {$x_2$};
	\node[draw=none,fill=none] at (1.75, 1.1) {$\bx(t_0)$};
	\node[draw=none, fill=none] at (4.75,3) {$\Phi\left(.\right)$};
	\hspace{6cm}
	\draw (0, 0) rectangle (3.5, 3.5);
	\node[draw=none,fill=none] at (1.75, -0.25) {$x_1$};
	\node[draw=none,fill=none] at (-0.25, 1.75) {$x_2$};
	\node[draw=none,fill=none] at (1.75, 1.3) {$\bx(t_1) = $};
	\node[draw=none,fill=none] at (1.75, 0.7) {\small{$\left(f_1(\bx(t_0)), f_2(\bx(t_0))\right)^\top$}};
	\draw[thick,black, ->] (-4.25, 1.75) .. controls (-1, 3) .. (1.75,1.75);
	\end{tikzpicture}
	\caption{Left: space of the initial conditions $\bx(t_0) = \left(x_1(t_0), x_2(t_0)\right)^\top$. Right: space of the solution of the system at time $t_1 = t_0 + \Delta t$, i.e. $\bx(t_1)$. The flow map $\Phi(.)$ maps from $\bx(t_0)$ to $\bx(t_1)$ : $\bx(t_1) = \Phi(\bx(t_0)) = \left(f_1(\bx(t_0)), f_2(\bx(t_0))\right)^\top$ where $f_1$ and $f_2$ are unknown functions approximated by GPs.} 
	\label{2D_example}
\end{figure}
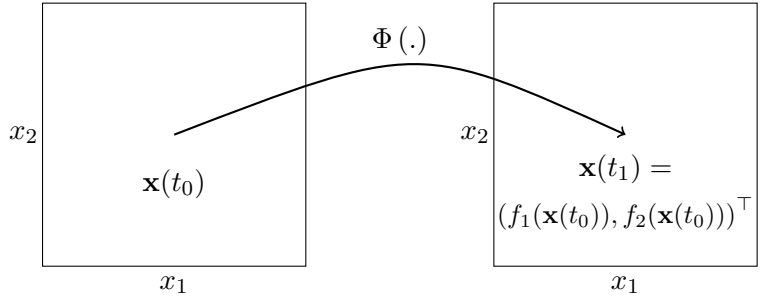

To predict $\bx(t)$ over time,  we assume that $\Phi (.)$ consists of $d$ components given by
\begin{align}
\Phi (.) \coloneqq \left(f_1\left(.\right), \dots,  f_d(.)\right)^\top ~, ~~ f_l: \mathbb{R}^d \longmapsto \mathbb{R} \, , ~ 1 \leq l \leq d ,
\label{flow_map}
\end{align}
such that each $f_l$ maps $\bx(t_0)$ to $x_l(t_1)$ the $l$-th component of $\bx(t_1)$. A $2D$ example is illustrated in Fig. \ref{2D_example} to clarify our assumption. Then, $f_l$s are treated as black-box functions that are replaced with their emulators denoted by $\hat{f}_l$s which are iteratively used for one-step ahead predictions over the time horizon $T$. 

The training set consists of $n$ initial conditions with the corresponding outputs, which are the solution of the system at time $t_1$, obtained by running the simulator over the short time horizon $\Delta t$. This training set is then used to approximate each function $f_l$ by a GP. The instructions are summarized in Algorithm \ref{alg1}. 

\begin{algorithm}[!hb] 
	\caption{Emulation of dynamic non-linear computer models} \label{alg1}
	\begin{algorithmic}[1]
		\State Select $n$ samples of initial conditions:  $\mathbf{X}_n = \left\lbrace \mathbf{x} ^{1}(t_0), \dots , \mathbf{x} ^{n}(t_0) \right\rbrace $  
		\State Run the simulator to obtain $\left\lbrace \mathbf{x} ^{1}(t_1), \dots , \mathbf{x} ^{n}(t_1) \right\rbrace $
		\For{$l=1$ to $d$}
		\State $\mathbf{y} \gets \left( x_l ^{1}(t_1), \dots , x_l ^{n}(t_1) \right)^\top$
		\State Build the $l$-th emulator $\hat{f}_l$, based on $D = \left\lbrace  \mathbf{X}_n,  \mathbf{y} \right\rbrace $
		\EndFor
		\State Predict over time horizon $T$ given initial condition $\mathbf{x} (t_0)$ as follow 
		\begin{align*}
		\hat{\mathbf{x}}^* (t_{1}) &= \left(\hat{f}_1 \left(\mathbf{x} (t_{0}) \right) , \dots ,  \hat{f}_d \left( \mathbf{x} (t_{0}) \right) \right)^\top \quad  \triangleright ~ \textit{$\bx$ with a superscript $\ast$ indicates that it is uncertain.} \\
		\hat{\mathbf{x}}^* (t_{2}) &= \left(\hat{f}_1 \left(\hat{\mathbf{x}}^* (t_{1}) \right) , \dots ,  \hat{f}_d \left(\hat{\mathbf{x}}^* (t_{1}) \right) \right)^\top \\
		&\vdots  \hspace{3cm}  \vdots\\
		\hat{\mathbf{x}}^* (t_{T}) &= \left(\hat{f}_1 \left(\hat{\mathbf{x}}^* (t_{T-1}) \right) , \dots ,  \hat{f}_d \left(\hat{\mathbf{x}}^* (t_{T-1}) \right) \right)^\top
		\end{align*} 
	\end{algorithmic}
\end{algorithm}

Note that in Algorithm \ref{alg1}, only the initial input to the emulators is certain. Thereafter, inputs are actually outputs of the emulators in the previous step. For example, to predict $\bx$ at $t_2= t_0 + 2\Delta t$, the input is $\hat{\mathbf{x}}^* (t_{1}) = \left(\hat{f}_1 \left(\mathbf{x}(t_{0}) \right) , \dots ,  \hat{f}_d \left( \mathbf{x}(t_{0}) \right) \right)^\top$ in which \newline $p\left( \hat{f}_l \left(\mathbf{x} (t_{0}) \right) \right) \sim \mathcal{N} \left( m_l(\bx(t_{0})), s_l^2(\bx(t_{0})) \right),~ 1 \leq l \leq d,$ see Eqs. (\ref{post_mean}) and (\ref{post_var}). So, we need to incorporate the input uncertainty in our modelling which is discussed below. Propagating such uncertainty, which is neglected in \cite{conti2009}, results in a more accurate representation of the uncertainty in the emulator over the time horizon.
\subsection{Emulation with uncertain input: uncorrelated emulators}
GPs with uncertain inputs have been studied in \cite{girard2003}, \cite{candela2003, kuss2006}. Suppose $\bx^*$ is drawn from a distribution that has mean $\boldsymbol{\mu}^*$ and variance $\boldsymbol{\Sigma}^*$. The probability distribution of the prediction at $\bx^*$ with the GP emulator $\hat{f}$ is determined by 
\begin{equation}
\begin{small}
p\left(\hat{f}(\bx^*)| \boldsymbol{\mu}^*, \boldsymbol{\Sigma}^*, D \right) = \int p\left(\hat{f}(\bx^*) |\bx^*, D \right) p(\bx^*) \,d\bx^* ,
\label{integral}
\end{small}
\end{equation}
where $p\left(\hat{f}(\bx^*) |\bx^*, D \right)$ has a normal distribution whose mean and variance are specified by Eqs. (\ref{post_mean}) and (\ref{post_var}). The integral in (\ref{integral}) is analytically intractable \cite{girard2003}. However, it can be approximated by different techniques which are divided into two groups: Monte Carlo-based methods and deterministic techniques such as Laplace's approximation. In this work, the former approach is used because it is simple, we only need to sample from $\mathcal{N} \left(\boldsymbol{\mu}^\ast , \boldsymbol{\Sigma}^\ast \right)$, and the approximated distribution will converge to the true distribution as the number of  samples grows \cite{girard2003, rasmussen1997}. We refer the reader to \cite{mackay2002} for more information on the deterministic techniques.

\begin{figure}[!hb] 
	\centering
	\includegraphics[width=0.49\textwidth]{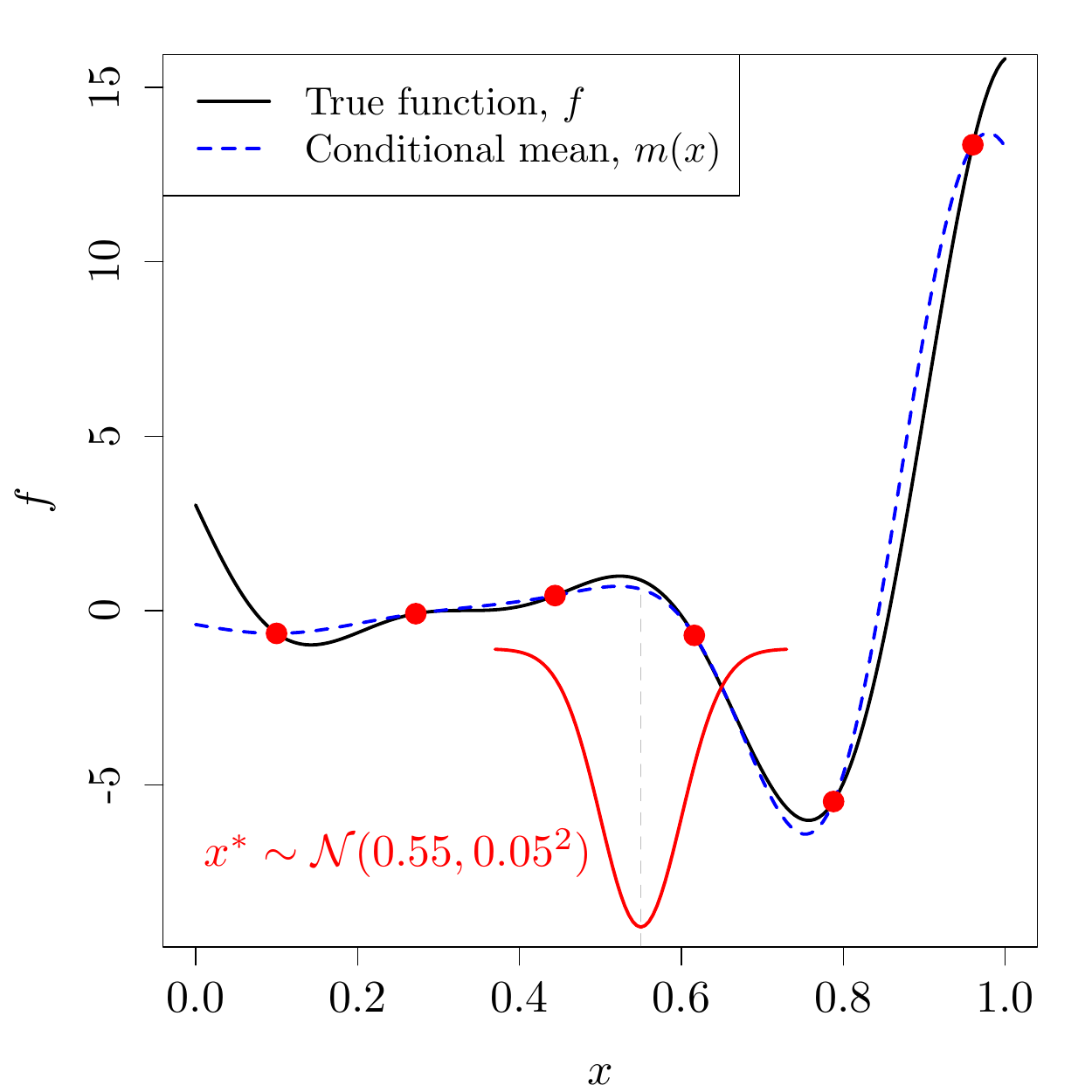}
	\includegraphics[width=0.49\textwidth]{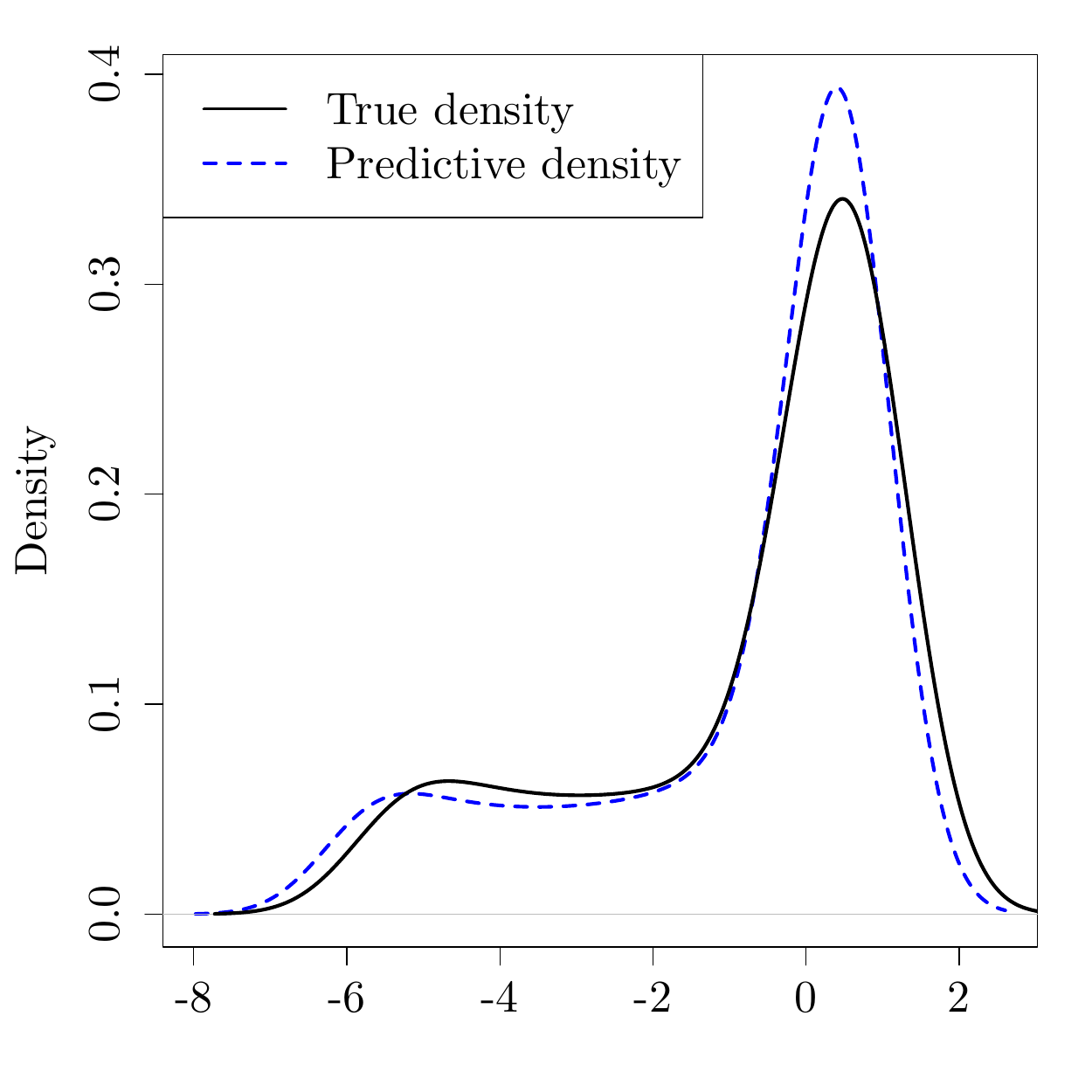}
	\caption{Estimating the probability distribution of the output given a distribution of inputs using the Monte Carlo method. Left: the model input is uncertain with the probability distribution: $x^\ast \sim \mathcal{N}(0.55, 0.05^2)$, as shown by the red line. Right: the estimated output distributions for the true function (solid) and the emulator (blue dashed). These estimates are based on $n_{MC} = 10000$ samples drawn from $x^\ast$ that are propagated in $f(.)$ and $m(.)$. Performing such a large number of function evaluations is impractical if $f$ is computationally expensive, but in our case this step is performed using the emulator.}
	\label{MC_method}
\end{figure}

Let $\bx^* \sim \mathcal{N} \left(\boldsymbol{\mu}^\ast , \boldsymbol{\Sigma}^\ast \right)$, the first and second moments of $p\left(\hat{f}(\bx^*)| \boldsymbol{\mu}^*, \boldsymbol{\Sigma}^*, D \right)$ using the law of iterated expectations and conditional variance  are given by
\begin{align}
\label{mean_uncertain_input}
\nonumber \Exp\left[ \hat{f}(\bx^*) | \boldsymbol{\mu}^*, \boldsymbol{\Sigma}^* \right] &=   \Exp_{\bx^*} \left[ \Exp_{\hat{f}(\bx^*)}   \left[  \hat{f}(\bx^*) \big| \bx^* \right]   \right]  \\
&= \Exp_{\bx^*} \left[  m(\bx^*)  \right]\\
\nonumber\Var\left[ \hat{f}(\bx^*) | \boldsymbol{\mu}^*, \boldsymbol{\Sigma}^* \right] &= \Exp_{\bx^*} \left[ \Var_{\hat{f}(\bx^*)}   \left[  \hat{f}(\bx^*) \big| \bx^* \right]   \right] \\
&+ \Var_{\bx^*} \left[ \nonumber \Exp_{\hat{f}(\bx^*)}   \left[  \hat{f}(\bx^*) \big| \bx^* \right]   \right] \\
&= \Exp_{\bx^*} \left[ s^2(\bx^*) \right] + \Var_{\bx^*} \left[ m(\bx^*) \right].
\label{var_uncertain_input}
\end{align}  
Computing quantities in (\ref{mean_uncertain_input}) and (\ref{var_uncertain_input}) is not straightforward because they are functions of the random variable $\bx^*$. In this work, $m(\bx^*)$ and $s^2(\bx^*)$ are approximated using a Monte Carlo (MC) method which relies on samples repeatedly drawn from a probability distribution and statistical analysis to infer the results \cite{raychaudhuri2008}. For example, to approximate $\Exp_{\bx^*} \left[ m(\bx^*) \right]$ in Eq. (\ref{mean_uncertain_input}), samples are repeatedly drawn from the random variable $\bx^\ast$. Then, they are propagated through the function $m(.)$ defined in Eq. (\ref{post_mean}). Finally, the desired quantity is approximated using 
\begin{equation}
\Exp_{\bx^*} \left[ m(\bx^*) \right] \approx \frac{1}{n_{MC}} \sum_{i= 1}^{n_{MC}} m(\bx^{*i}),
\label{MC_example}
\end{equation}
where $n_{MC}$ denotes the number of MC samples. To shed more light on the MC method, an illustrative example is demonstrated in Fig. \ref{MC_method}.
\subsection{Emulation with uncertain input: correlated emulators}
In the previous section, each element of $\bx (t) = \left(x_1(t), x_2(t), \dots, x_d(t)\right)^\top$ is emulated separately; $d$ different GP emulators denoted by $\hat{f}_1, \hat{f}_2, \dots , \hat{f}_d$ are employed independently such that the $l-$th emulator $\hat{f}_l$ emulates the transition function $f_l$ defined as $f_l: \bx(t_0) \longmapsto  x_l(t_1)$. However, we may lose some information if correlation between emulators is neglected. 

\begin{figure}[!hb] 
	\centering
	\includegraphics[width=0.49\textwidth]{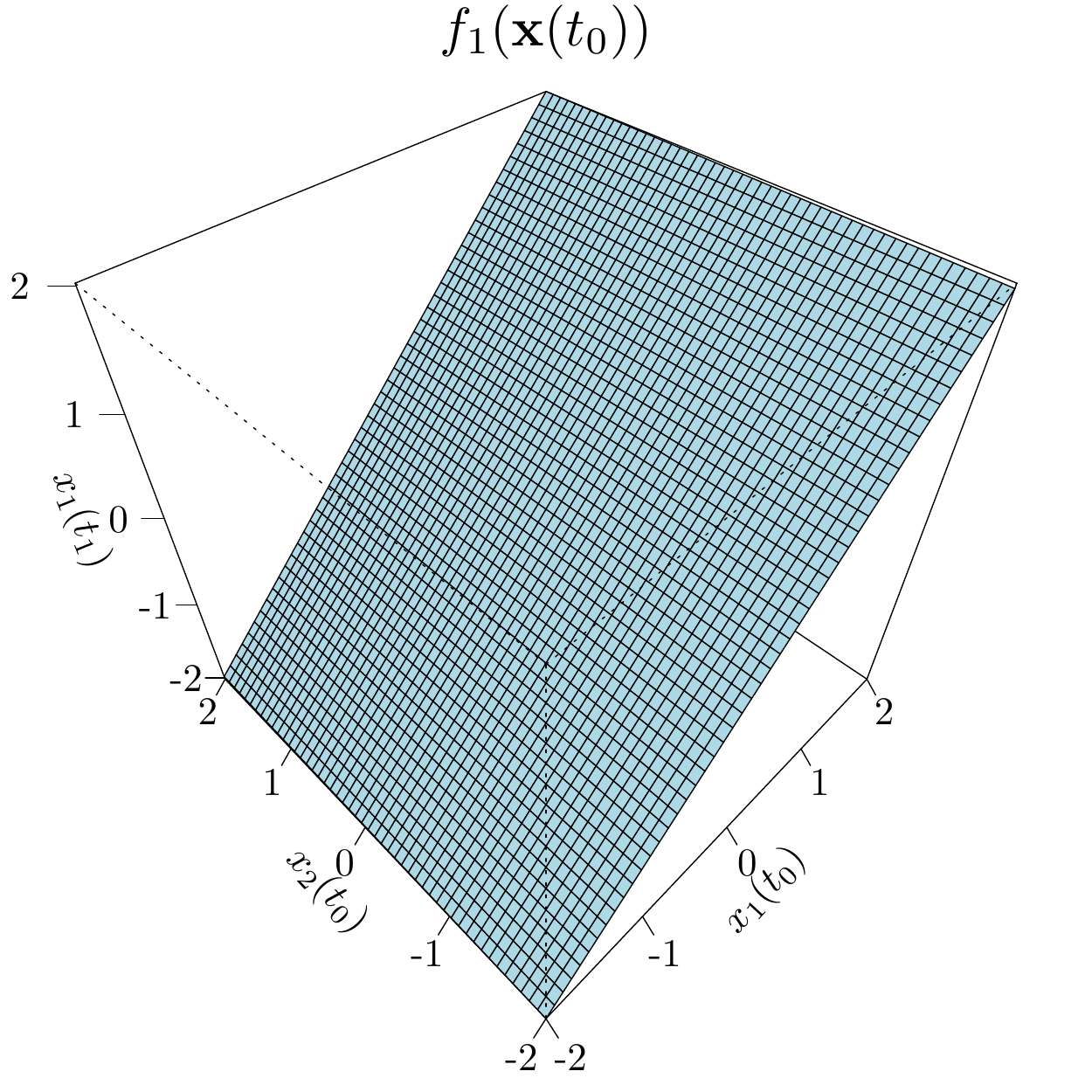}
	\includegraphics[width=0.49\textwidth]{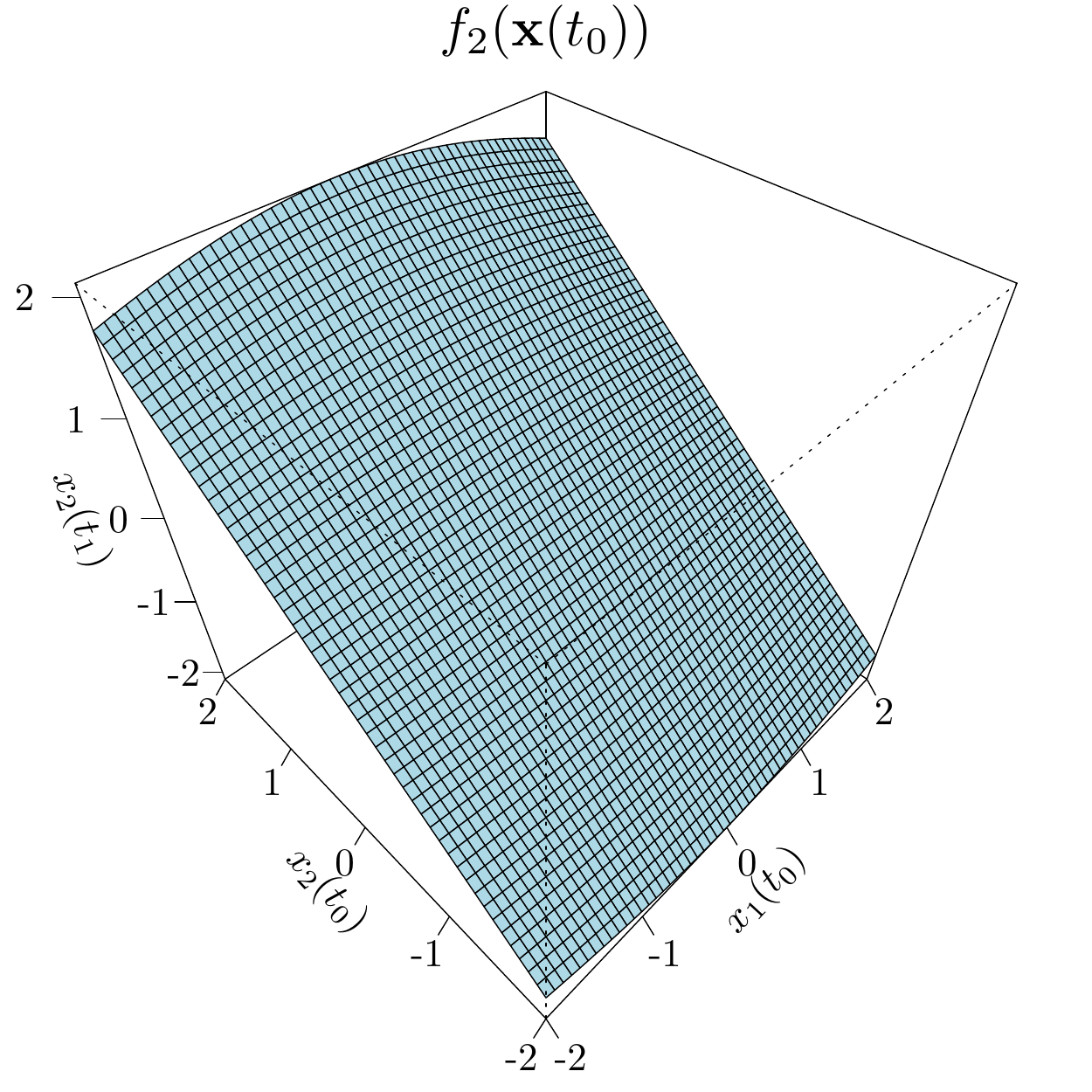}
	\caption{The flow map of the Van der Pol model as a function of the initial condition $\mathbf{x}(t_0)$ for the time step $\Delta t = 0.01$. Since the time step is small, the flow map is approximately linear. The two functions $f_1$ and $f_2$ are the components of the flow map defined as $\bx(t_1) = \Phi(\bx(t_0)) = \left(f_1(\bx(t_0)), f_2(\bx(t_0))\right)^\top$.} 
	\label{Vander_flowmap}
\end{figure}

Let $\bx^* \sim \mathcal{N}(\boldsymbol{\mu}^*, \boldsymbol{\Sigma}^*)$ be an uncertain input to the $d$ emulators. As a result, $\bx^{**} = \left( \hat{f}_1 (\bx^*), \dots,  \hat{f}_d (\bx^*) \right)^\top$ is a random vector whose mean is determined by
\begin{equation}
\boldsymbol{\mu}^{**} = \left( \Exp\left[ \hat{f}_1(\bx^*) | \boldsymbol{\mu}^*, \boldsymbol{\Sigma}^* \right], \dots, \Exp\left[ \hat{f}_d(\bx^*) | \boldsymbol{\mu}^*, \boldsymbol{\Sigma}^* \right]   \right)^\top.
\label{mean_approx}
\end{equation}
The elements in $\boldsymbol{\mu}^{**}$ are approximated using (\ref{MC_example}). Notice that $\bx^{**}$ is not necessarily a random normal variable, see Fig. \ref{MC_method} as an example in which the underlying function is highly non-linear. However, we approximate it by a Gaussian which has been used in similar works such as \cite{girard2003}. It is convenient to generate samples from a Gaussian distribution as we use the Monte Carlo method to approximate the unknown quantities. Note that if we use a very small time step relative to the size of the vector field ($\Delta t \to 0$), the change in $\bx(t_1)$ under the flow map for $\Delta t$ is very small. Therefore, the function will be approximately $f_l\left(\bx(t_0)\right) = \bx(t_0)$ (this is the limit) and the assumption of normality on $\bx^{**}$ is quite reasonable, see Fig. \ref{Vander_flowmap}.

Let $\boldsymbol{\Sigma}^{**}$ be the covariance matrix of $\bx^{**} $. In order to include the correlation between emulators, $\boldsymbol{\Sigma}^{**}$ must be of the form: 
\[
\begin{small}
\begin{bmatrix}
\Var\left[ \hat{f}_1(\bx^*) | \boldsymbol{\mu}^*, \boldsymbol{\Sigma}^* \right]&\dots&\Cov\left[ \hat{f}_1(\bx^*) ,  \hat{f}_d(\bx^*)| \boldsymbol{\mu}^*, \boldsymbol{\Sigma}^* \right]  \\
\vdots&\ddots&\vdots \\
\Cov\left[ \hat{f}_d(\bx^*) ,  \hat{f}_1(\bx^*)| \boldsymbol{\mu}^*, \boldsymbol{\Sigma}^* \right]&\dots& \Var\left[ \hat{f}_d(\bx^*) | \boldsymbol{\mu}^*, \boldsymbol{\Sigma}^* \right]
\end{bmatrix}.
\end{small}
\]
The diagonal elements of $\boldsymbol{\Sigma}^{**}$ are calculated using Eq. (\ref{var_uncertain_input}) which is approximated by the MC method. The off-diagonal elements, i.e. cross covariances, are given by:
\begin{small}
	\begin{align}
	\nonumber &\Cov\left[ \hat{f}_l(\bx^*) ,  \hat{f}_j(\bx^*)| \boldsymbol{\mu}^*, \boldsymbol{\Sigma}^* \right] = \Exp\left[ \hat{f}_l(\bx^*)   \hat{f}_j(\bx^*)| \boldsymbol{\mu}^*, \boldsymbol{\Sigma}^* \right] \\
	&- \Exp\left[ \hat{f}_l(\bx^*) | \boldsymbol{\mu}^*, \boldsymbol{\Sigma}^* \right] \Exp\left[ \hat{f}_j(\bx^*) | \boldsymbol{\mu}^*, \boldsymbol{\Sigma}^* \right]  ,~ 1 \leq l, j \leq d, l \neq j,
	\label{cross_cov}
	\end{align}
\end{small}
which are approximated by the MC method as below
\begin{small}
	\begin{align}
	\nonumber & \Cov\left[ \hat{f}_l(\bx^*) ,  \hat{f}_j(\bx^*)| \boldsymbol{\mu}^*, \boldsymbol{\Sigma}^* \right]  \approx 
	\frac{1}{n_{MC}}  \sum_{i = 1}^{n_{MC}} \hat{f}_l(\bx^{*i}) \hat{f}_j(\bx^{*i}) \\
	& - \left( \frac{1}{n_{MC}}  \sum_{i = 1}^{n_{MC}} \hat{f}_l(\bx^{*i}) \right)   \left( \frac{1}{n_{MC}}   \sum_{i = 1}^{n_{MC}} \hat{f}_j(\bx^{*i}) \right).
	\label{cross_cov_approx}
	\end{align}
\end{small}
For example, the mean and covariance of $\hat{\bx}^\ast(t_2)$ are obtained analogous to the way that $\boldsymbol{\mu}^{**}$ and $\boldsymbol{\Sigma}^{**}$ are computed. The uncertain input is $\hat{\bx}^\ast(t_1)$ with the following distribution
\begin{small}
\begin{equation*}
\mathcal{N} \left(
\begin{bmatrix} 
\Exp \left[ \hat{f}_1(\bx(t_0)) \right]  = m_1\left(\bx(t_0)\right)\\
\vdots \\
\Exp \left[ \hat{f}_d(\bx(t_0)) \right]	= m_d\left(\bx(t_0)\right)																																					
\end{bmatrix} ,
\begin{bmatrix} 
\Var \left[ \hat{f}_1(\bx(t_0)) \right] = s_1^2\left(\bx(t_0)\right) & \dots &  0 \\
\vdots & \vdots & \vdots \\
0 & \dots & \Var \left[ \hat{f}_d(\bx(t_0)) \right]	= s_d^2\left(\bx(t_0)\right)																																						
\end{bmatrix} \right) ,
\end{equation*}
\end{small}
where the mean and covariance are equivalent to $\boldsymbol{\mu}^\ast$ and $\boldsymbol{\Sigma}^\ast$ in Eqs. (\ref{mean_approx})-(\ref{cross_cov_approx}).
The same rules apply to predict the state variable at times $t = 3, 4, \ldots, T$.

It is worth mentioning that by applying a Gaussian distribution as a ``prior" on the input at each time step ($t \geq 2$), the process can be considered as part of a family of ``Deep Gaussian Processes" in which some properties of a GP are themselves another GP, see \cite{dunlop2017}. In that paper, they show that for the squared exponential kernel and certain parameter values, the process can become degenerate. We have not encountered this problem in any of our examples.

To shed more light on this method, the mean and the covariance matrix of the input in one-step ahead predictions are presented in Table \ref{tab1}. If cross covariances are not calculated and $\boldsymbol{\Sigma}^{**}$ is assumed to be a diagonal matrix, then it means the emulators are independent. Note that the GP emulators $\hat{f}_1, \dots, \hat{f}_d$ are independent and hence, the cross covariances are zero if the input to the emulators is deterministic, e.g. at $t = 1$. However, when the input is uncertain ($t \geq 2$), the predictive distribution of two independent GP become correlated \cite{kuss2006, deisenroth2009}.
\begin{table*}[htpb]	
	\caption{Mean and covariance matrix of the input in one-step ahead predictions}
	\label{tab1} 
	\begin{tabular}{lcr}
		\rule{14cm}{0.05cm} \\
		\textbullet ~  $t = 0$, the input is deterministic: \\ [5pt]
		
		$\bx(t_0) \sim \mathcal{N} \left(\bx(t_0),
		\begin{bmatrix} 
		0 & \dots &  0 \\
		\vdots & \vdots & \vdots \\
		0 & \dots & 0																																							
		\end{bmatrix} \right)$ \\ [30pt]
		\textbullet ~ $t = 1$, the cross covariances are zero because the emulators are independent: \\[5pt]
		
		\begin{footnotesize}
			$\hat{\bx}^*(t_1) \sim \mathcal{N} \left(
			\begin{bmatrix} 
			\Exp \left[ \hat{f}_1(\bx(t_0)) \right] \\
			\vdots \\
			\Exp \left[ \hat{f}_d(\bx(t_0)) \right]																																						
			\end{bmatrix} ,
			\begin{bmatrix} 
			\Var \left[ \hat{f}_1(\bx(t_0)) \right] & \dots &  0 \\
			\vdots & \vdots & \vdots \\
			0 & \dots & \Var \left[ \hat{f}_d(\bx(t_0)) \right]																																							
			\end{bmatrix} \right) $
		\end{footnotesize}  \\ [30pt]
		\textbullet ~ $t = 2$, the input to emulators are no longer deterministic and the method described \\
		in this section should be applied:\\[5pt]
		
		\begin{footnotesize}
			$\hat{\bx}^*(t_2) \stackrel{app.}{\sim} \mathcal{N} \left(
			\begin{bmatrix} 
			\Exp \left[ \hat{f}_1( \hat{\bx}^*(t_1)) \right] \\
			\vdots \\
			\Exp \left[ \hat{f}_d( \hat{\bx}^*(t_1)) \right]																																						
			\end{bmatrix} ,
			\begin{bmatrix} 
			\Var \left[ \hat{f}_1(\hat{\bx}^*(t_1)) \right] & \dots &  	\Cov\left[ \hat{f}_1(\hat{\bx}^*(t_1)) ,  \hat{f}_d(\hat{\bx}^*(t_1)) \right]  \\
			\vdots & \vdots & \vdots \\
			\Cov\left[ \hat{f}_d(\hat{\bx}^*(t_1)) ,  \hat{f}_1(\hat{\bx}^*(t_1)) \right]  & \dots & \Var \left[ \hat{f}_d(\hat{\bx}^*(t_1)) \right]																																							
			\end{bmatrix} \right) $
		\end{footnotesize} \\ [30pt]
		\hspace{0.6cm} $\vdots$	\hspace{2.4cm} $\vdots$	\hspace{5.7cm} $\vdots$  \\ [20pt]
		\textbullet ~ $t = T$: \\ [5pt]
		
		\begin{footnotesize}
			$\hat{\bx}^*(t_T) \stackrel{app.}{\sim} \mathcal{N} \left(
			\begin{bmatrix} 
			\Exp \left[ \hat{f}_1( \hat{\bx}^*(t_{n-1})) \right] \\
			\vdots \\
			\Exp \left[ \hat{f}_d( \hat{\bx}^*(t_1)) \right]																																						
			\end{bmatrix} ,
			\begin{bmatrix} 
			\Var \left[ \hat{f}_1(\hat{\bx}^*(t_{n-1})) \right] & \dots &  	\Cov\left[ \hat{f}_{1}(\hat{\bx}^*(t_{n-1})) ,  \hat{f}_d(\hat{\bx}^*(t_{n-1})) \right]  \\
			\vdots & \vdots & \vdots \\
			\Cov\left[ \hat{f}_d(\hat{\bx}^*(t_{n-1})) ,  \hat{f}_{1}(\hat{\bx}^*(t_{n-1})) \right]  & \dots & \Var \left[ \hat{f}_d(\hat{\bx}^*(t_{n-1})) \right]																																							
			\end{bmatrix} \right)$.
		\end{footnotesize} \\
		\rule{14cm}{0.055cm}
	\end{tabular}	
\end{table*}	
\section{Application to nonlinear dynamical systems}
In this section we first describe the emulator which is applied for predicting dynamic models. We then examine the prediction capability of the emulator on two well studied dynamical systems: the Lorenz and the Van der Pol systems, which are described in subsequent sections. 

The GP emulator we use in our experiments consists of a first order polynomial regression for the mean function in Eq. (\ref{prior_mean}) (i.e., $\mu(\bx) = \beta_0 + \beta_1\bx$) and a squared exponential kernel, given in Eq. (\ref{SE_kernel}), for the covariance kernel $k$. These choices of $\mu$ and $k$ are recommended in \cite{conti2009}. 
A set of training samples of size $n = 12d$, as recommended in \cite{jones1998, loeppky2009}, is drawn over the space of initial conditions.
Our training sample is constrained to lie in a cube. For example, a suitable boundary for the Lorenz attractor, as described in \cite{afraimovich1977, williams1979}, is given by the unstable manifold of the origin. To define a bounding box that contains the attractor, we therefore simulated the system with initial conditions close to the origin and chose as boundaries in each coordinate the extremes of the simulation.
Note we do assume that the system is constrained to lie in the same volume, although clearly we would like to capture most of its variation.
The points of the training samples should be selected based on a space-filling sampling scheme, and we therefore use a Latin hypercube \cite{stein1987, pronzato2012}. The goal in a space-filling design is to spread the points \emph{evenly} within the input space.
No attempt is made to have the points lie along the stable manifold, we simply try to `fill' space.

To build each emulator $\hat{f}_l ,~ 1 \leq l \leq d$, the training data consists of 
$\mathbf{X}= \left\lbrace \mathbf{x} ^{1}(t_0), \dots , \mathbf{x} ^{n}(t_0) \right\rbrace$ with the corresponding outputs $\mathbf{y}= \left(  x_l ^{1}(t_1), \dots , x_l ^{n}(t_1) \right)^\top$. 
The \texttt{R} package \texttt{DiceKriging} \cite{roustant2012} is employed to fit the GP emulator. The unknown parameters of the SE kernel $k$ (i.e. $\sigma$ and $\theta_l$s) and the mean function $\mu$ (i.e. $\beta_0$ and $\beta_1$) are estimated by maximum likelihood implemented in \texttt{DiceKriging}. After building the emulators, their accuracies are assessed by the leave-one-out cross-validation mean squared error ($MSE_{LOO}$) defined as 
\begin{equation}
MSE_{LOO} = \frac{1}{n} \sum_{i = 1}^{n} \left( \hat{f}_{l, -i} \left(\mathbf{x}^{i}(t_0)\right) - x_l^i (t_1) \right)^2.
\label{LOO}
\end{equation}
In the above equation, $\hat{f}_{l, -i} \left(\mathbf{x}^{i}(t_0)\right)$ is the prediction obtained by the GP emulator   $\hat{f}_l$ based on all the data points in $\mathbf{X}$ except the $i$-th one.

The ODEs are solved by the default solver of the \texttt{R} package \texttt{deSolve} \cite{soetaert2010}  which is called ``\texttt{ode}". It is based on a variable order method to integrate the system over the next step ahead, i.e. $t_1= t_0 + \Delta t$. More precisely, it uses the LSODA (Livermore solver for ordinary differential equations with automatic switching between stiff and nonstiff methods) method \cite{petzold1983}. A full Jacobian matrix is used which is calculated internally by LSODA. In these two examples, we use a fixed time step equal to $\Delta t = 0.01$. 
In the following sections, we first apply the method of uncorrelated emulators and subsequently examine the method of correlated emulators on two dynamical systems, i.e. the Lorenz and Van der Pol models.
\subsection{Lorenz system: uncorrelated emulators}
The Lorenz system was first proposed by Edward Lorenz in 1963 \cite{lorenz1963} as a mathematical representation of atmospheric convection. It is a three-dimensional system of  ordinary differential equations. Under certain choices of parameters it can display chaotic behaviour, i.e. its behaviour is highly sensitive to initial conditions. The evolution of three state variables is described by \cite{soetaert2010}
\begin{equation}
\begin{cases}
\frac{d x_1}{d t} = a x_1 + x_2x_3 \\  \frac{d x_2}{d t} = b(x_2 - x_3) \\  \frac{d x_3}{d t} = -x_1x_2 + c x_2 - x_3,
\end{cases}
\label{lorenz}
\end{equation}
where $a$, $b$ and $c$ are parameters. Here, we assume $a = -8/3$, $b = -10$ and $c = 28$. We focus on the case with initial conditions $\bx(t_0) = (x_1(t_0)= 1, x_2(t_0) = 1, x_3(t_0) = 1)^\top$.  The accuracy of the emulators is high based on the $MSE_{LOO}$ criterion which is given below.  
\begin{center}
	\begin{tabular}{l | l c r }
		& $\hat{f}_1$ & $\hat{f}_2$ & $\hat{f}_3$ \\
		\hline
		$MSE_{LOO}$ & $1.946 \times 10^{-4}$ & $3.533 \times 10^{-7}$ & $1.288 \times 10^{-4}$
	\end{tabular}
\end{center}

\begin{figure}[htpb] 
	\includegraphics[width=0.49\textwidth]{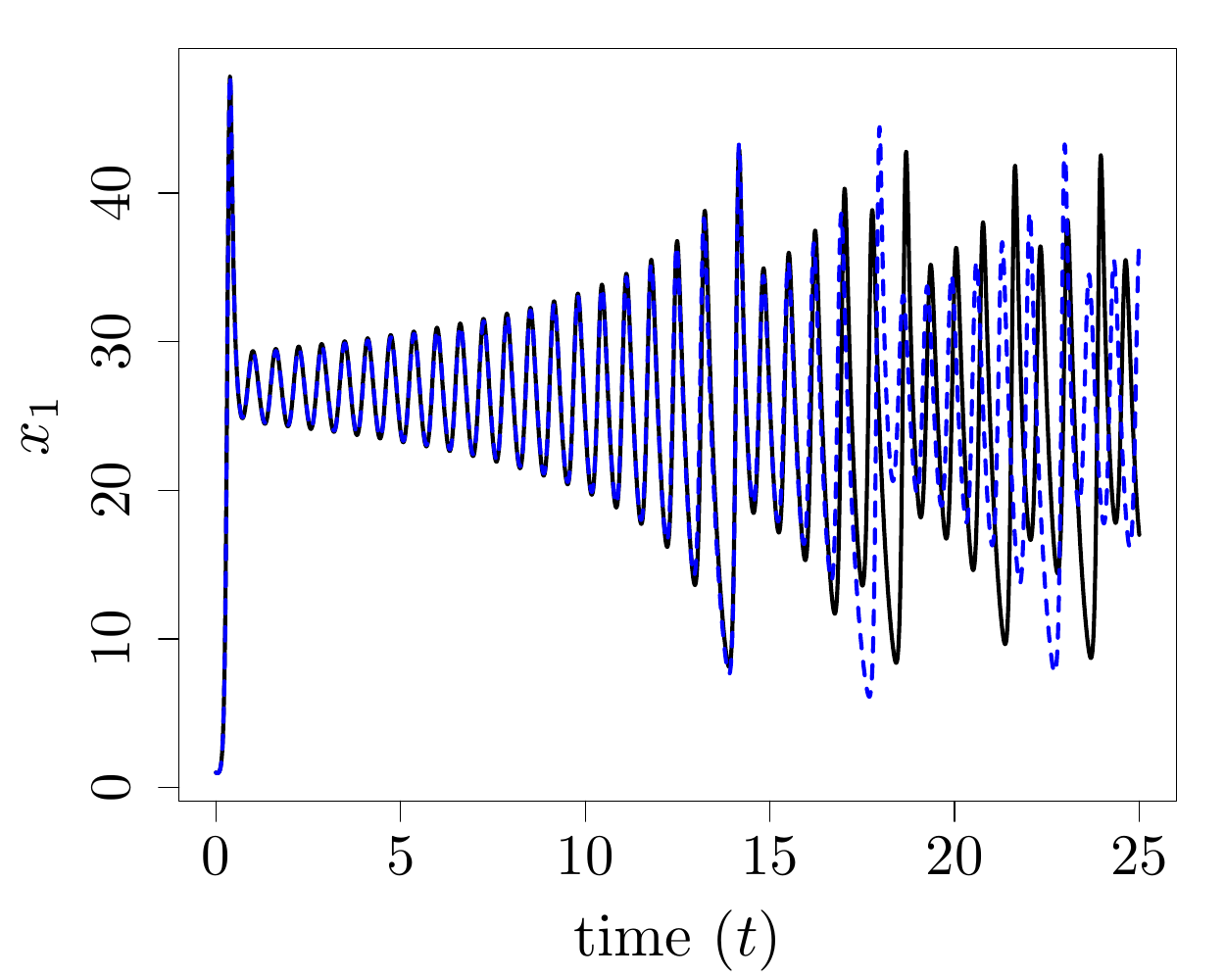}
	\includegraphics[width=0.49\textwidth]{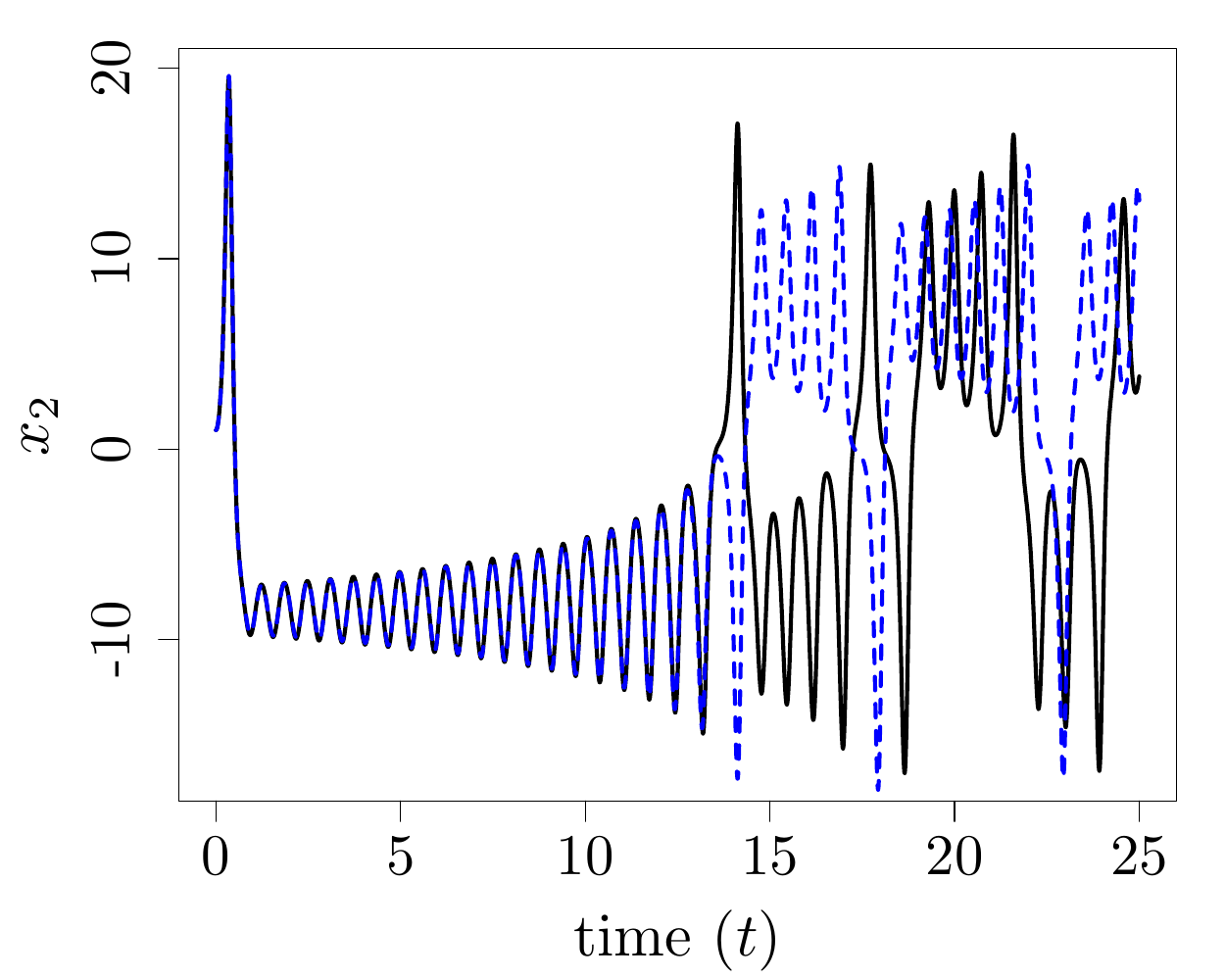}
	\includegraphics[width=0.49\textwidth]{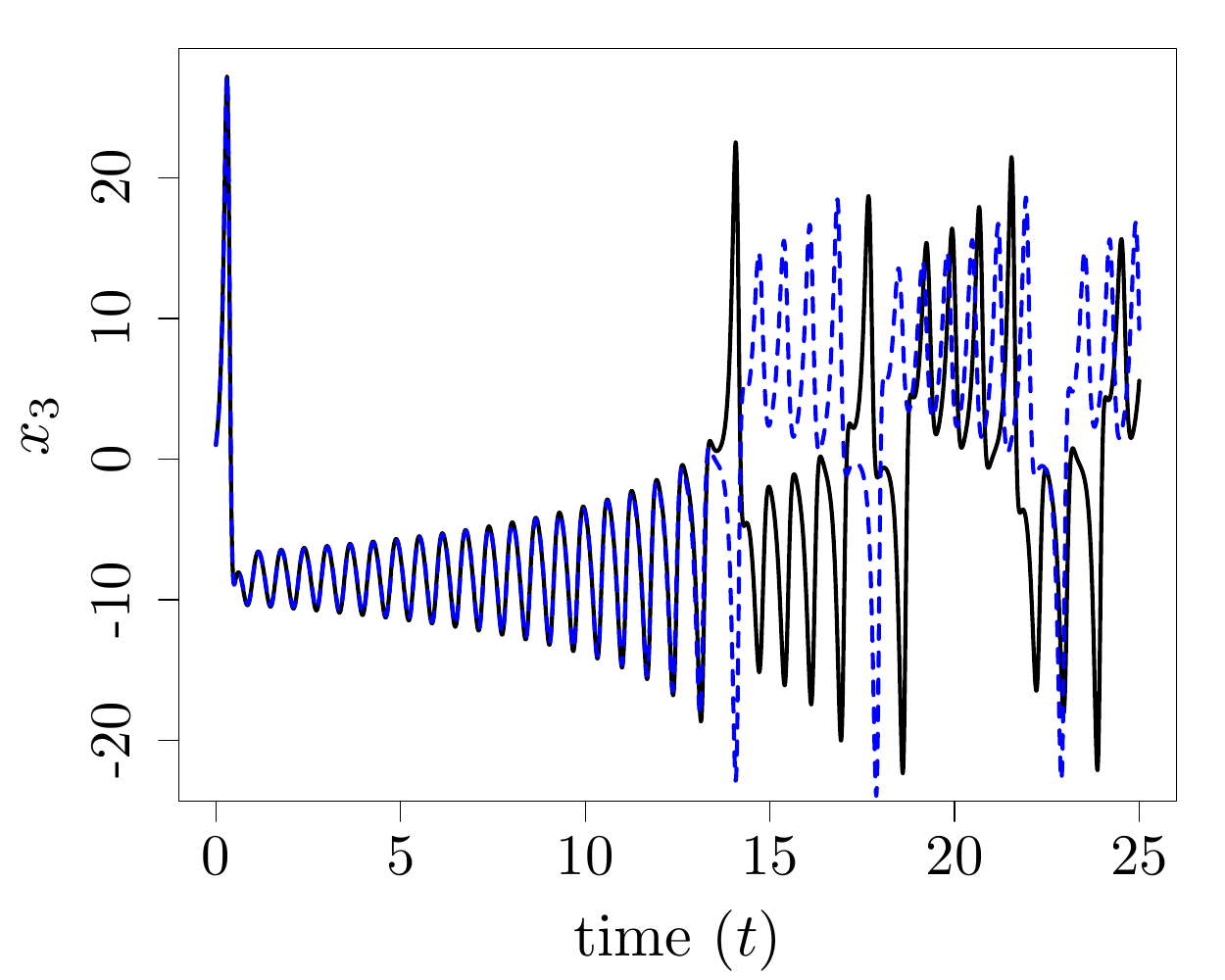} 
	\includegraphics[width=0.47\textwidth]{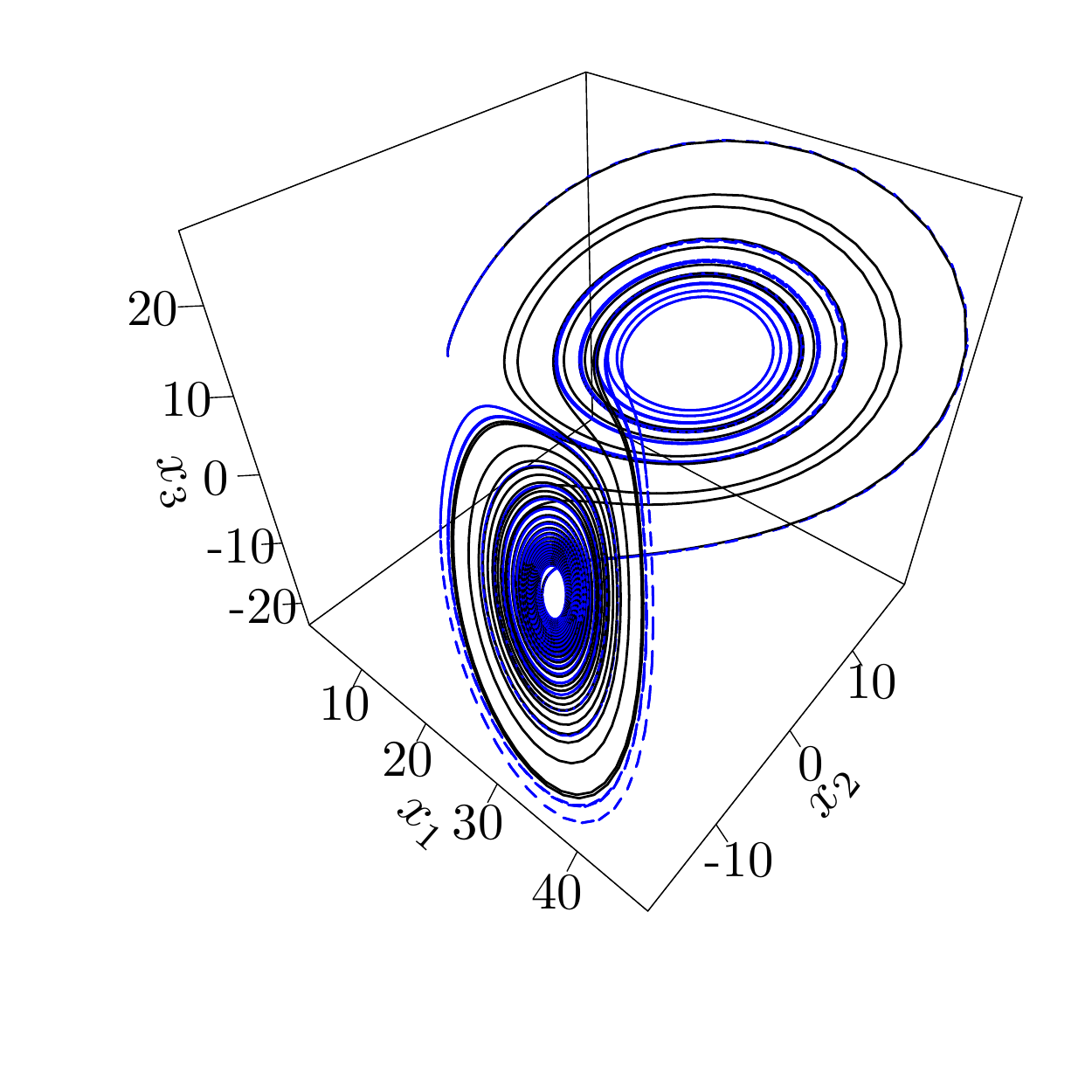}
	\caption{The Lorenz system (solid black) and its emulation (blue dashed) considering the input uncertainty, but neglecting correlation between emulators, as described in Algorithm \ref{alg1}. The 1-D pictures illustrate evolution of state variables $x_1, x_2, x_3$ and their predictions. The emulators built based on iterative one-step ahead predictions are able to well predict up to about $t = 14$. The 3-D picture shows the evolution and prediction of the whole system.}
	\label{Lorenz_MC}
\end{figure}

Emulation of the Lorenz model using the iterative one-step ahead predictions considering the input uncertainty, but neglecting correlation between emulators, as described in Algorithm \ref{alg1} is demonstrated in Fig. \ref{Lorenz_MC}.  We show the evolution of predictions for each system variable over time, as well as a three-dimensional picture showing the evolution of the whole system, $\left(x_1(t), x_2(t), x_3(t) \right)^\top$. The solid line represents the true model and the blue dashed line is the GP prediction. It can be seen that the prediction precision is high at the beginning of the time course, for example $t \leq 14$. However, the emulator deviates from the true model as time progresses. Fig. \ref{Lorenz_MC} suggests that the emulator is well suited to describing the evolution of the system within a ``wing" of the Lorenz attractor, but that predictions break down upon switching to the other part of the attractor.  

\begin{figure}[htpb] 
	\includegraphics[width=0.49\textwidth]{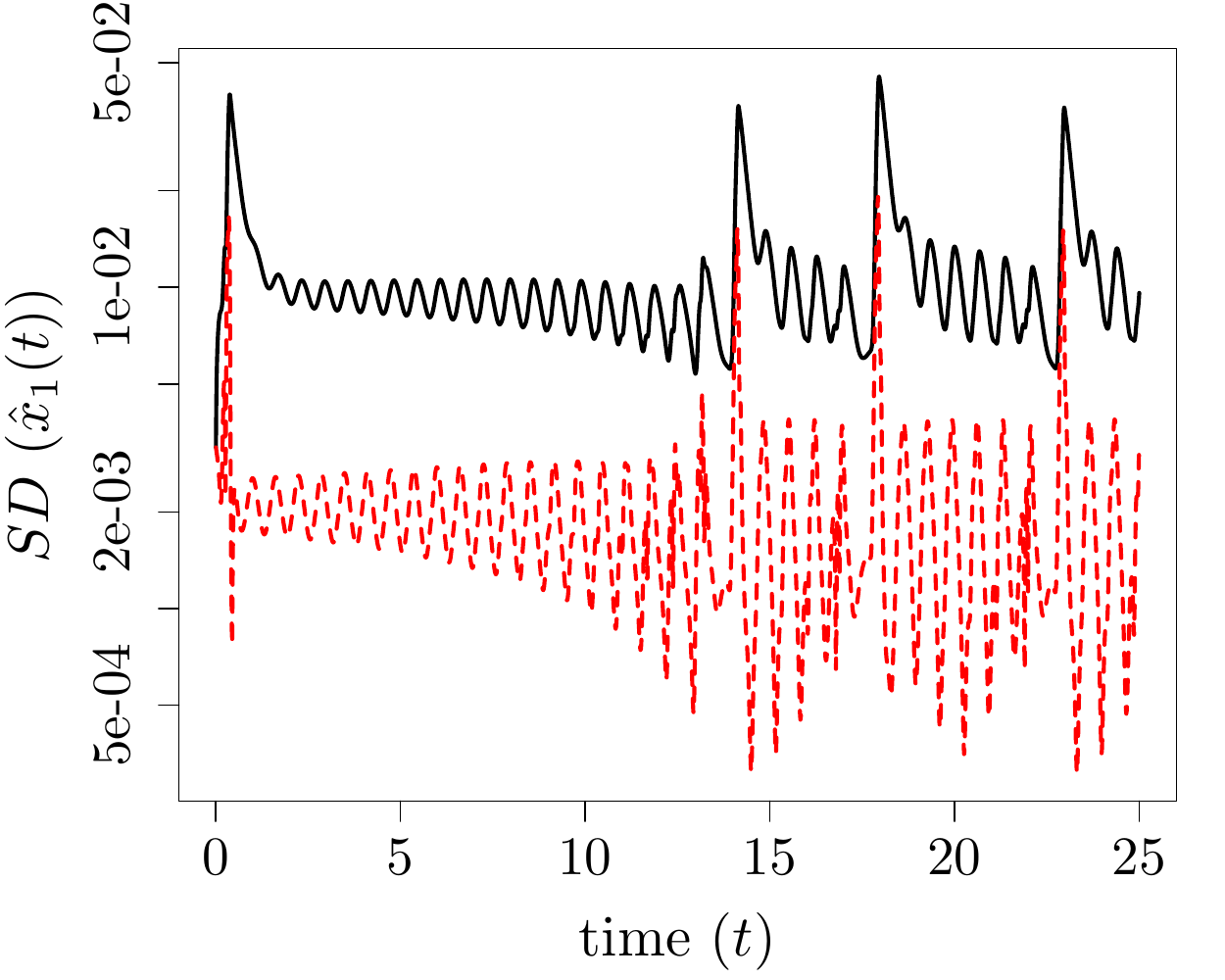}
	\includegraphics[width=0.49\textwidth]{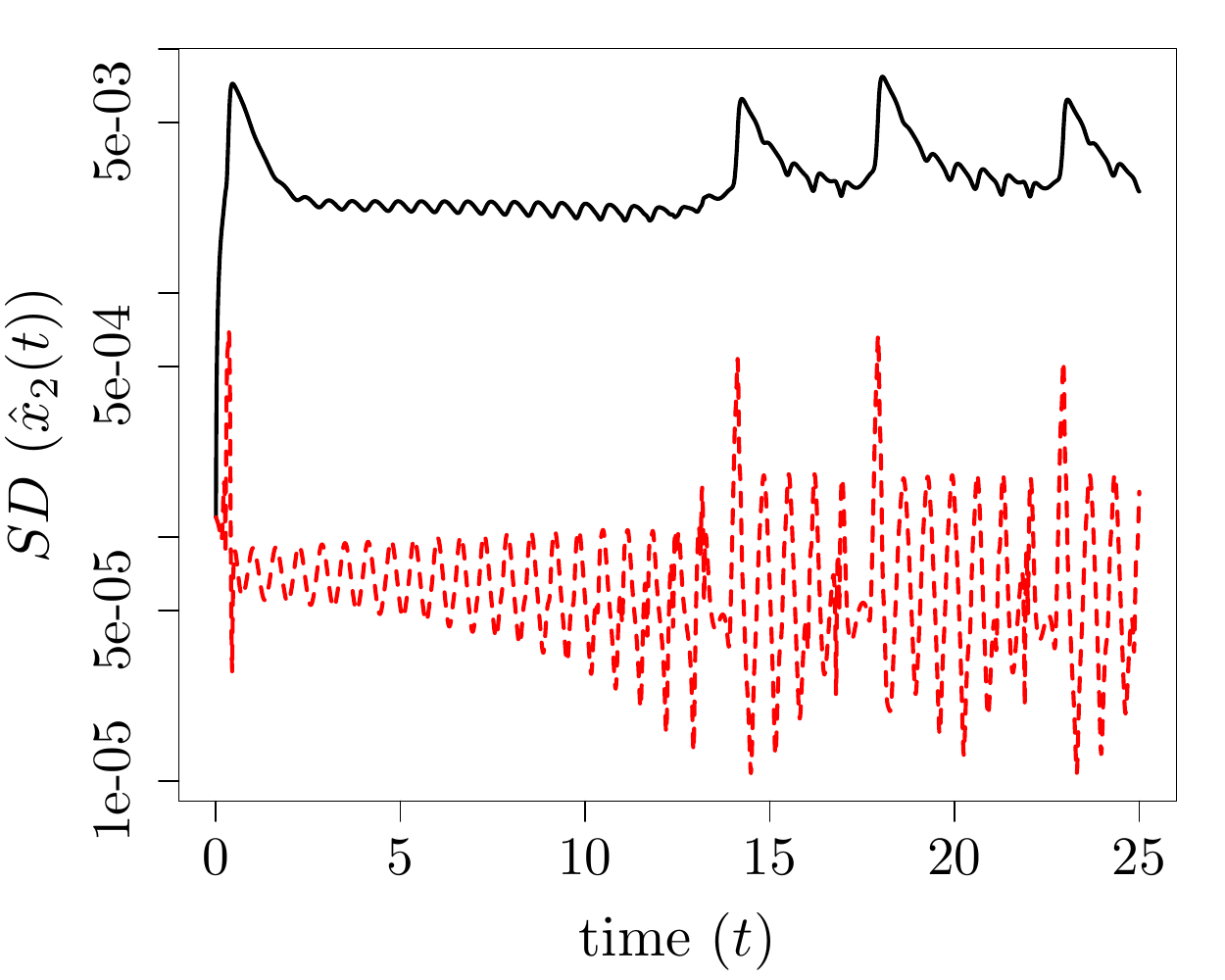} \\
	\centering
	\includegraphics[width=0.49\textwidth]{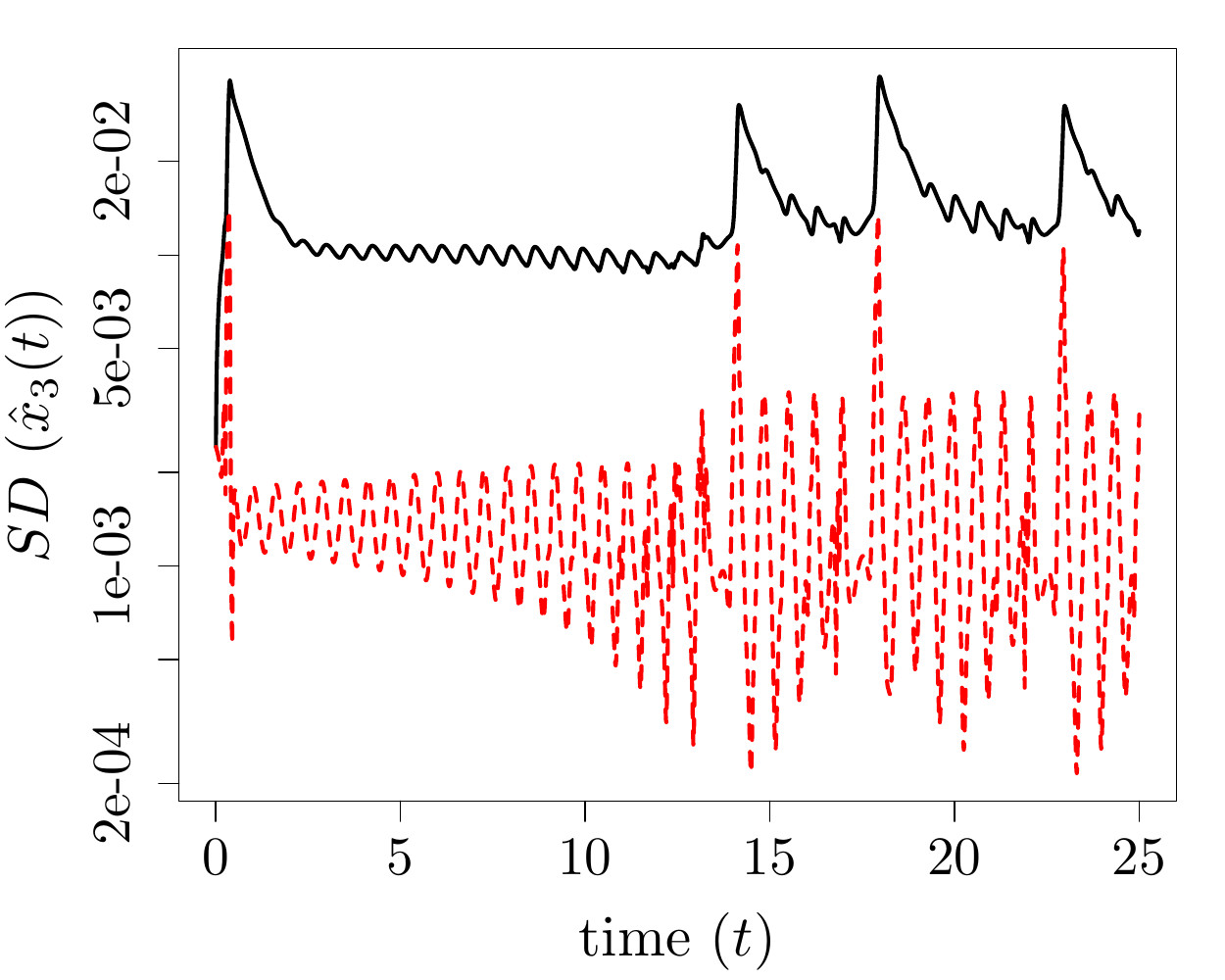}
	\caption{Standard deviation (SD) associated with the predictions of state variables of Lorenz system (Fig. \ref{Lorenz_MC}) with  and without considering input uncertainty, solid black and red dashed lines respectively. Incorporating input uncertainty augments the prediction SD. Contrary to our expectations, the uncertainties do not grow with time.}
	\label{Lorenz_MC_SD}
\end{figure}

Fig. \ref{Lorenz_MC_SD} shows the uncertainties (solid black) associated with the predictions illustrated in Fig. \ref{Lorenz_MC}. The uncertainties are compared with the case in which the input uncertainty is not considered (red dashed line). Generally, if emulation is carried out with uncertain inputs, the magnitude of uncertainties is higher. 
Nevertheless, they are still too small and contrary to our expectations do not increase over time as the uncertainty builds up from step to step.
The true model is not inside the credible intervals, which are defined as $m(\bx) \pm 2s(\bx)$. Note the credible intervals are not shown on Fig. \ref{Lorenz_MC}, but can easily be derived from Figs. \ref{Lorenz_MC} and \ref{Lorenz_MC_SD}. In particular, we would expect the uncertainty to ``blow up" when we reach the point of switching between wings of the attractor (at about $t=14$) where our emulator can be on a different wing to the true model but still has very small uncertainty.       
\subsection{Van der Pol oscillator: uncorrelated emulators}
\begin{figure}[htpb] 
	\centering
	\includegraphics[width=0.49\textwidth]{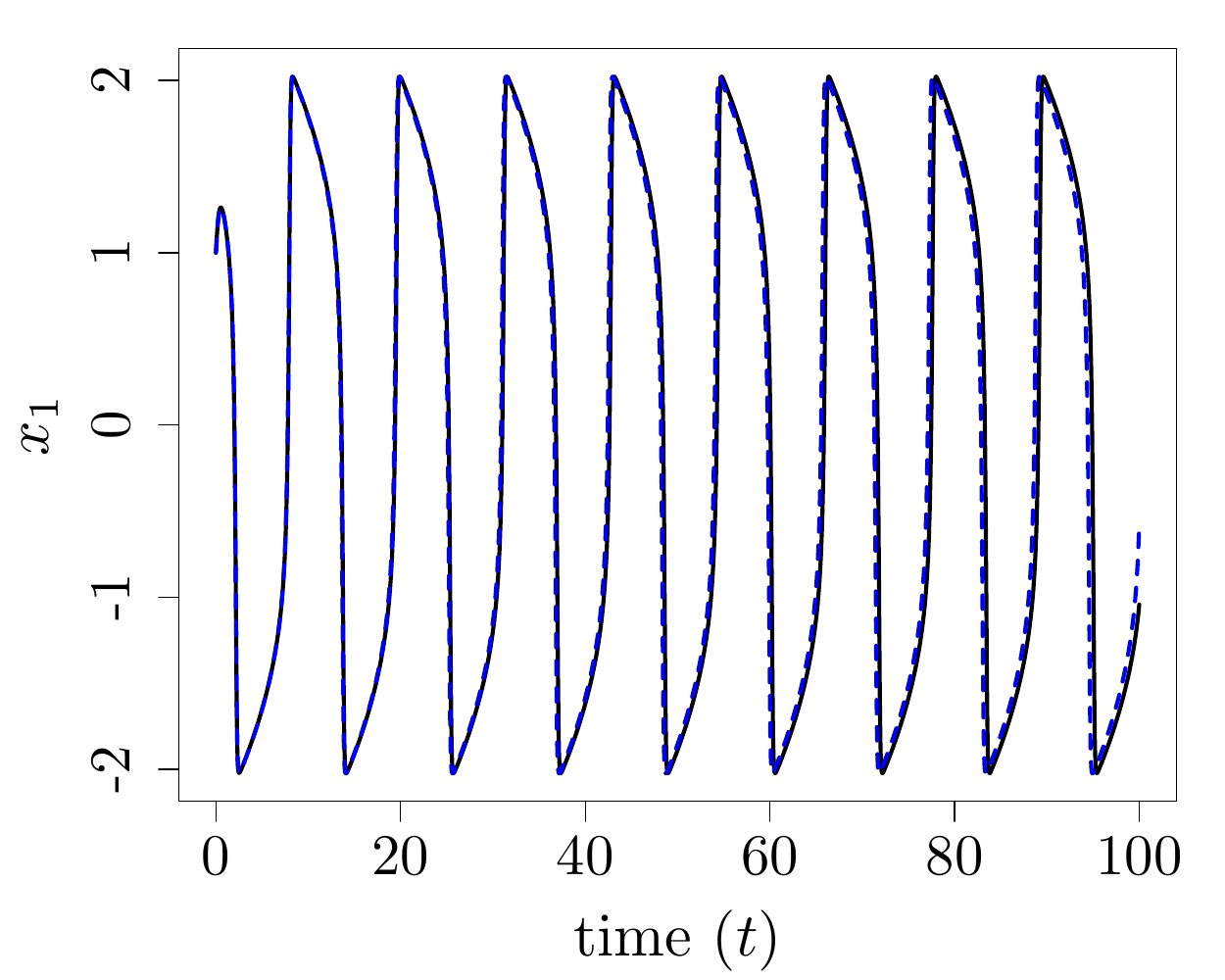}
	\includegraphics[width=0.49\textwidth]{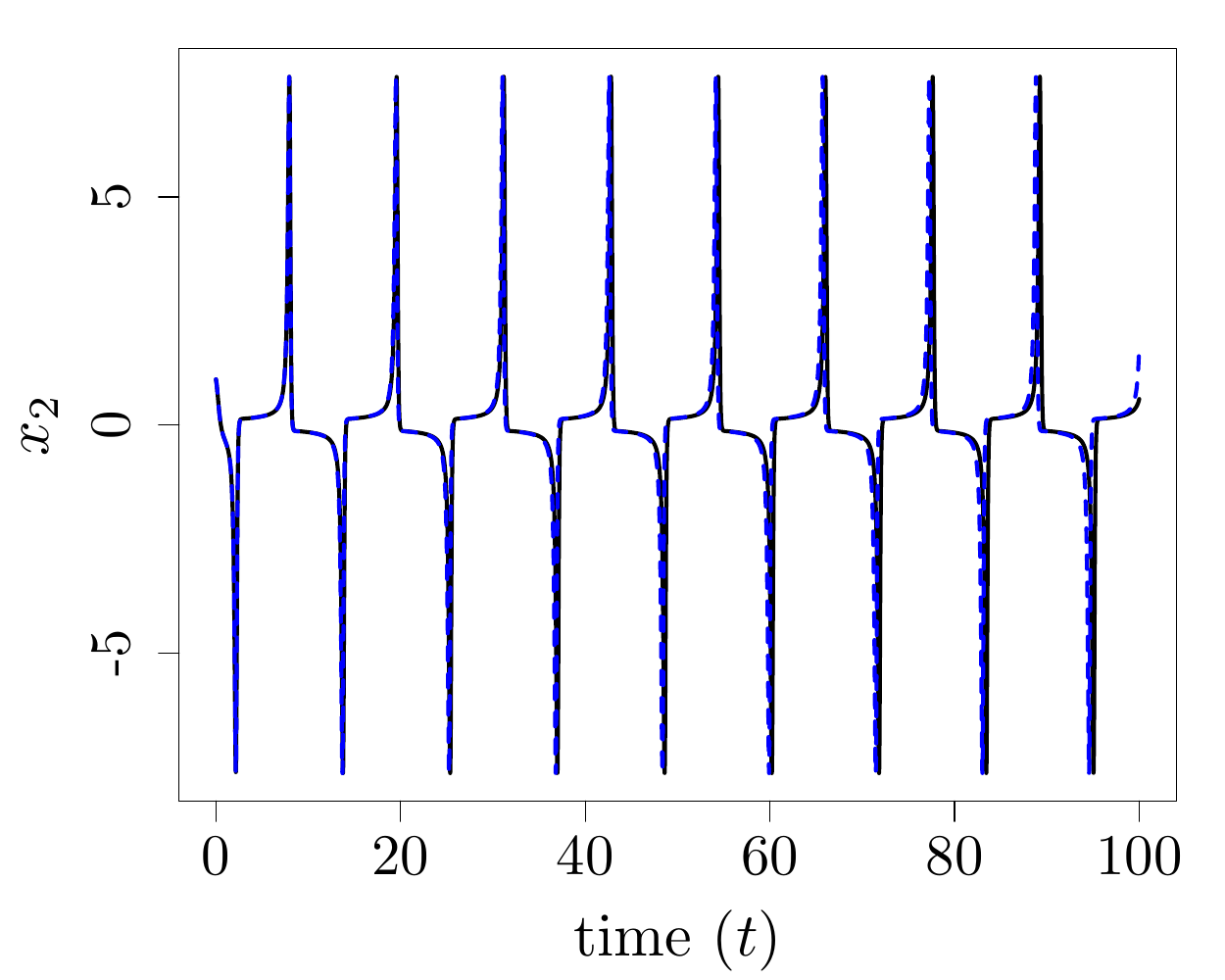}
	\includegraphics[width=0.49\textwidth]{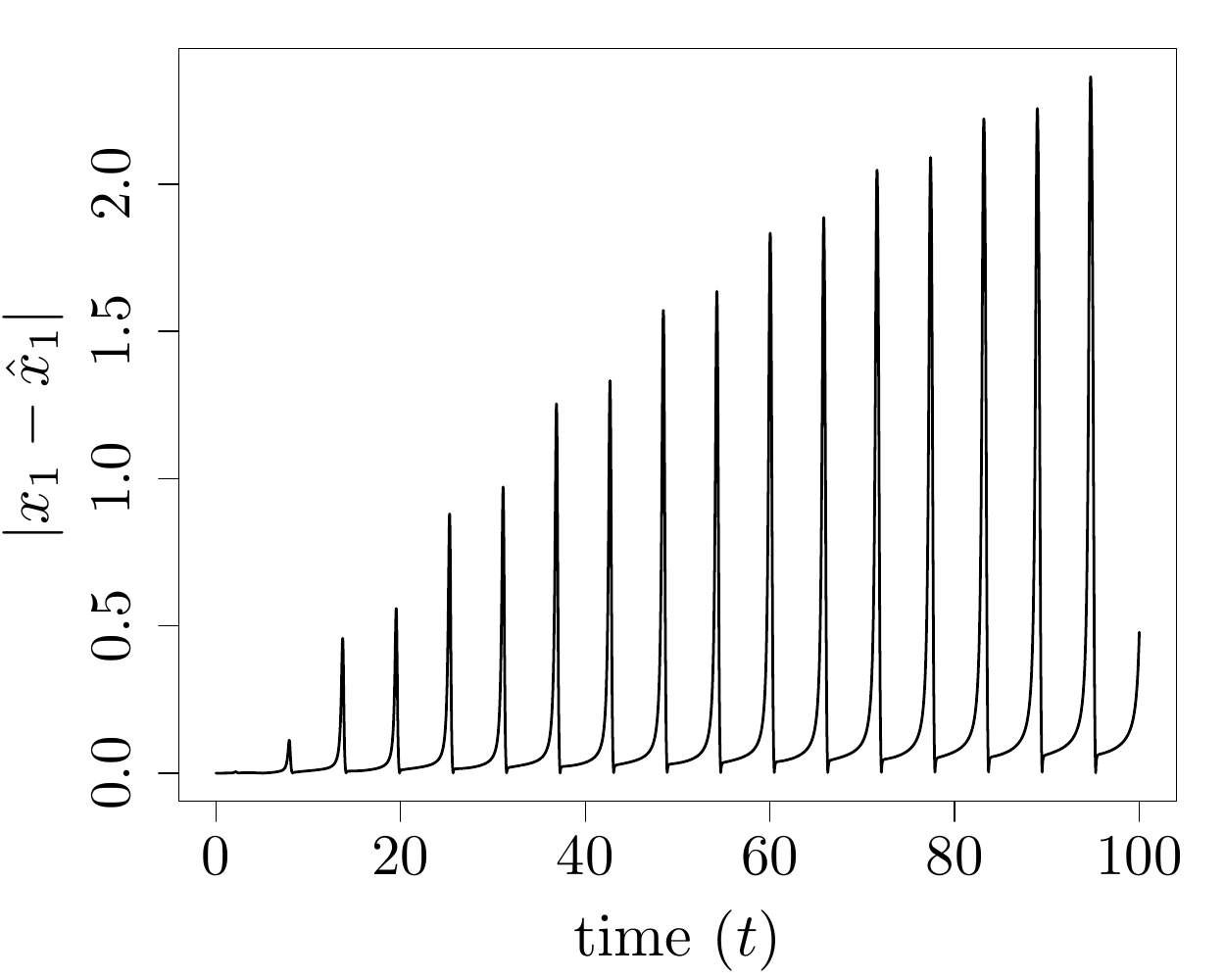}
	\includegraphics[width=0.49\textwidth]{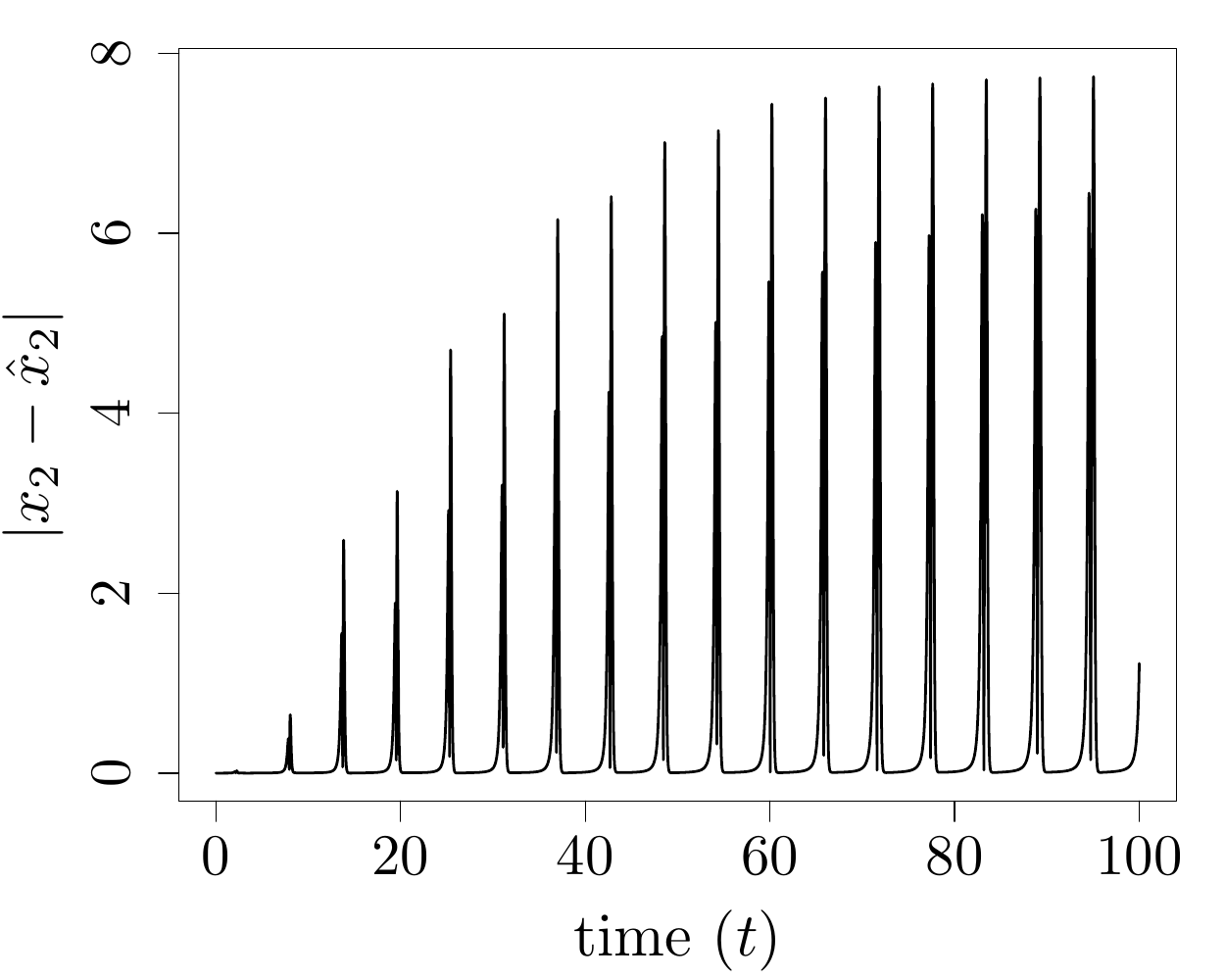}
	\caption{Top: Evolution of state variables of the Van der Pol oscillator (solid black) and their predictions (blue dashed) considering the input uncertainty, but neglecting correlation between emulators, as described in Algorithm \ref{alg1}. Bottom: Difference between the state variables of the Van der Pol system and their predictions, i.e. $\vert x_i - \hat{x}_i \vert$. The prediction capability of emulators is high up to approximately $t = 30$.}
	\label{vander_MC}
\end{figure}

\begin{figure}[htpb] 
	\includegraphics[width=0.49\textwidth]{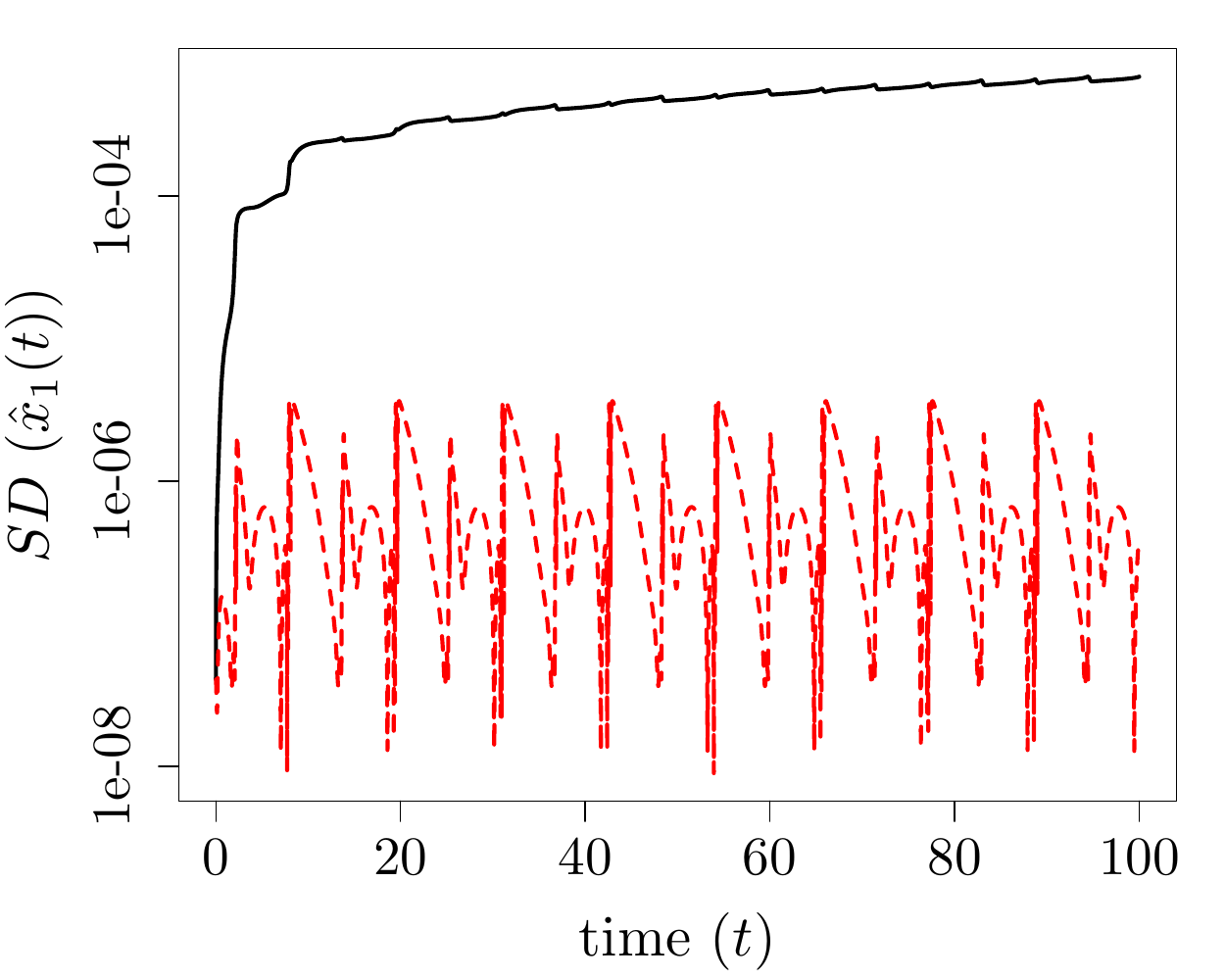}
	\includegraphics[width=0.49\textwidth]{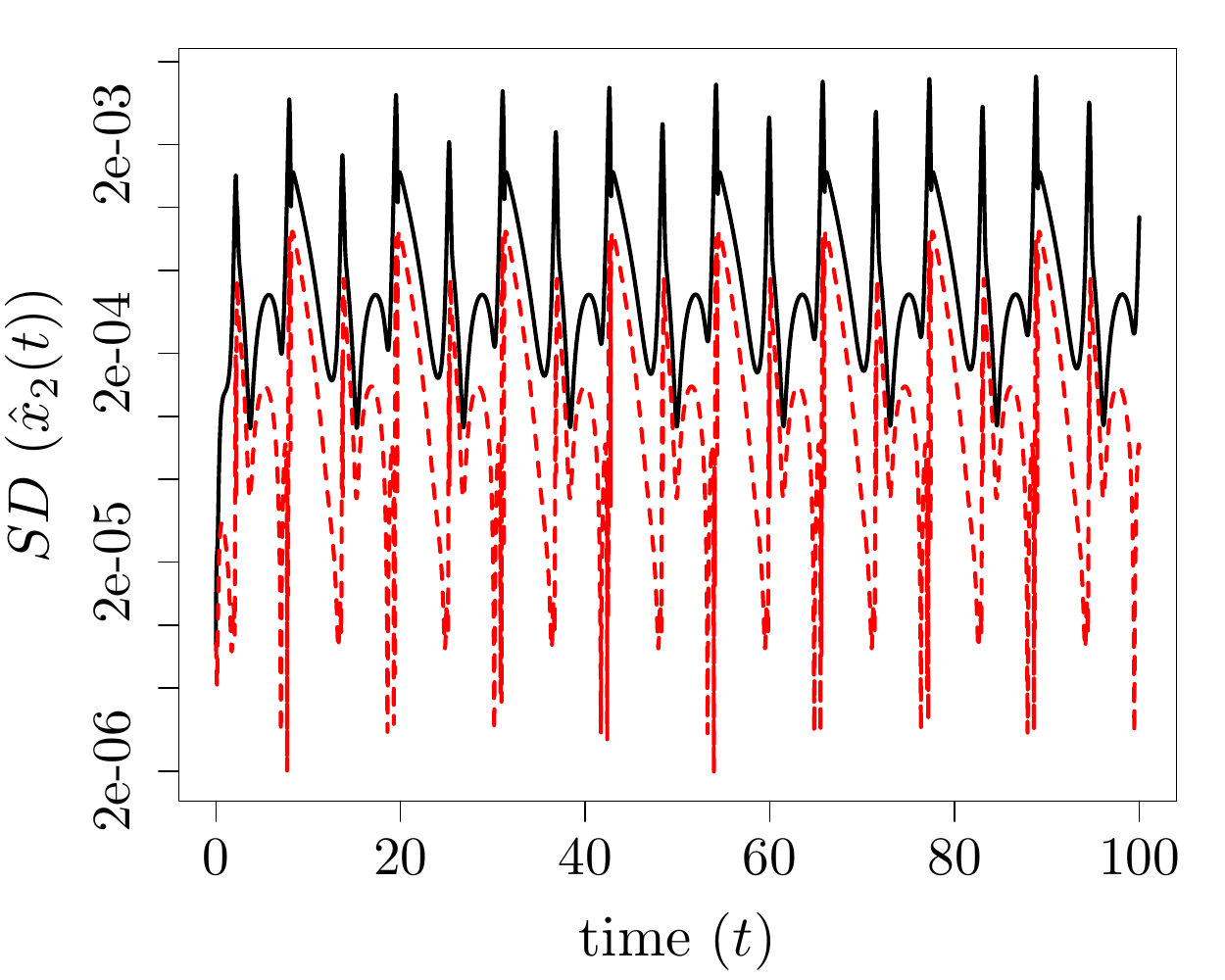}
	\caption{Standard deviation (SD) of predicting the Van der Pol system with  and without considering input uncertainty, solid black and red dashed lines respectively.}
	\label{vander_MC_SD}
\end{figure}

The Van der Pol model was first introduced by the Dutch electrical engineer Balthasar van der Pol in 1920. The Van der Pol oscillator models expresses the behaviour of nonlinear vacuum tube circuits. In its two-dimensional form, it is given by the following equations \cite{strogatz2007}
\begin{equation}
\begin{cases}
\frac{d x_1}{d t} = x_2\\  \frac{d x_2}{d t} = \alpha(1 - x_1^2)x_2 - x_1.
\end{cases}
\label{van_der_pol}
\end{equation}
Here, the scalar $\alpha > 0$ determines the nonlinearity and the strength of damping. Here, we use the initial condition $\bx(t _0) = (x_1 = 1, x_2 = 1)^\top$ and $\alpha = 5$. The accuracy of the emulators is given below.
\begin{center}
\begin{tabular}{l | c r }
	  & $\hat{f}_1$ & $\hat{f}_2 $ \\
	\hline
	$MSE_{LOO}$ & $7.887 \times 10^{-7}$ & $2.580 \times 10^{-7}$
\end{tabular}
\end{center}
The results of predicting state variables of the Van der Pol oscillator, neglecting correlations between emulators is illustrated in Fig. \ref{vander_MC}. The corresponding uncertainties are given by Fig. \ref{vander_MC_SD} where they are compared with the case that the input uncertainty is not considered. The difference between emulation and the true model is low up to approximately $t = 30$.  Again, taking into account the input uncertainty augments prediction uncertainties everywhere. But, since they are small, the true model is not inside the credible intervals (= prediction $\pm$ 2$\times$ prediction standard deviation) when prediction accuracy declines. As is shown in the next section, considering both the input uncertainty and the correlation between emulators allows uncertainties to grow over time. 
\subsection{Application of correlated emulators to the Lorenz and Van der Pol systems}
\label{correlated_emuls_results}
\begin{figure}[htpb] 
	\includegraphics[width=0.49\textwidth]{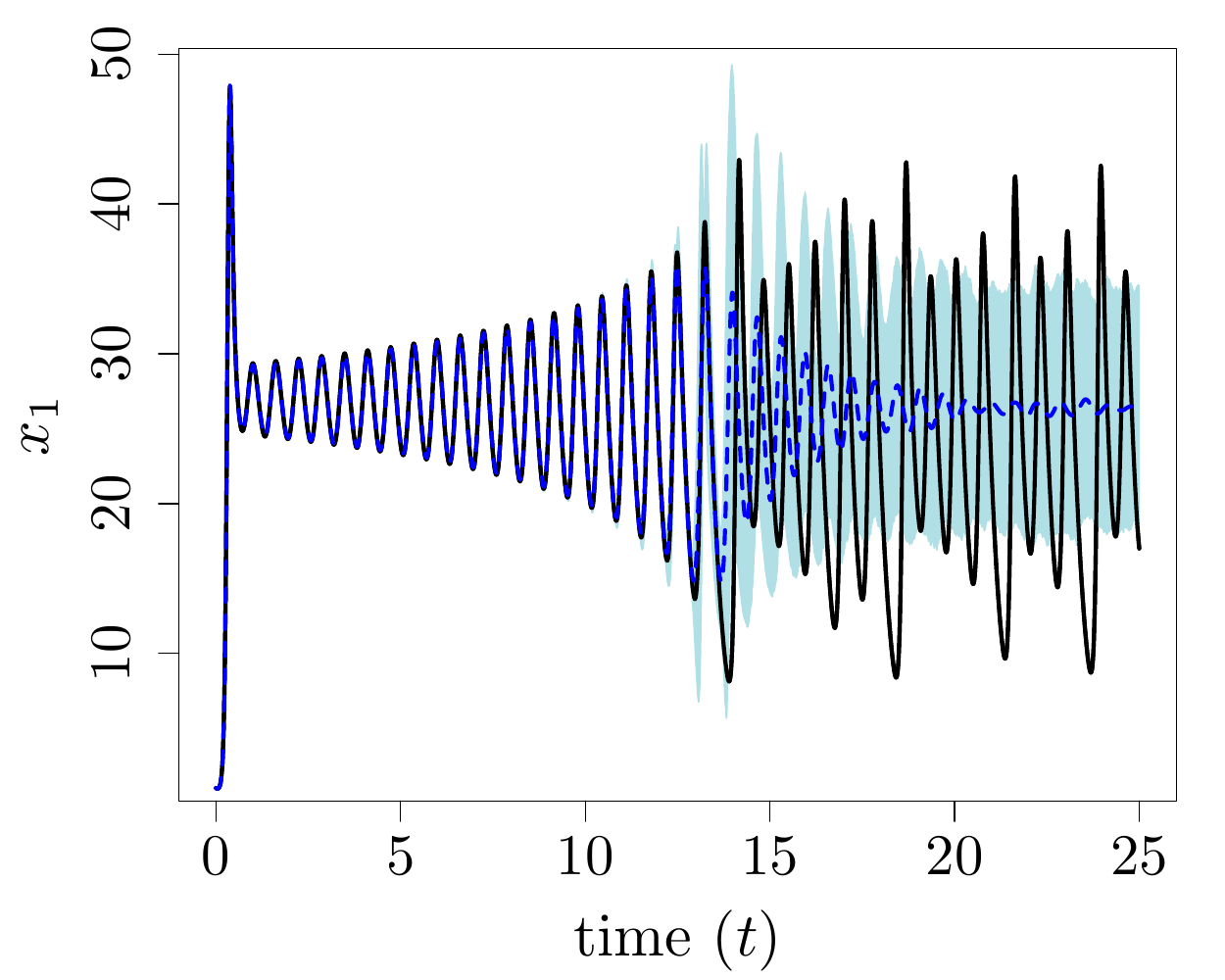}
	\includegraphics[width=0.49\textwidth]{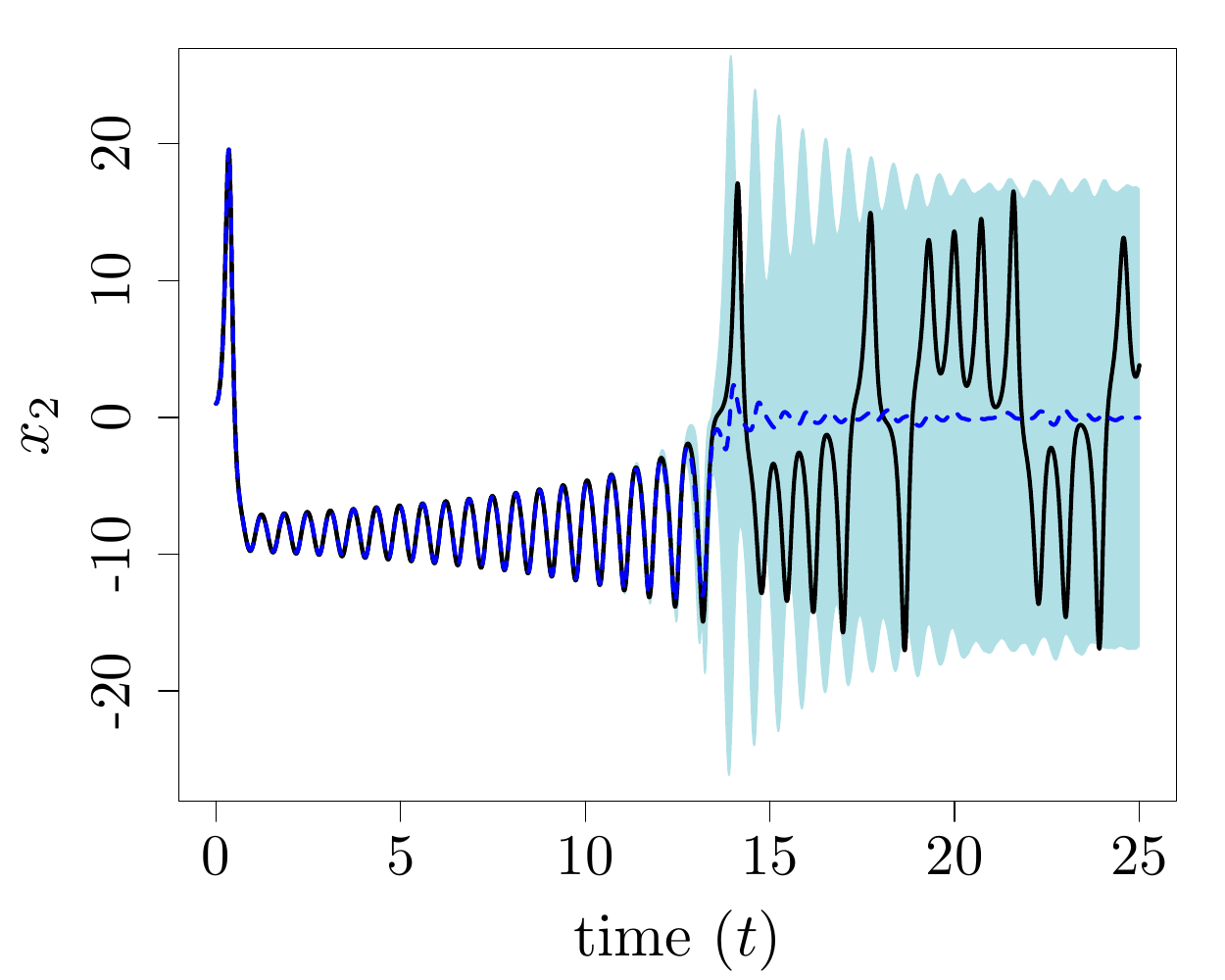}
	\centering
	\includegraphics[width=0.49\textwidth]{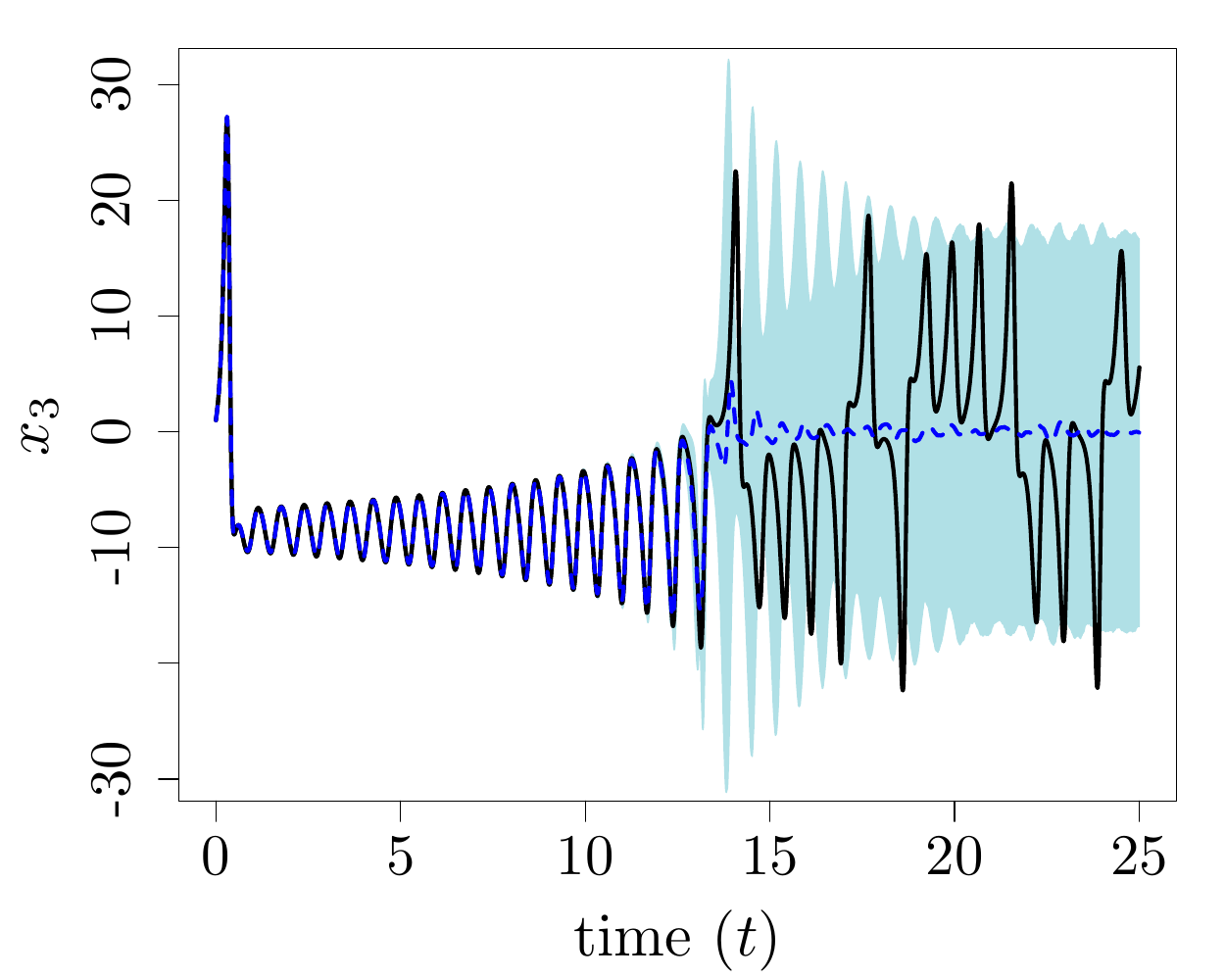}
	\caption{The Lorenz system (solid black) and its prediction (blue dashed) considering input uncertainties and the correlation between emulators. The prediction accuracy is high at the beginning of the time course, say up to $t \approx13$. From this point onwards, i.e. where the emulator is not able to predict the true model well, the prediction tends to the average of the process. However, the credible intervals (shaded) are large enough to contain the true model most of the time. Note when deviation from the true model occurs the prediction uncertainty reaches its maximum.}
	\label{lorenz_method2}
\end{figure}

\begin{figure}[htpb] 
	\includegraphics[width=0.49\textwidth]{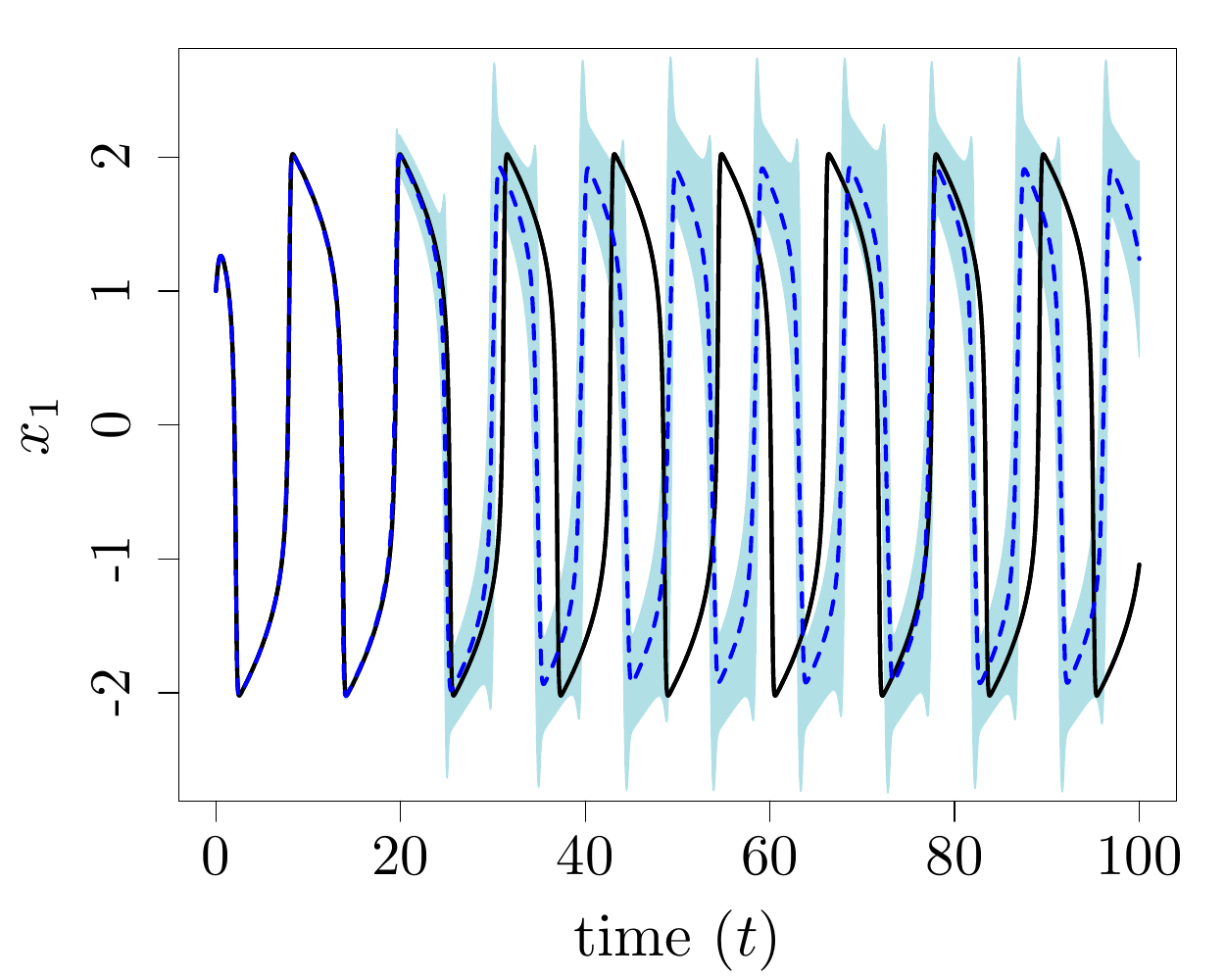}
	\includegraphics[width=0.49\textwidth]{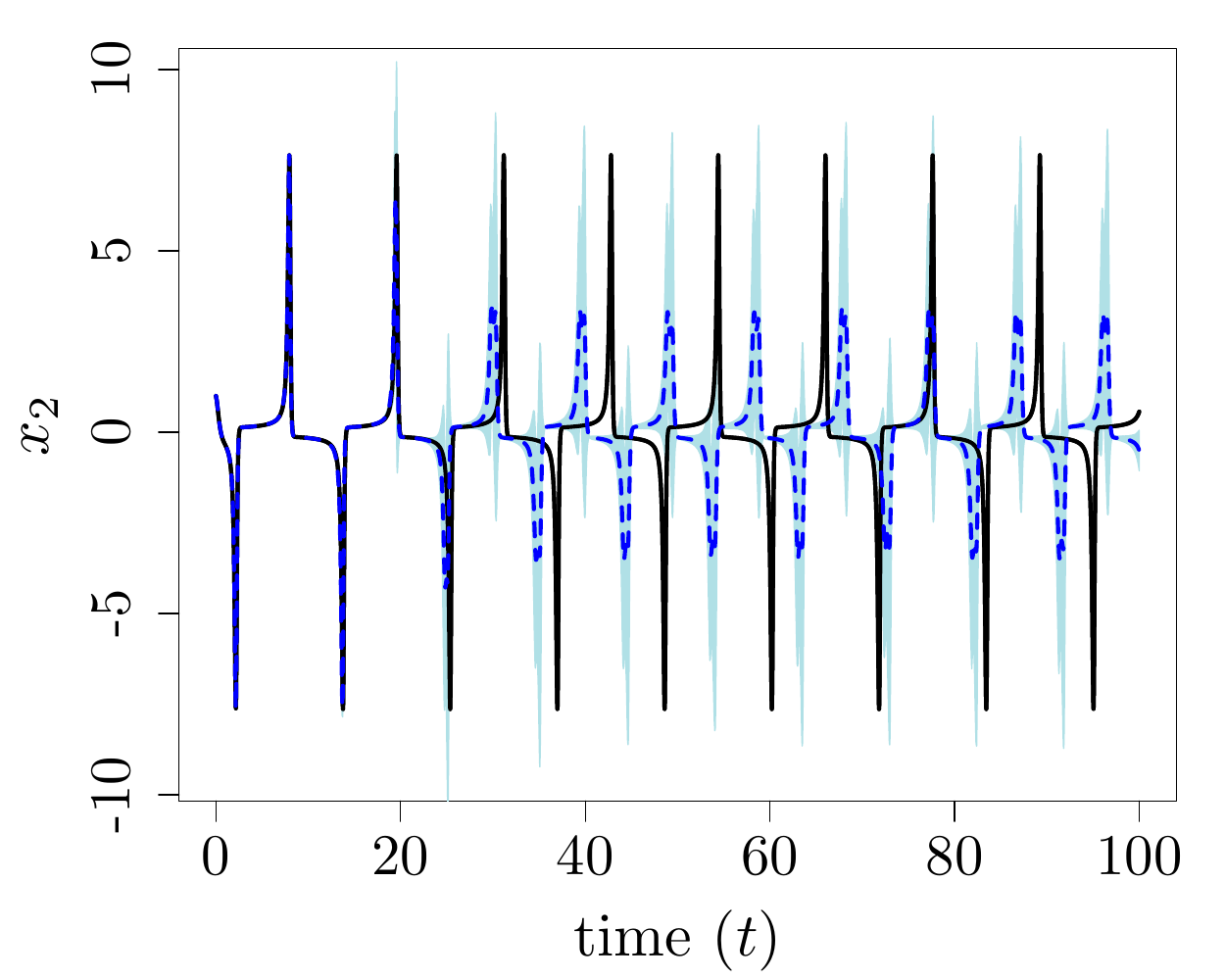}
	\caption{The Van der Pol model (solid black) and its prediction (blue dashed) considering input uncertainties and the correlation between emulators. The shaded area represents the credible intervals. 
		The predictive capability of the emulator is high at the beginning of the time course, up to $t \approx 25$, but subsequently is a frequency miss-match and the prediction damps. From this point onwards, the prediction tends to the long term average of the system. However, the prediction uncertainty blows up when the emulator deviates from the true model.}
	\label{vander_method2}
\end{figure}

The results of emulating the Lorenz and Van der Pol systems considering input uncertainties together with the correlation between emulators are illustrated in Figs. \ref{lorenz_method2} and \ref{vander_method2}, respectively. The predictive capability of these emulators is high at the beginning of the time course, say up to $t \approx 13$ for the Lorenz  and $t \approx 25$ for the Van der Pol models. In both cases, when the emulators deviate from the true models, the prediction uncertainties blow up which can be used to identify the time for which the prediction obtained by the emulator is reliable. This will be discussed later.

In the emulation of the Lorenz model, the prediction (blue dashed lines) tends to the average of the system after the emulator no longer predicts the true model well, i.e. $t \approx 13$. However, the uncertainty is large enough to encompass most values of the system such that the true model predominantly remains inside the credible intervals represented by the shaded area. Recall that the main drawback of the uncorrelated emulators method is that the prediction uncertainty is too small and the true model is not inside the credible intervals.

In the case of the Van der Pol system, the prediction is accurate up to $t \approx 25$. From this time onwards, a frequency miss-match happens and deviation from the true trajectory grows such that the simulator output is rarely inside the credible intervals after $t \approx 40$. Also, the amplitude of the prediction gradually damps which can be interpreted as the emulator in effect ``giving up" on trying to emulate the value at a particular time $t$ and instead falling back on a prediction for a random time. Such a prediction is useless in practice but statistically makes sense.

\begin{figure}[htpb] 
	\includegraphics[width=0.49\textwidth]{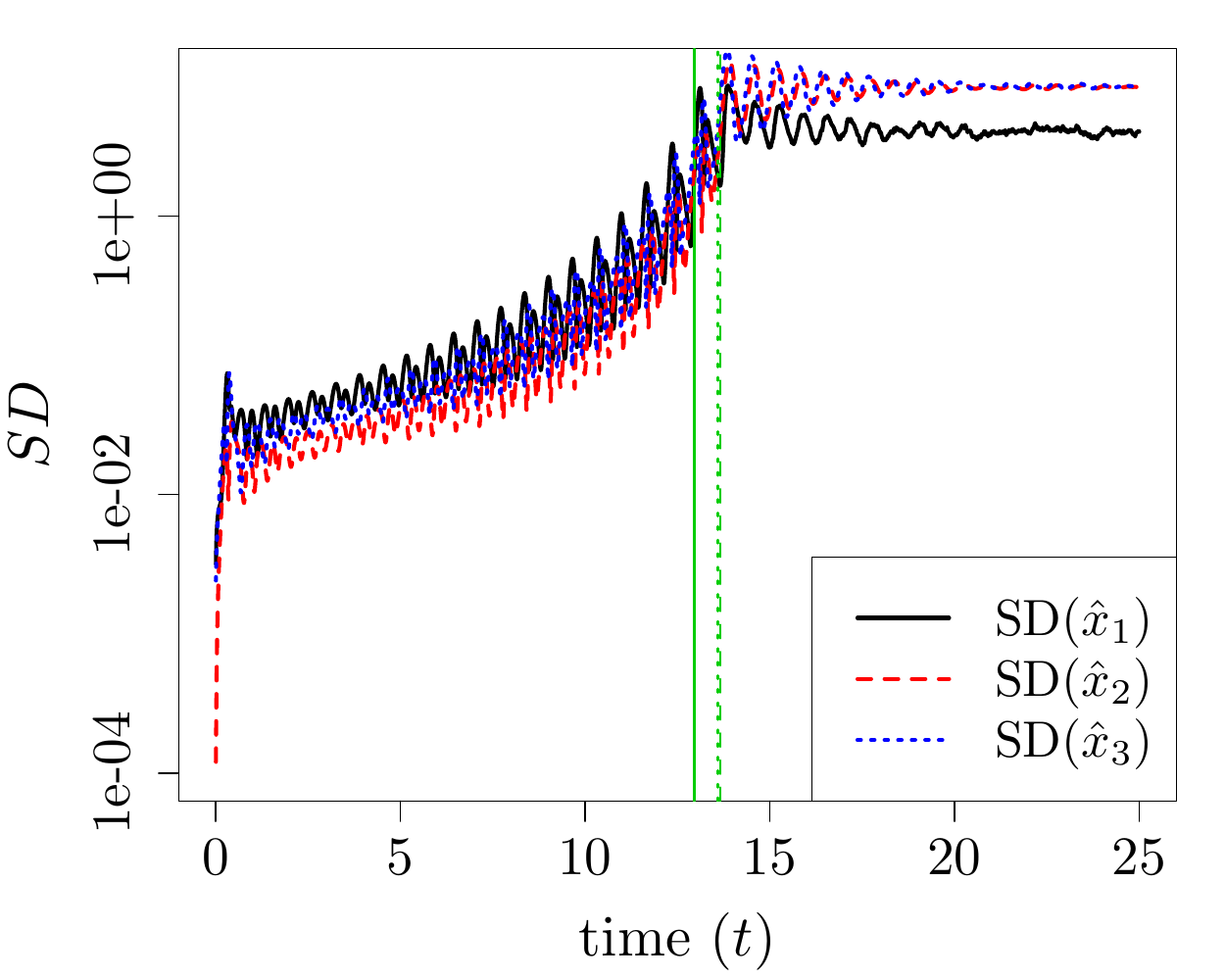}
	\includegraphics[width=0.49\textwidth]{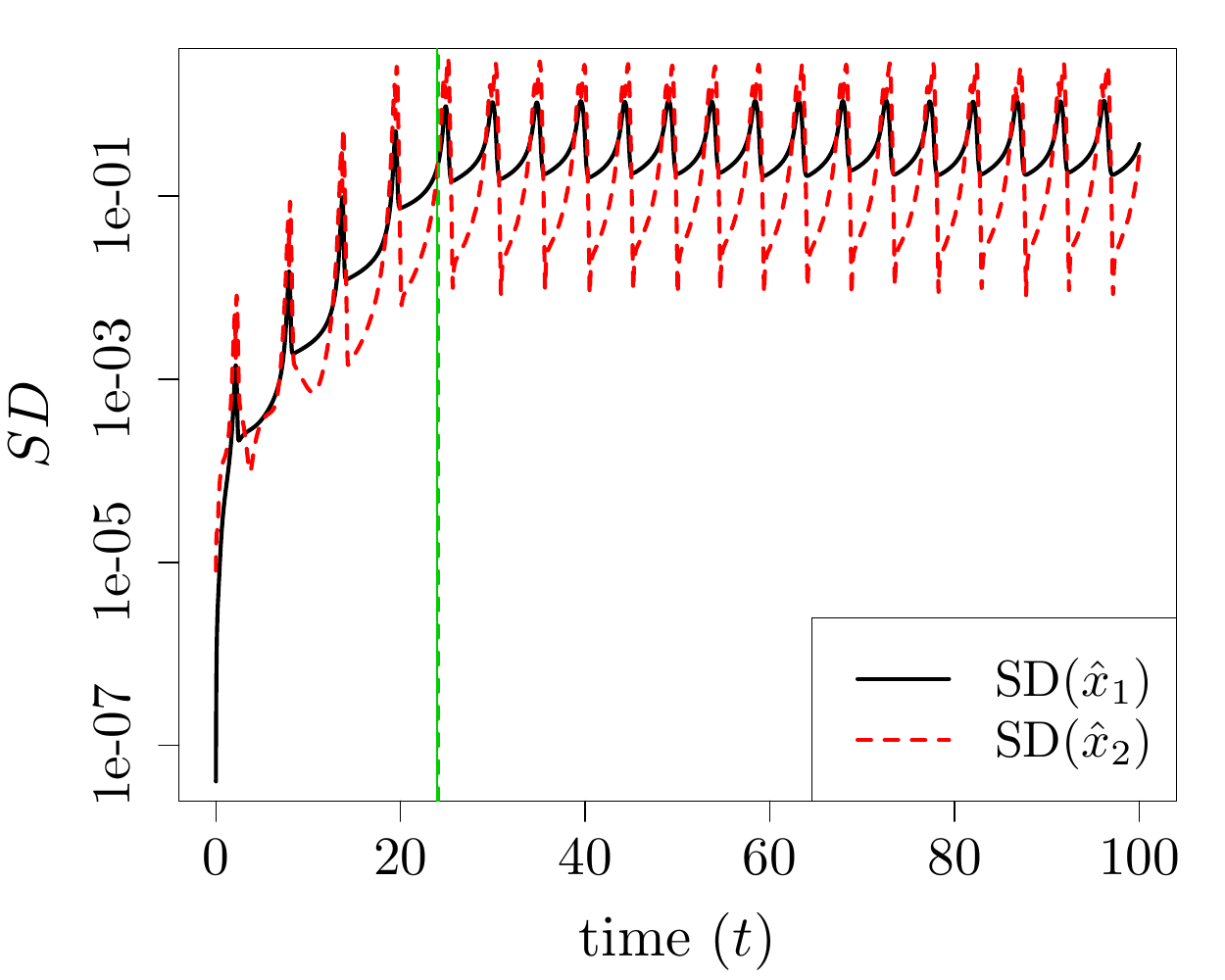}
	\caption{Left: standard deviation (SD) of prediction associated with the three variables in the Lorenz model. Right: SD of prediction associated with the two variables in the Van der Pol model. The $y$-axis is on logarithmic scale. In both cases, an SD reaches its maximum where the corresponding emulator can not well predict the true model. This can be used as a criterion to detect the predictability horizon of dynamic emulators. The vertical green lines show the predictability horizon in the corresponding time series, change point in $\hat{x}_1(t)$ (solid),$\hat{x}_2(t)$ (dashed) and $\hat{x}_2(t)$ (dotted).}
	\label{SD_method2}
\end{figure}

Fig. \ref{SD_method2} shows uncertainties associated with predictions of the variables in the Lorenz (left) and Van der Pol systems (right) considering the correlation between the emulators through the time series. As can be seen, the uncertainty grows and reaches its maximum, interestingly, when deviation from the true model begins. From this point onwards, the expected value of the emulator is the long term average of the underlying model while the uncertainty of prediction is large. This point can be used as a measure for the \emph{predictability horizon} of dynamic emulators. More precisely, the time at which the mean value of uncertainty changes significantly is considered as the predictability horizon. To identify the \emph{change point} in the mean of time series, the \texttt{cpt.mean} function implemented in the \texttt{changepoint  R} package \cite{killick2014} is applied to the prediction uncertainties. The vertical green lines in Fig. \ref{SD_method2} represent the change points in the time series. 
Similar results are obtained when emulating the Lorenz and Van der Pol systems with different initial conditions using the correlated method. In Appendix \ref{append1}, the emulation of the two systems is demonstrated for additional initial conditions.

\begin{figure}[htpb] 
	\includegraphics[width=0.49\textwidth]{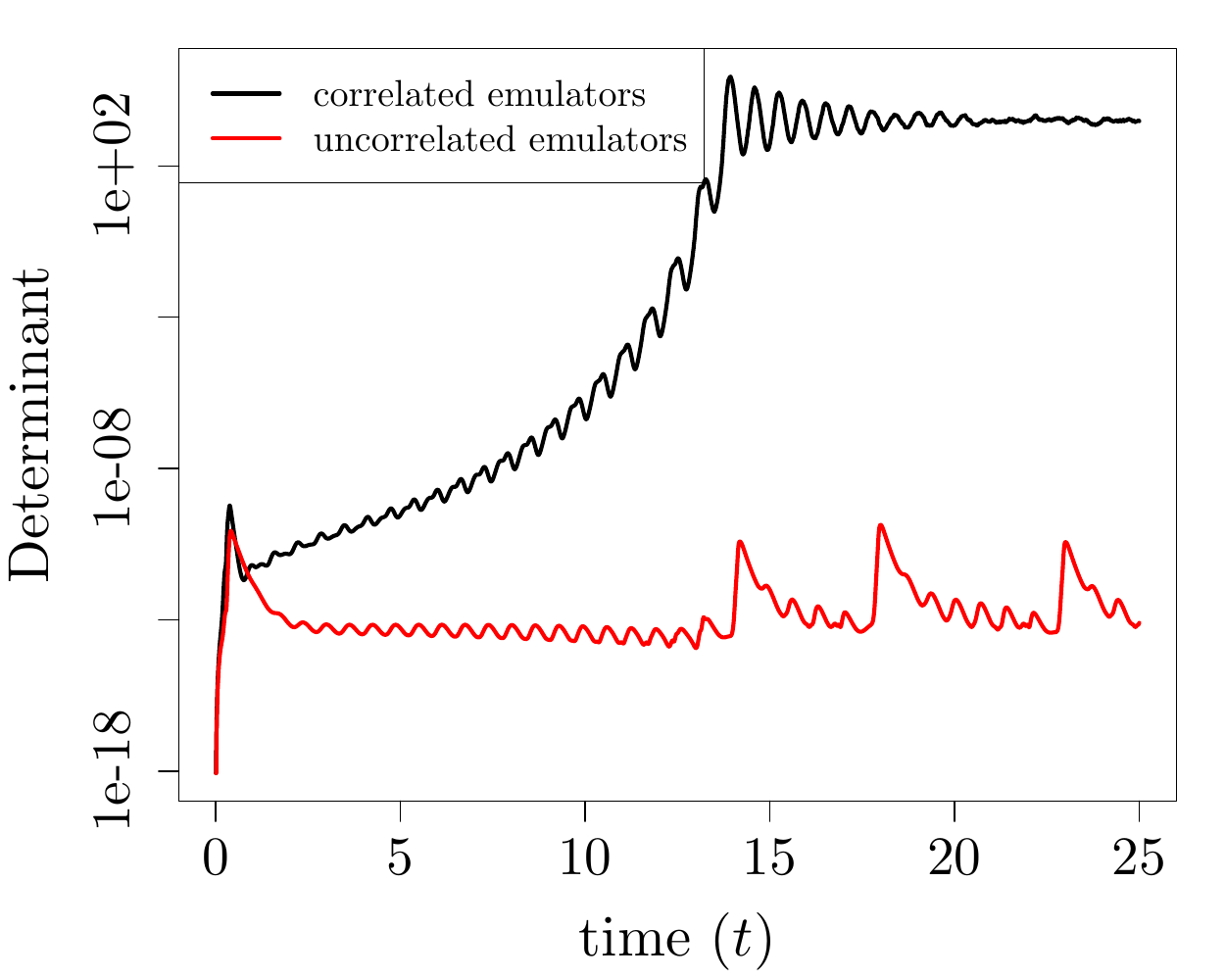}
	\includegraphics[width=0.49\textwidth]{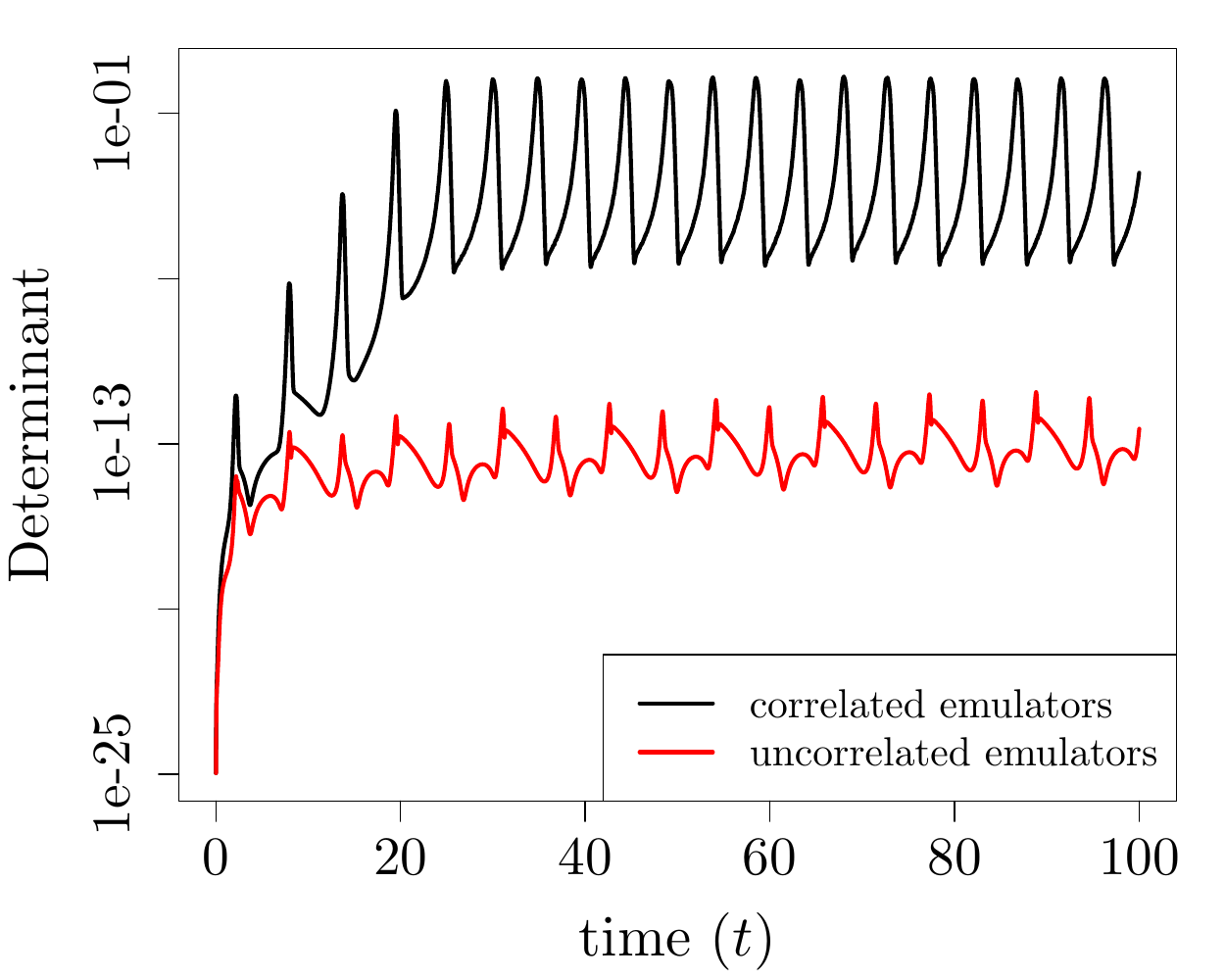} 
	\caption{The determinant of the covariance matrix, used in Table \ref{tab1}, over time in correlated (black) and uncorrelated (red) emulators in the Lorenz (left) and Van der Pol (right) systems. In both examples, the determinant is larger in the correlated method which lead to a wider sampling of update directions for the next step. This is in line with the magnitude of uncertainties in the correlated and uncorrelated methods.}
	\label{Determinant}
\end{figure}

Using correlated emulators, the magnitude of uncertainty is generally larger than for the uncorrelated case, which would lead to a wider sampling of update directions for the next step, and hence greater deviation from the true underlying trajectory. This is especially the case of the Van der Pol model where the prediction accuracy is higher in the uncorrelated method, see Figs. \ref{vander_method2} and \ref{vander_MC}. However, in this approach the uncertainty is too small and the true model is not inside the credible intervals. The larger uncertainty in the correlated method can be justified by looking at the determinant of the covariance matrix (Fig. \ref{Determinant}) used in Table \ref{tab1}  to generate samples from the input distribution. As can be seen, the determinant of the covariance matrix is consistent with the prediction uncertainties. Note that the determinant measures overall dispersion of a multidimensional random variable and is referred to as generalised variance.
\section{Conclusion}
In this paper we develop a general framework for dynamically emulating highly non-linear functions with time series outputs using Gaussian processes. Such functions show the behaviour of phenomena evolving with time. One advantage of our method is that it is easy to implement in comparison to alternative methods; it uses a GP emulator to perform one-step ahead predictions in an iterative way over the whole time series. Moreover, we propose a number of ways to propagate uncertainty through the time series based on both the uncertainty of inputs to the emulators and the correlation between them. The capability of our method is illustrated in application to two non-linear dynamical systems: the Lorenz and Van der Pol systems. In both examples, the prediction uncertainty obtained by the proposed method (i.e. correlated emulators) allows to measure a ``predictability horizon", within which the prediction accuracy is high. It should be noted that in our two examples the computer model run time is small and we can compare the model and emulator directly. This is not the case for many applications where the model run time is large.  

One can extend the idea of one-step to several-step ahead predictions. The main consideration is that the flow map becomes more non-linear if the number of steps increases. As a result, more training data is required to approximate the flow map well which increases the computational complexity of GPs. Recall that a GP has a computational complexity of $\mathcal{O}(n^3)$. With respect to the approximation of the distribution of inputs at subsequent time steps, as mentioned before, Eq. (\ref{integral}) can be approximated by the deterministic methods such as the Laplace's approximation. Another possible future research direction is to investigate such techniques in the framework of dynamic emulation. Fig. \ref{Laplace_Lorenz_fig} (Appendix \ref{append2}) illustrates the results of emulating the Lorenz system in which, at each iteration, the output distribution is approximated by the Laplace's method. In this case, the prediction uncertainty does not grow over time, similarly to the case of uncorrelated emulators.
\section*{Acknowledgements}
The authors gratefully acknowledge the financial support of the EPSRC via grant EP/N014391/1. The contribution of MG was generously supported by a Wellcome Trust Institutional Strategic Support Award (WT105618MA). MG further acknowledges support from the EPSRC [grant number EP/P021417/1]. We warmly thank Peter Ashwin and Jennifer Creaser for the constructive discussions.

\begin{appendices}
\section{Correlated emulators with different initial conditions}
\label{append1}
The Lorenz and Van der Pol systems are emulated (based on the correlated method) with six more different initial conditions selected randomly from $[-10, 10]^d$. These results are consistent with the previous observations in that the prediction uncertainty can be used to obtain the predictability horizon using the change point detection.
\begin{figure}[htpb] 
	\includegraphics[width=0.32\textwidth]{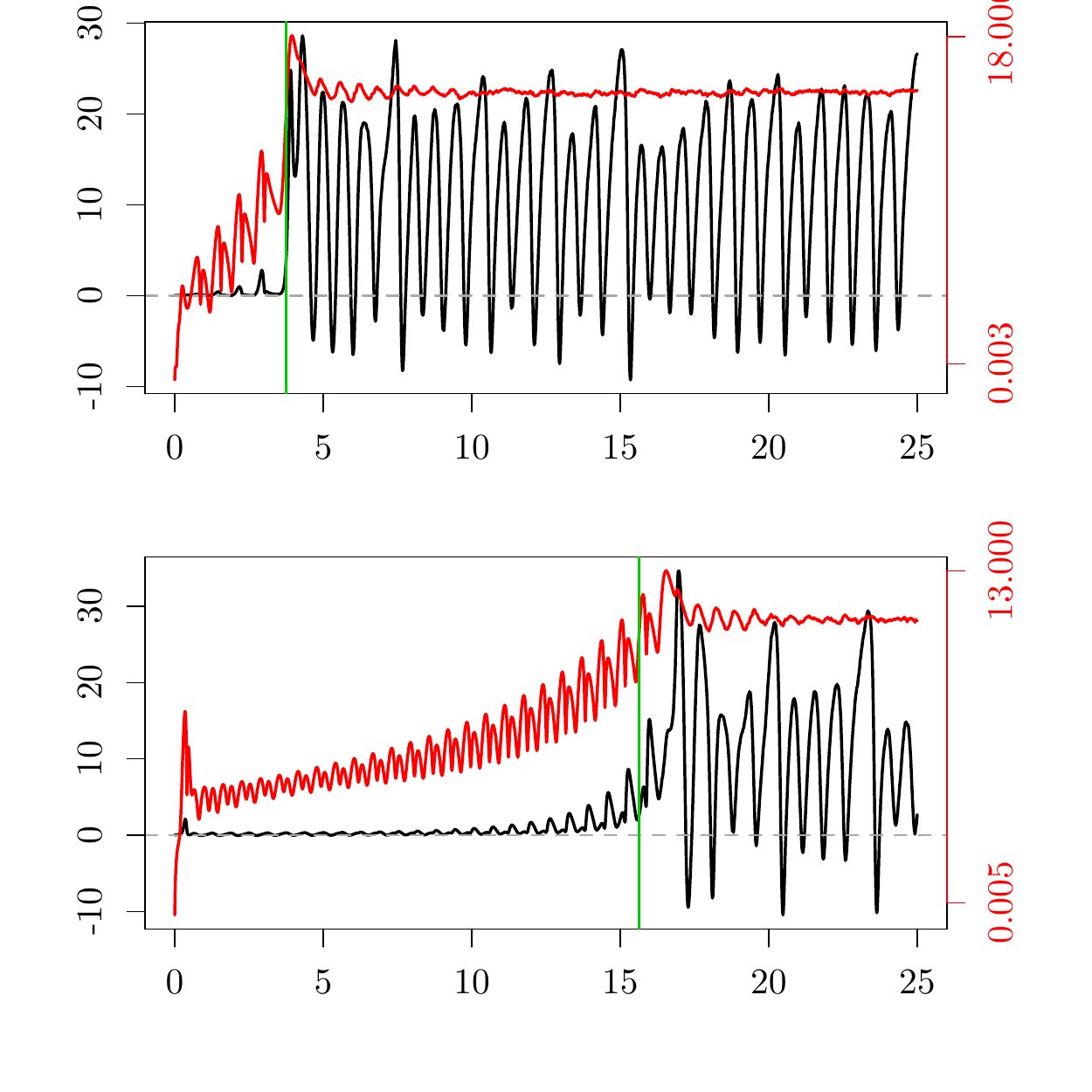}
	\includegraphics[width=0.32\textwidth]{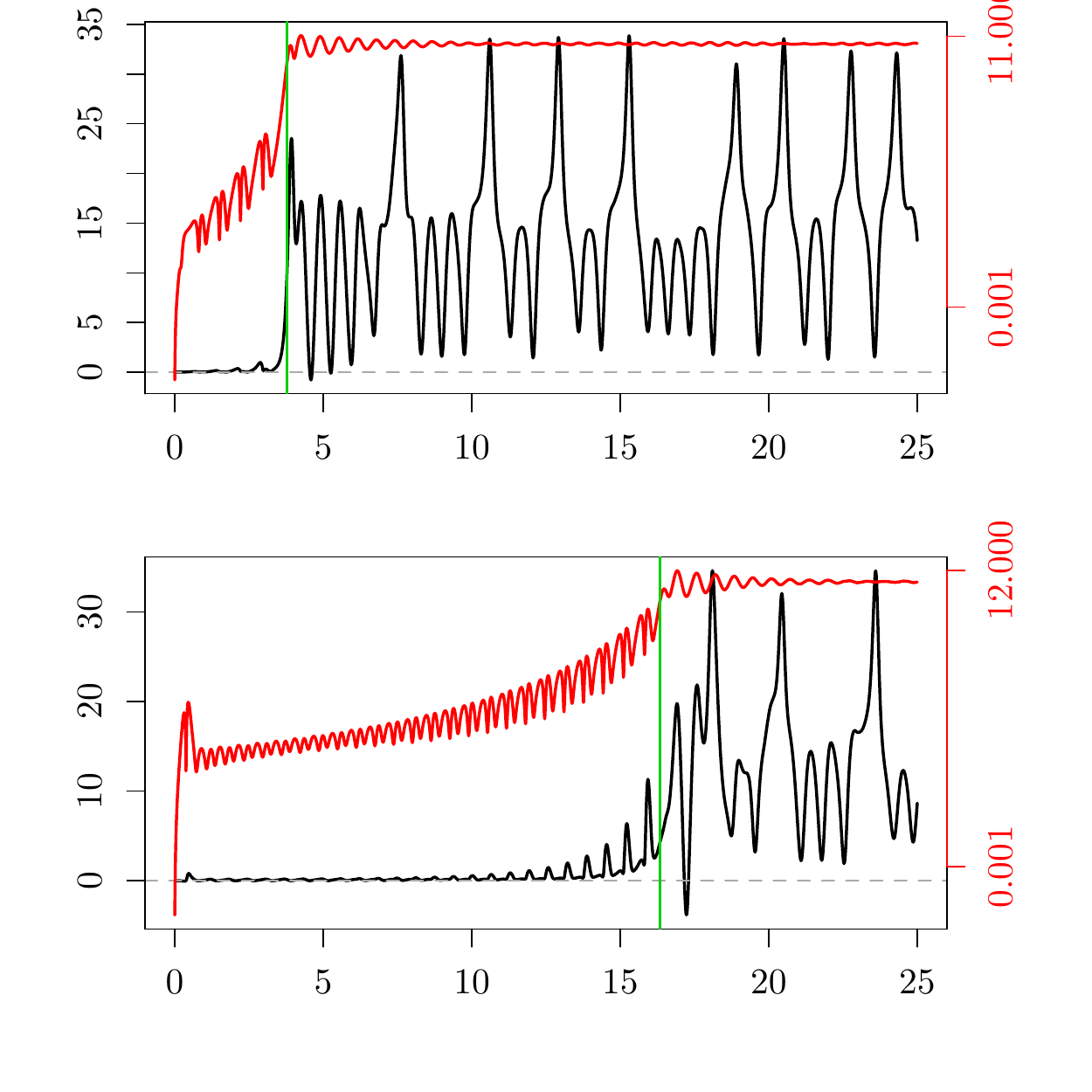}
	\includegraphics[width=0.32\textwidth]{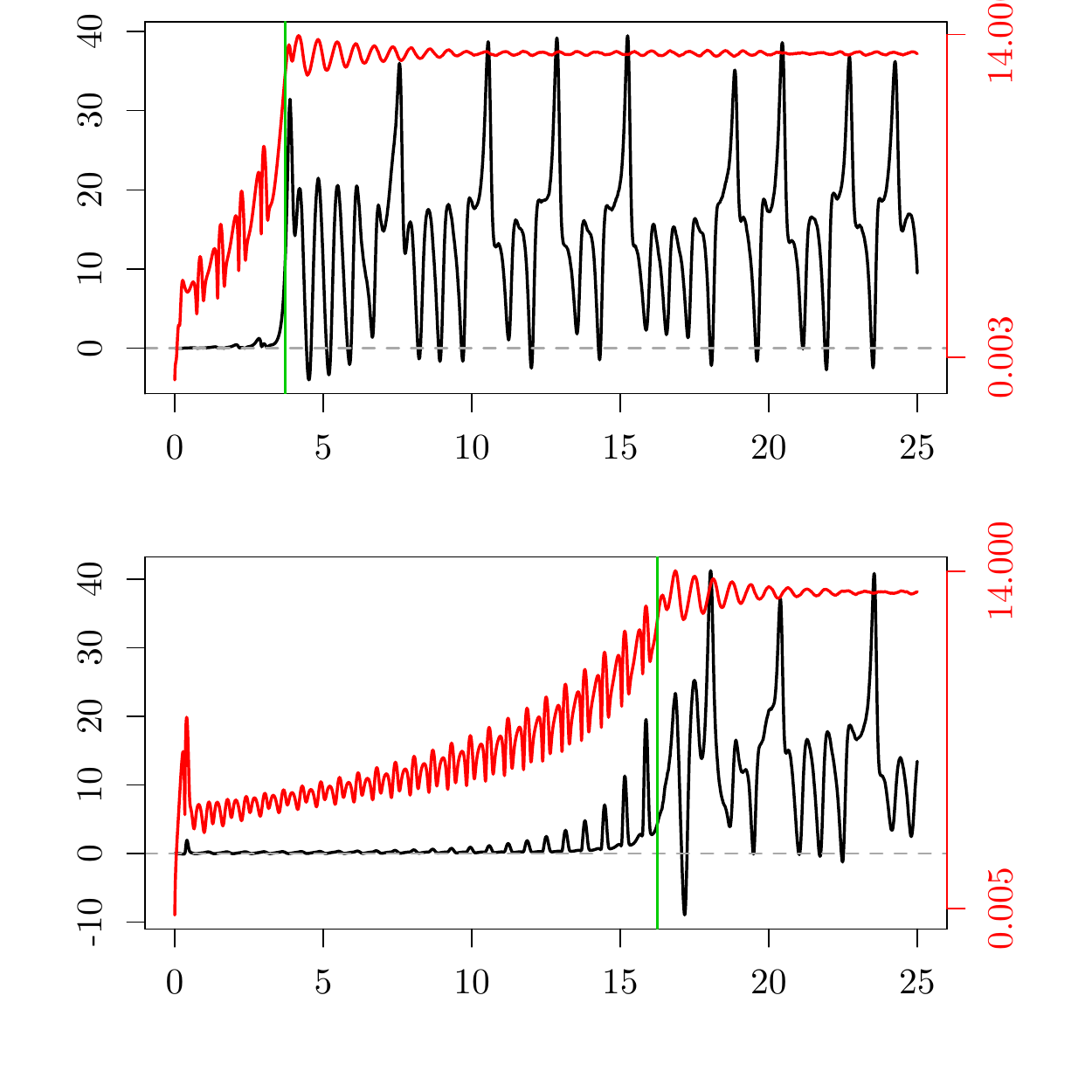} \\
	\includegraphics[width=0.32\textwidth]{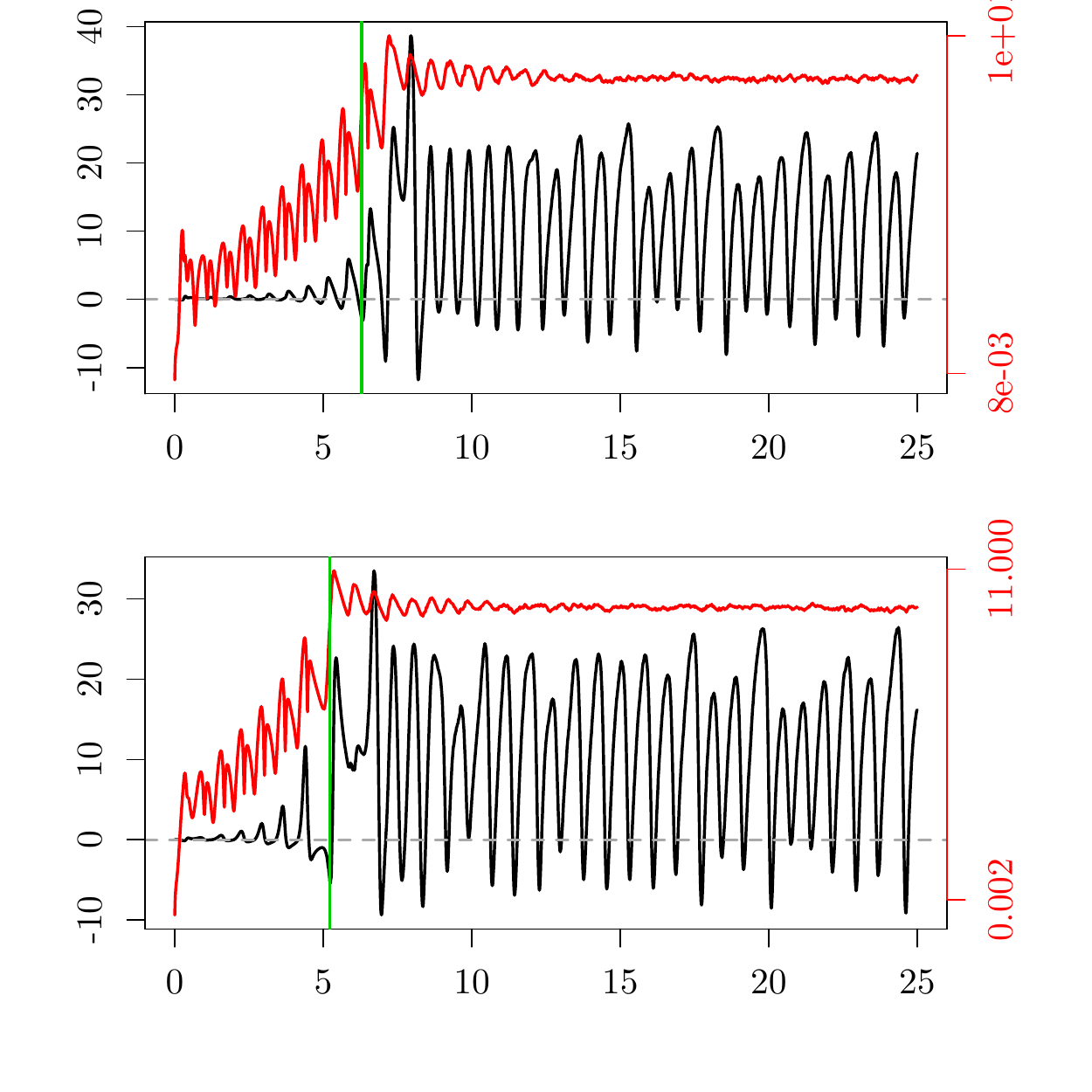}
	\includegraphics[width=0.32\textwidth]{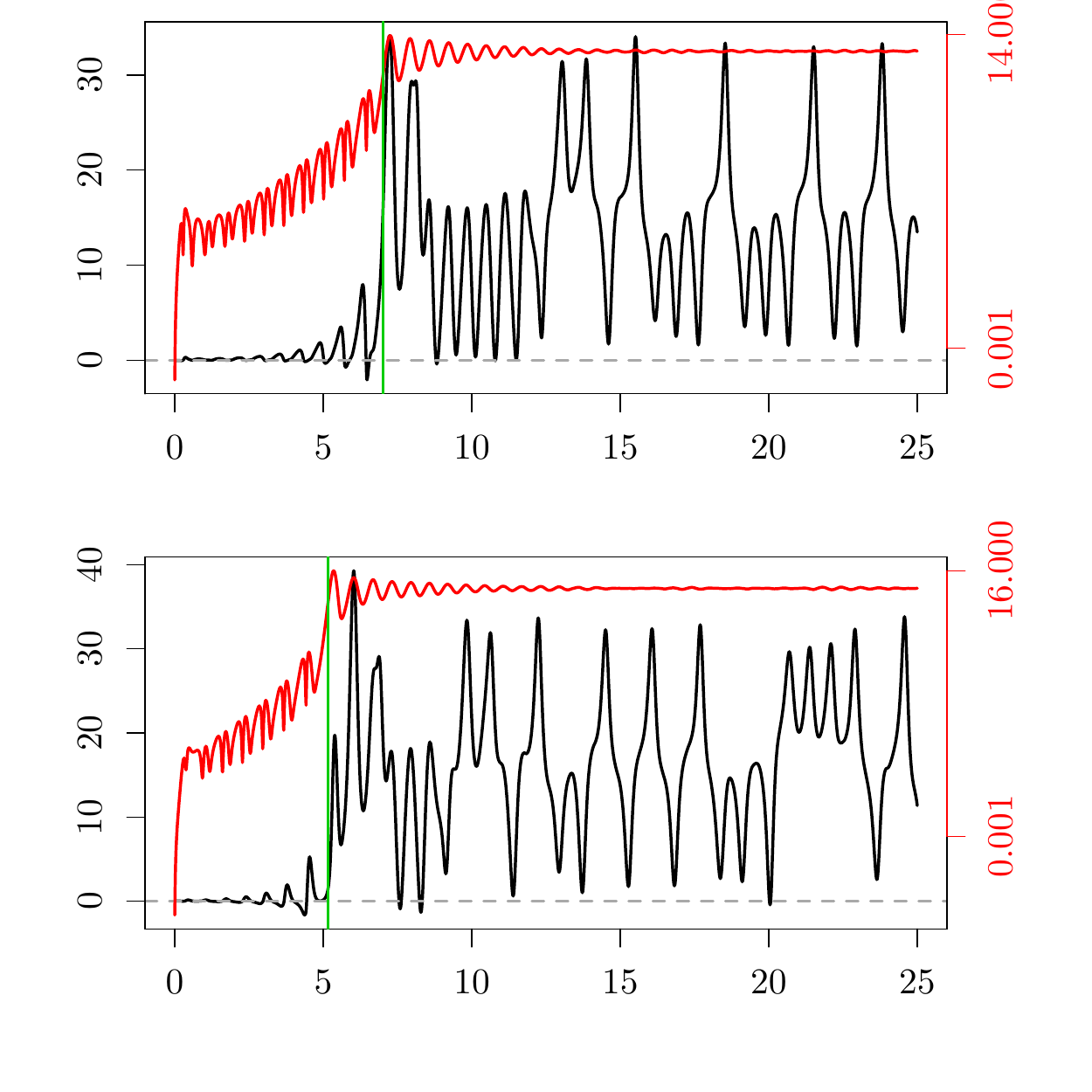} 
	\includegraphics[width=0.32\textwidth]{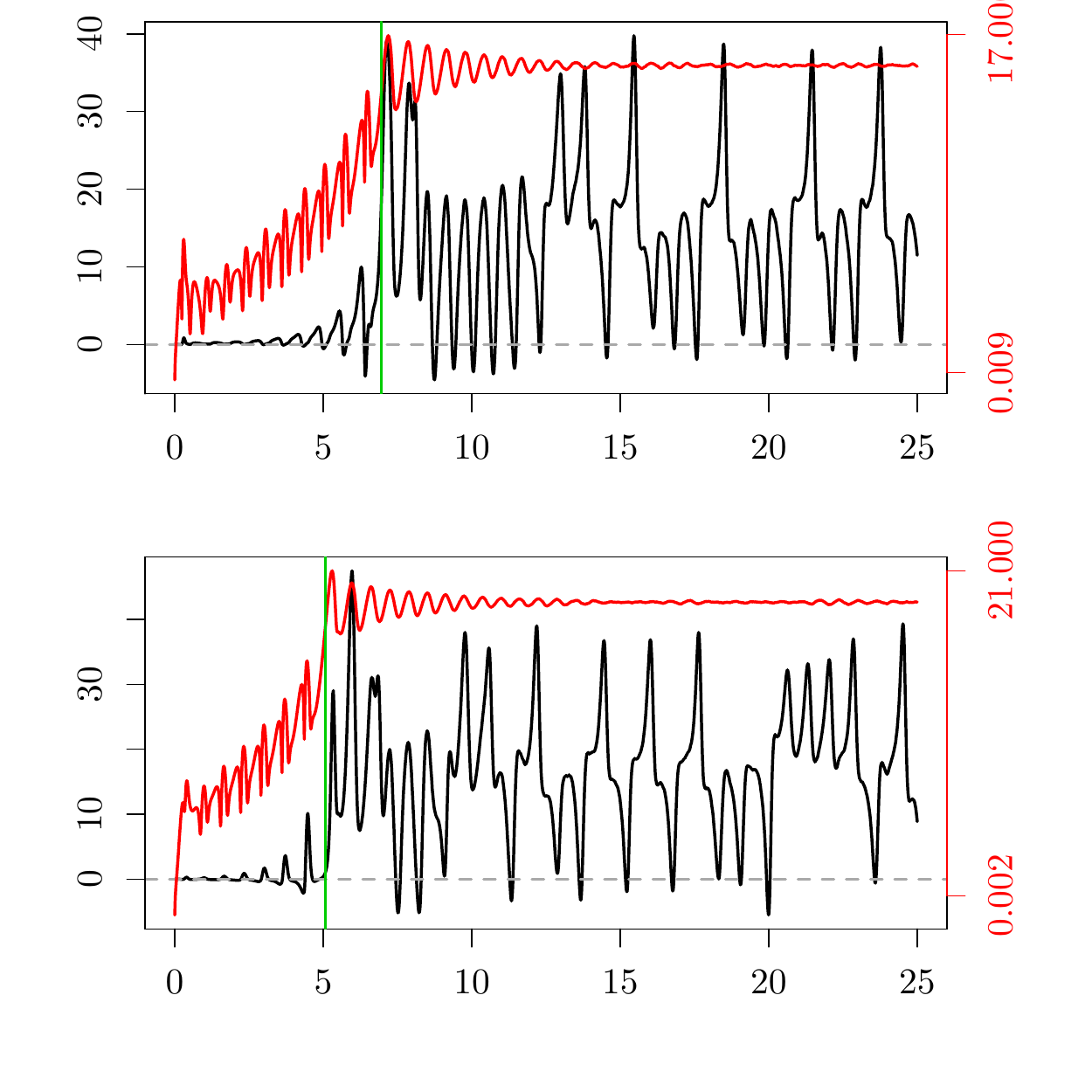} \\
	\includegraphics[width=0.32\textwidth]{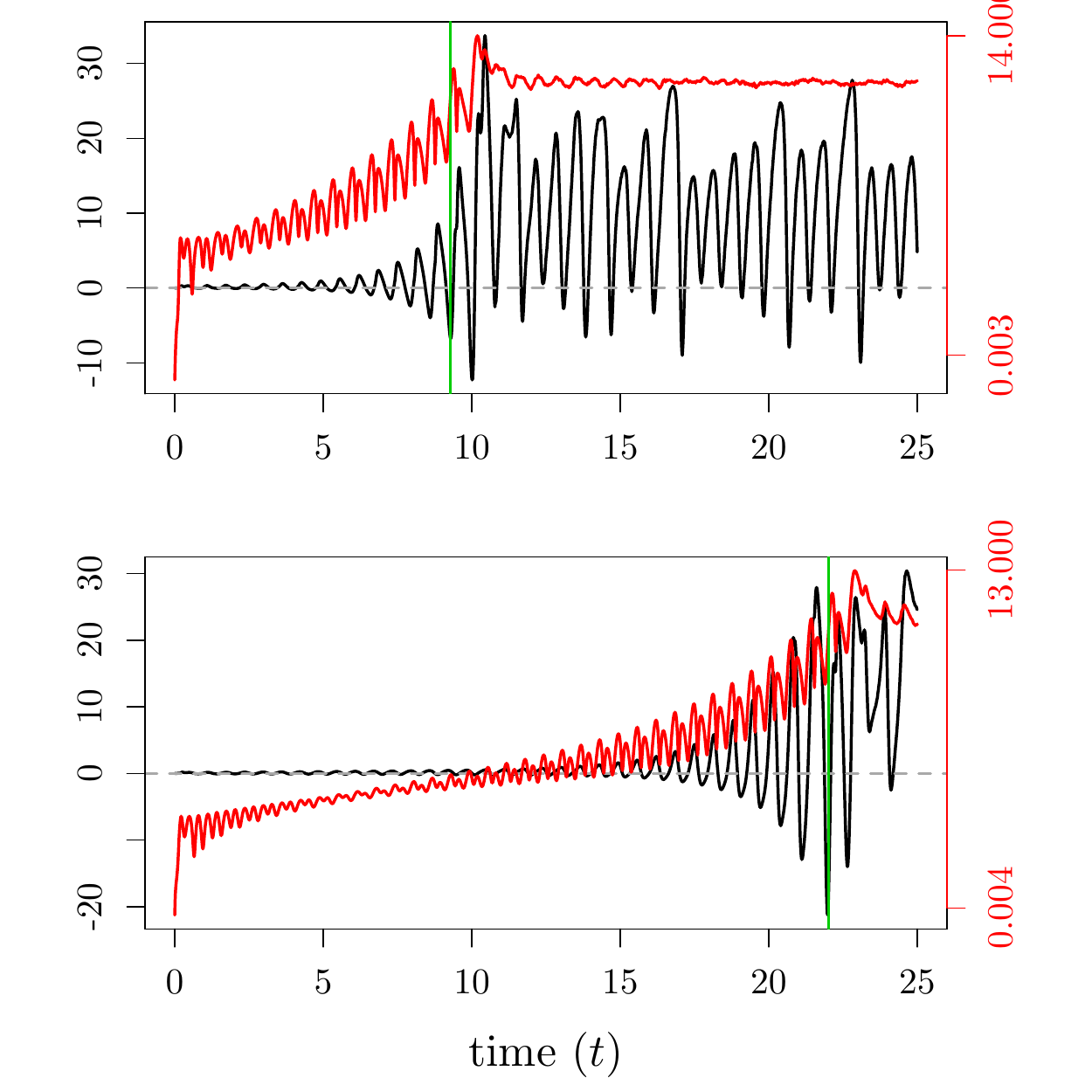}
	\includegraphics[width=0.32\textwidth]{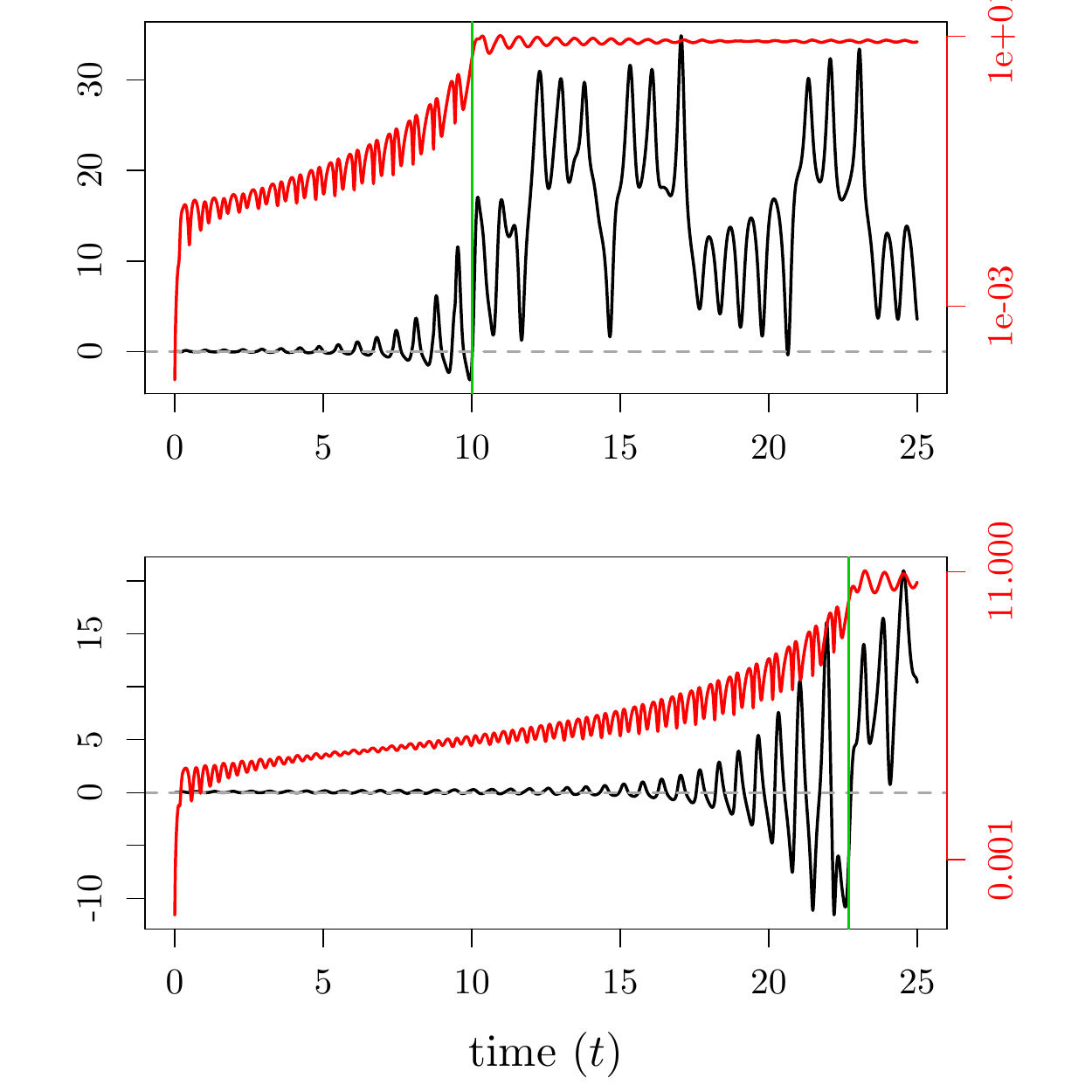} 
	\includegraphics[width=0.32\textwidth]{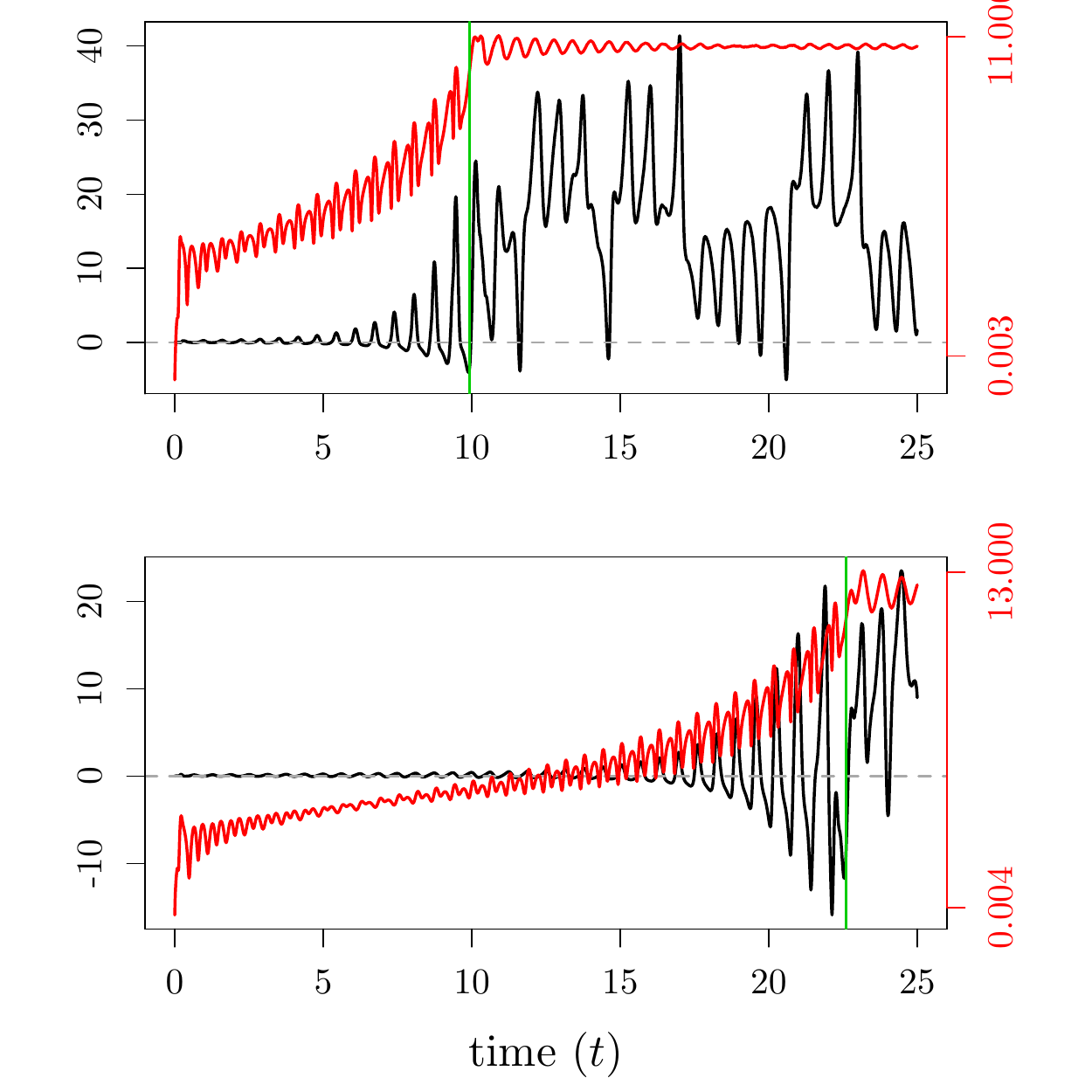} 
	\caption{Prediction of the Lorenz system ($x_1(t)$: first panel, $x_2(t)$: second panel and $x_3(t)$: third panel) with six different initial conditions (corresponding to each row). The black line represents: $\hat{x}_i(t) + 2SD\left(\hat{x}_i(t)\right) - x_i(t)$ and the red line is $SD \left(\hat{x}_i(t) \right)$ with the (log scale) $y$-axis shown in red on the right side. The change point is shown by the vertical green line.}
\end{figure}
\begin{figure}[htpb] 
	\includegraphics[width=0.49\textwidth]{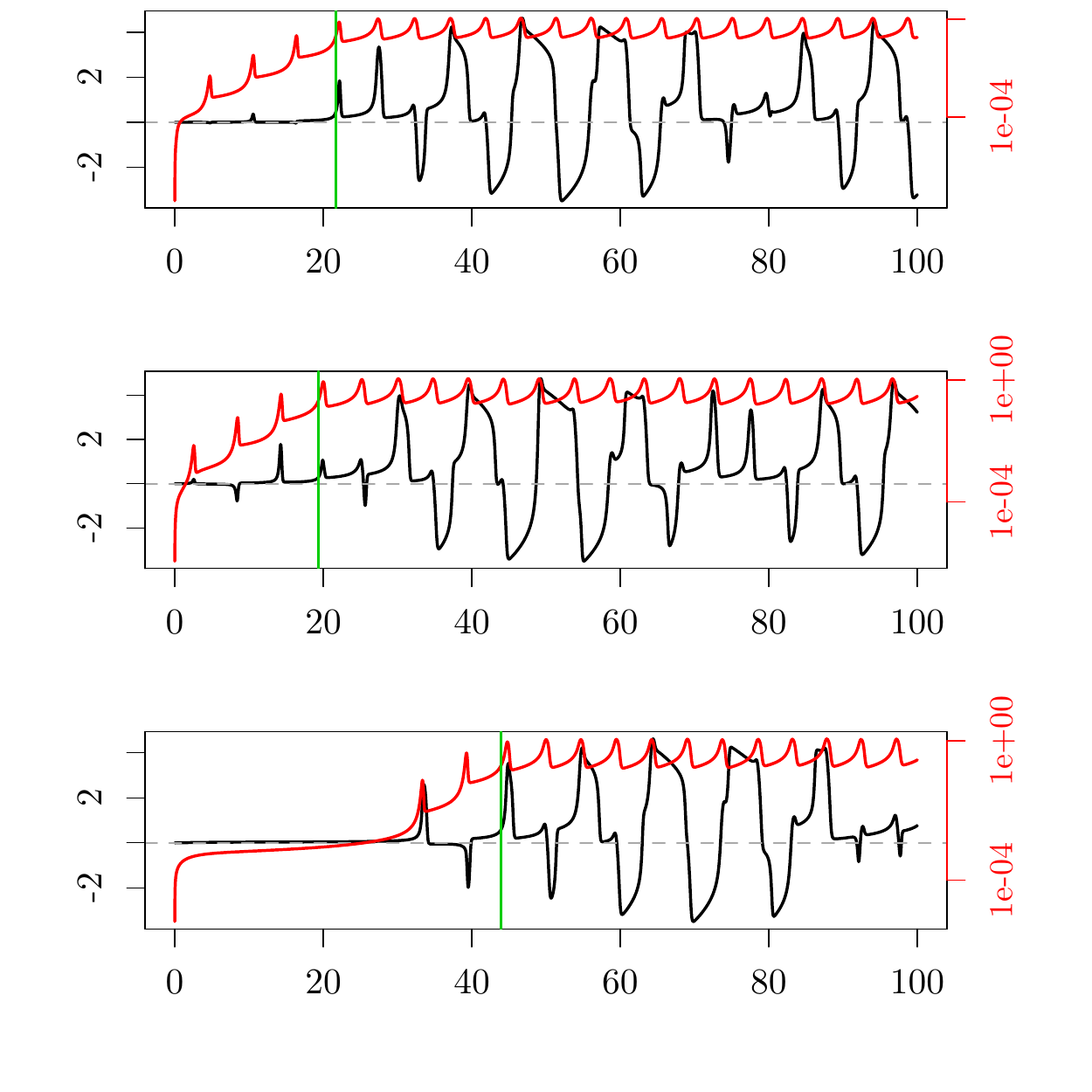}
	\includegraphics[width=0.49\textwidth]{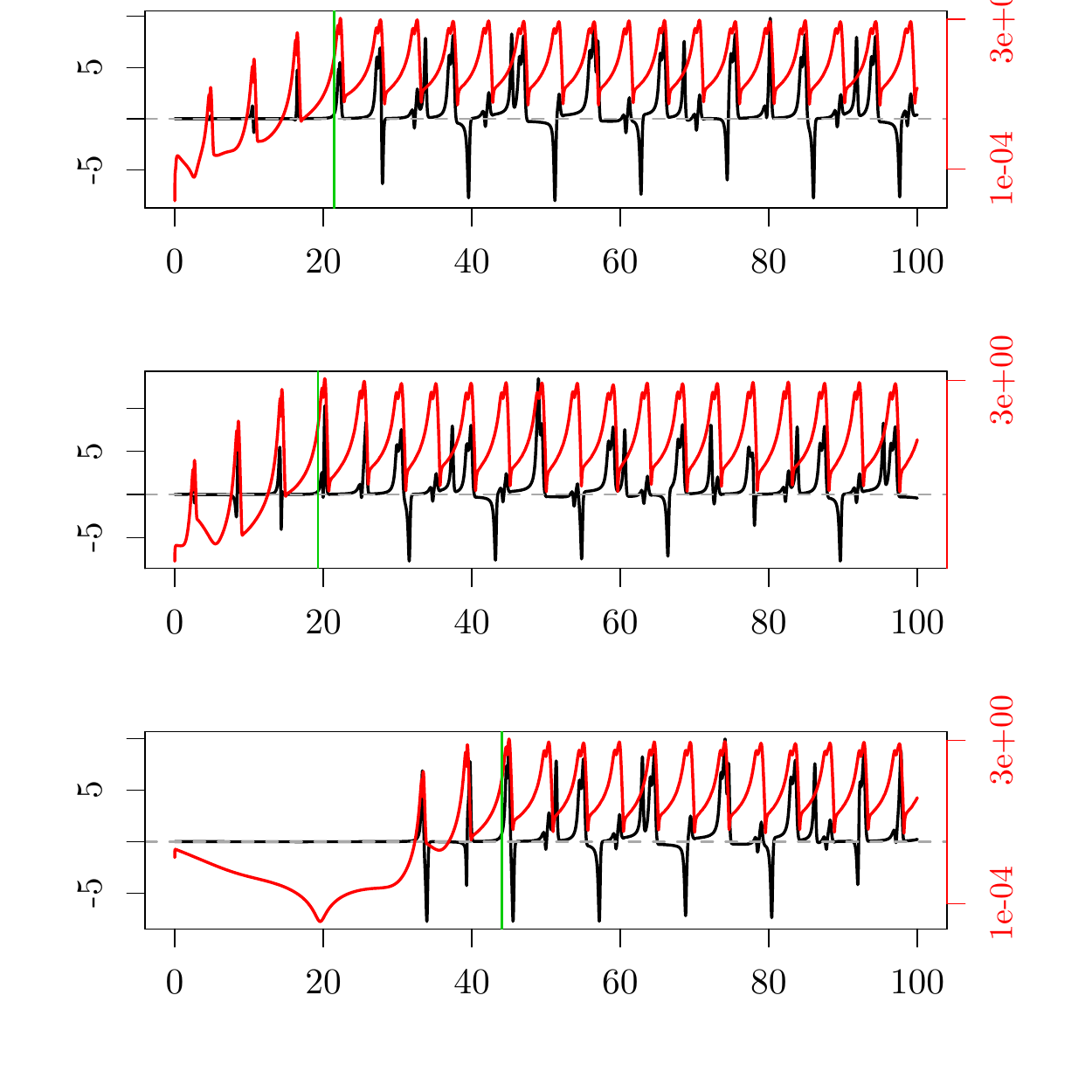} \\
	\includegraphics[width=0.49\textwidth]{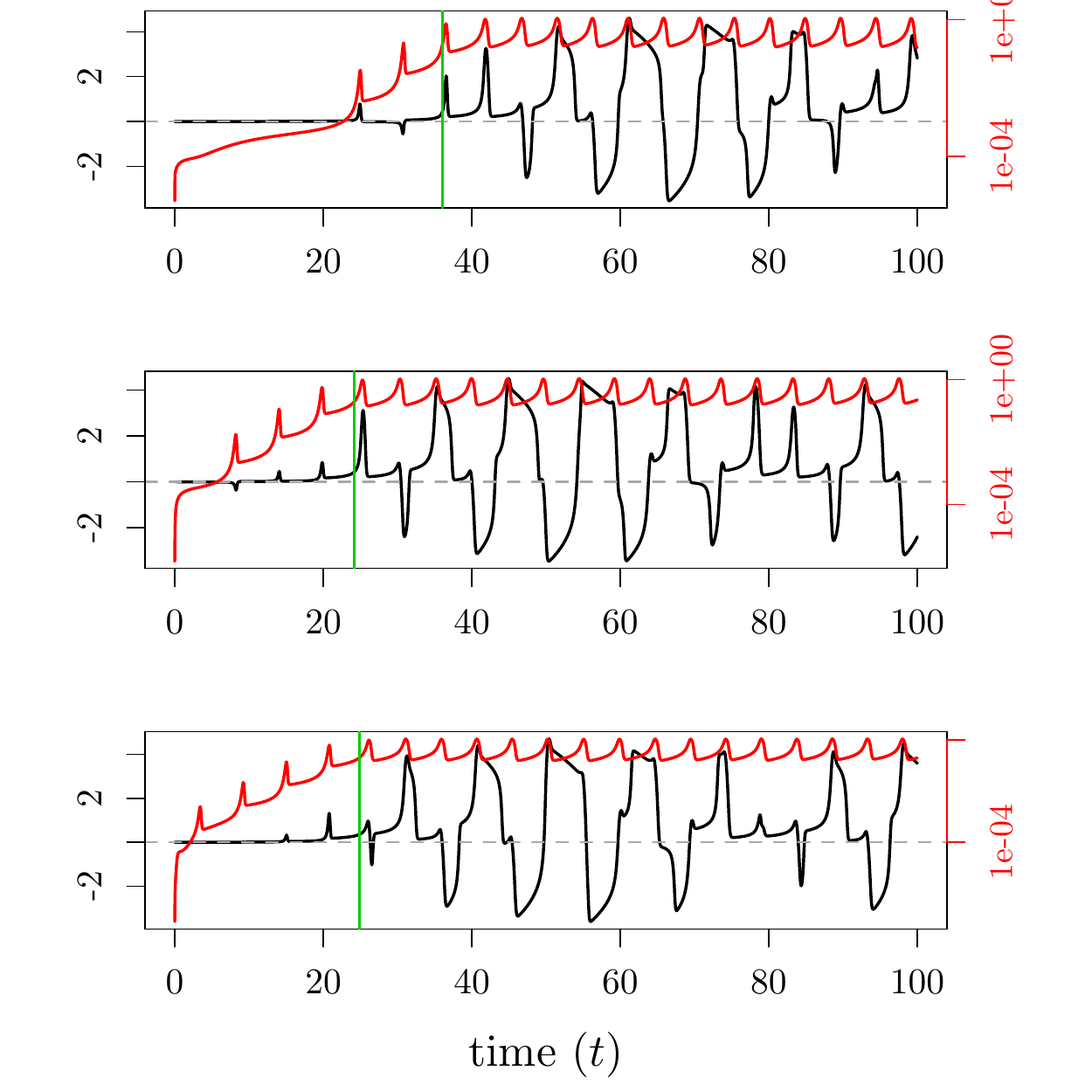}
	\includegraphics[width=0.49\textwidth]{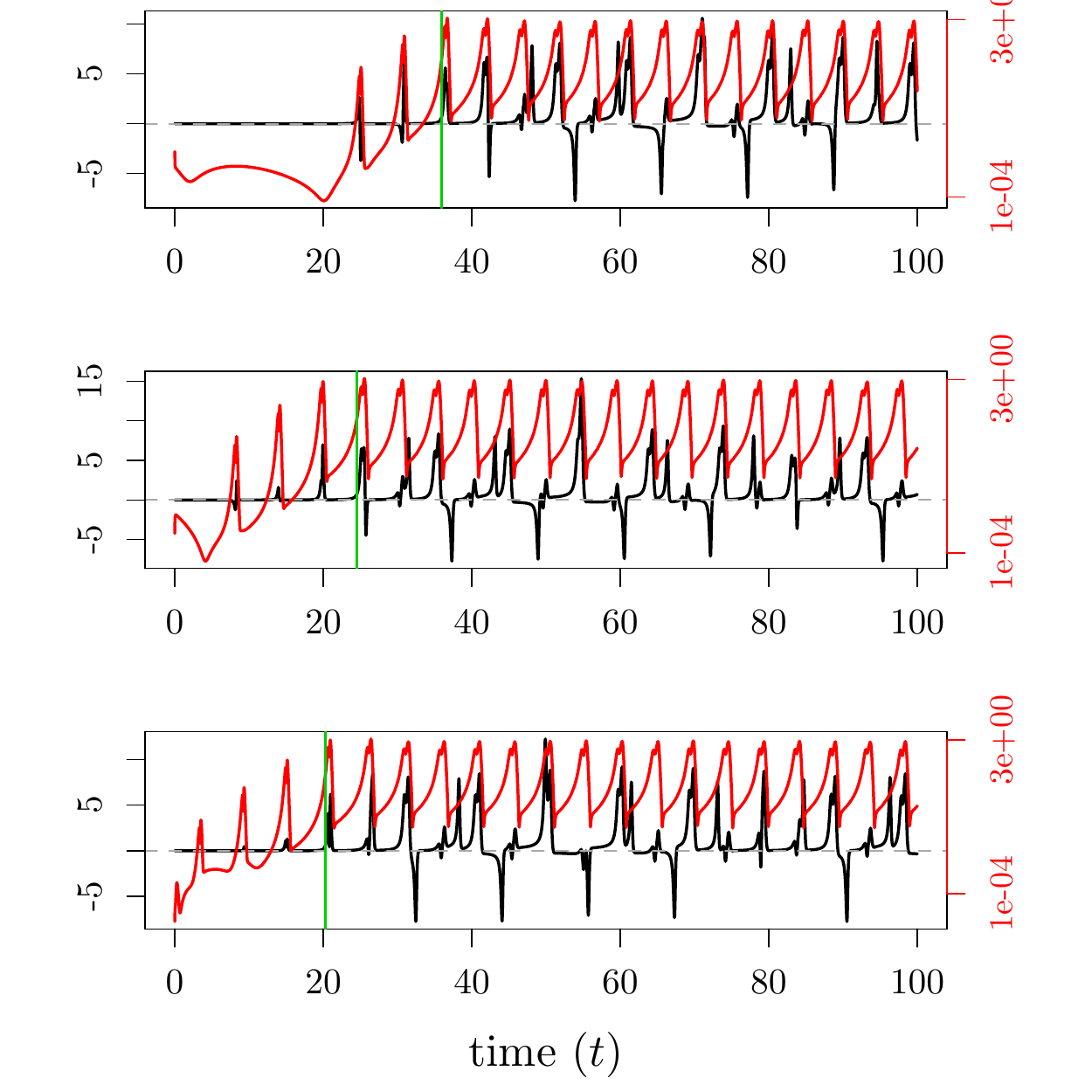} 
	\caption{Prediction of the Van der Pol model ($x_1(t)$: first panel and $x_2(t)$: second panel) with six different initial conditions (corresponding to each row). The black line represents: $\hat{x}_i(t) + 2SD\left(\hat{x}_i(t)\right) - x_i(t)$ and the red line is $SD \left(\hat{x}_i(t) \right)$. The change point is shown by the vertical green line.}
\end{figure}
\section{Laplace's approximation}
\label{append2}
\begin{figure}[htpb] 
	\includegraphics[width=0.49\textwidth]{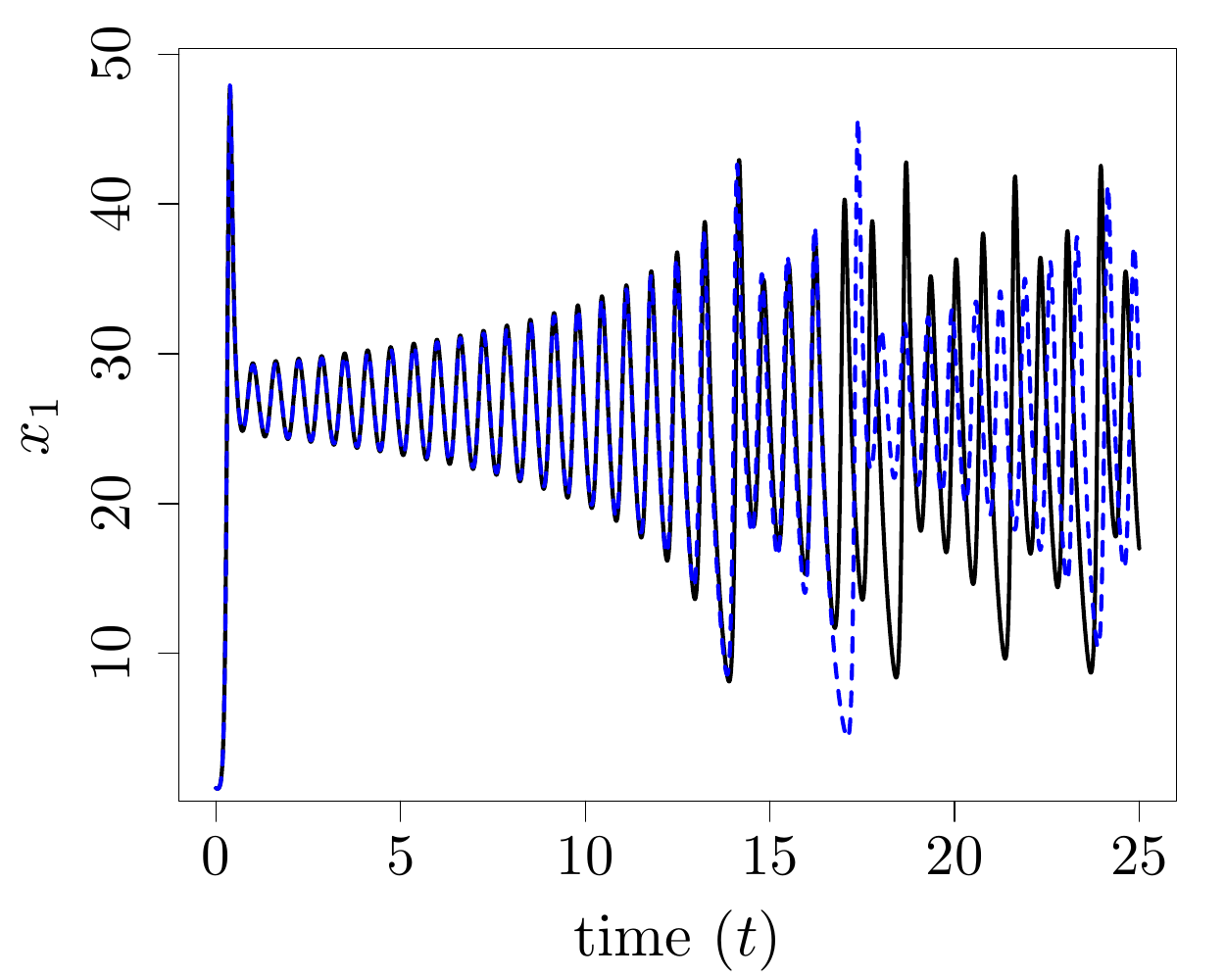}
	\includegraphics[width=0.49\textwidth]{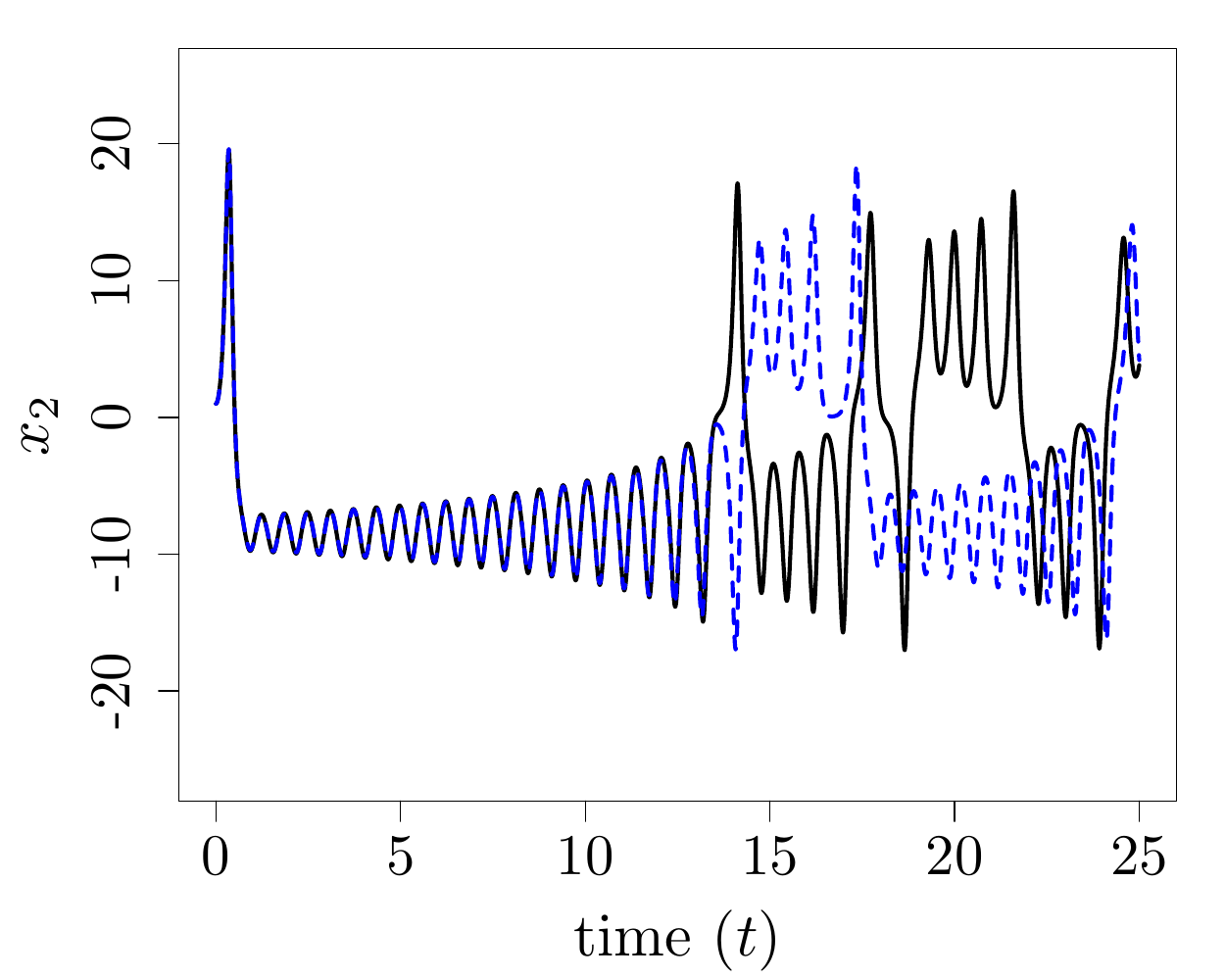} \\
	\includegraphics[width=0.49\textwidth]{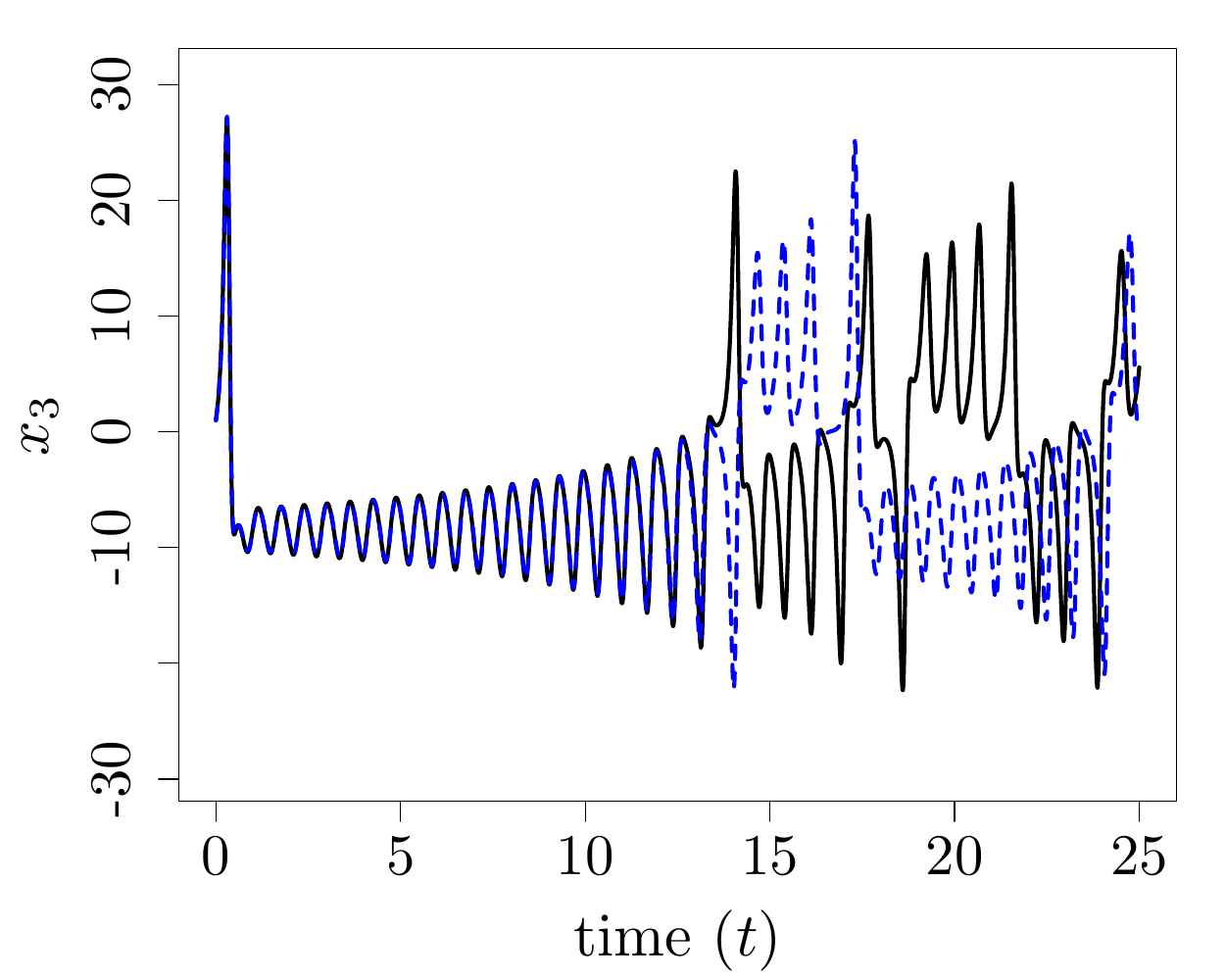}
	\includegraphics[width=0.49\textwidth]{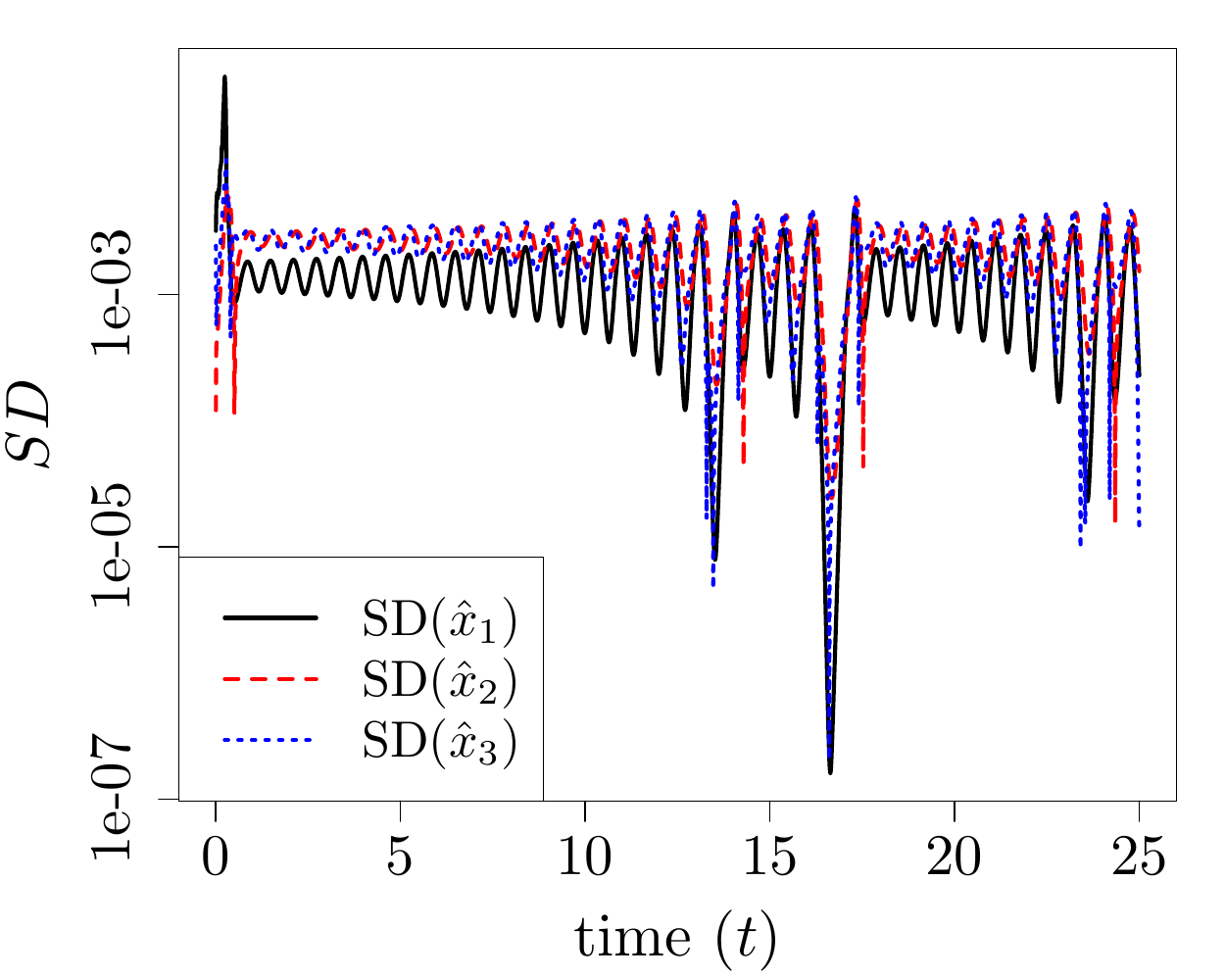} 
	\caption{Emulation of the Lorenz system in which, at each iteration, the output distribution is approximated by the Laplace's method.}
	\label{Laplace_Lorenz_fig}
\end{figure}

\end{appendices}

\newpage
\section*{References}
\bibliography{biblio.bib}
\bibliographystyle{plain}
\end{document}